%% file: main.tex
\documentclass[runningheads]{llncs}

\usepackage{eccv}

\usepackage{eccvabbrv}

\usepackage{graphicx}
\usepackage{booktabs}
\usepackage{wrapfig} %
\usepackage{multirow, makecell}
\usepackage{placeins}
\usepackage[accsupp]{axessibility}  %
\input{math_commands}

\usepackage[pagebackref,breaklinks,colorlinks,citecolor=eccvblue]{hyperref}

\usepackage{orcidlink}

\newcommand{\algoname}[0]{CalibAtt}

\begin{document}

\title{Accelerating Text-to-Video Generation with Calibrated Sparse Attention}

\author{Shai Yehezkel\inst{1,2} \and
Shahar Yadin\inst{1} \and
Noam Elata\inst{1} \and
Yaron Ostrovsky-Berman\inst{1} \and
Bahjat Kawar\inst{1}}

\authorrunning{S.~Yehezkel et al.}

\institute{Apple \and
Tel Aviv University}

\maketitle
\input{figures/video_frame_comp}

\begin{abstract}
    Recent diffusion models enable high-quality video generation, but suffer from slow runtimes.
    The large transformer-based backbones used in these models are bottlenecked by spatiotemporal attention.
    In this paper, we identify that a significant fraction of token-to-token connections consistently yield negligible scores across various inputs, and their patterns often repeat across queries.
    Thus, the attention computation in these cases can be skipped with little to no effect on the result.
    This observation continues to hold for connections among local token blocks.
    Motivated by this, we introduce \algoname, a training-free method that accelerates video generation via calibrated sparse attention. \algoname\ performs an offline calibration pass that identifies block-level sparsity and repetition patterns that are stable across inputs, and compiles these patterns into optimized attention operations for each layer, head, and diffusion timestep.
    At inference time, we compute the selected input-dependent connections densely, and skip the unselected ones in a hardware-efficient manner.
  Extensive experiments on Wan 2.1 14B, Mochi 1, and few-step distilled models at various resolutions show that \algoname\ achieves up to $1.58\times$ end-to-end speedup, outperforming existing training-free methods while maintaining video generation quality and text-video alignment.
\end{abstract}

\section{Introduction}

Transformers~\cite{vaswani2017attention} have been established as a foundational architecture, revolutionizing many domains such as natural language processing~\cite{brown2020language, grattafiori2024llama}, computer vision~\cite{dosovitskiy2021an, oquab2024dinov2}, and generative models such as image and video diffusion models~\cite{ho2020denoisingdiffusionprobabilisticmodels, peebles2023scalable, esser2024scaling, wan2025}.
Their state-of-the-art performance, however, comes at a significant inference-time computational cost.
The self-attention mechanism, central to the transformer architecture, has a quadratic complexity with respect to sequence length, making it particularly challenging for longer sequences inherent in many applications.
In this work, we propose training-free optimizations for self-attention, focusing on the video generation application for several reasons: 
(i) they represent a powerful and increasingly useful tool for content generation;
(ii) they inherently require long sequences to support higher resolutions and larger frame counts;
and (iii) video data often contains high levels of spatiotemporal redundancy, making it an ideal candidate for compute savings. %

To mitigate the computational demands of attention, various innovations have been proposed~\cite{zhang2026survey}.
Most notably, FlashAttention~\cite{dao2022flashattention,dao2023flashattention2fasterattentionbetter,shah2024flashattention3} significantly reduces the memory and runtime of the attention operation. However, it does not change the number of multiplication operations.
Many papers~\cite{zhang2026survey} explore reducing the amount of computation required for attention, often adhering to FlashAttention-friendly constraints leading to improved runtime gains.
Despite their advancements, these methods often require fine-tuning, %
or offer limited acceleration by disabling their methods on certain layers or diffusion timesteps.

In this work, we propose \algoname, a novel training-free method for accelerating attention in video diffusion transformers that automatically calibrates to any model, works in tandem with FlashAttention3~\cite{shah2024flashattention3}, and offers state-of-the-art runtime savings.
We begin by inspecting attention maps in video models and making observations about their typical behaviors at the token level and the block level.
We identify sparse and repetitive attention patterns, many of which are well-replicated across various text prompts and latent initial noise inputs. However, these patterns vary significantly in different transformer layers, attention heads, and in some cases, diffusion timesteps.
Based on these observations, we devise an algorithm that automatically identify data-independent patterns at the FlashAttention block level, and calibrate either an attention mask or mark repeating queries for each layer, head, and timestep combination in a model. %

We conduct extensive experiments on several prominent open-source video diffusion models: Wan 2.1~\cite{wan2025}, Mochi 1~\cite{genmo2024mochi}, and LightX2V~\cite{lightx2v}, a few-step distilled model. We apply our method on different output resolutions and number of diffusion steps.
In all scenarios, \algoname\ consistently maintains the generation quality of the slower dense attention baseline, while achieving state-of-the-art attention sparsity and end-to-end runtime among training-free acceleration methods.
Furthermore, because of the calibration's minimal reliance on heuristics, \algoname's performance persists across models with negligible sensitivity to hyperparameter tuning, and does not require arbitrary exclusion of specific layers or timesteps.
Finally, we conduct hyperparameter ablations that affect \algoname 's inference runtime,
memory footprint, and the one-time cost of the calibration step. %
Notably, the results show that \algoname\ produces impressively sparse masks even at low calibration budgets.

\section{Preliminaries and Related Work}
Attention~\cite{vaswani2017attention} has become a fundamental building block in modern neural architectures. %
Formally, the attention operation is defined over queries $\rmQ \in \R^{N_q \times d}$, keys $\rmK \in \R^{N_{kv} \times d}$, and values $\rmV \in \R^{N_{kv} \times d}$, where $d$ is the feature dimension, and $N_q$, $N_{kv}$ are the query and key-value sequence lengths, respectively.
The attention output $\rmA \in \R^{N_q \times d}$ is computed as
\begin{equation}
\nonumber
  \begin{minipage}{0.48\textwidth}
      \begin{equation}
      \label{eq:p}
      \rmP = \text{softmax} \left( \frac{\rmQ \rmK^\top}{\sqrt{d}} \right),
      \end{equation}
  \end{minipage}%
  \begin{minipage}{0.48\textwidth}
      \begin{equation}
      \label{eq:pv}
          \rmA = \rmP \rmV.
      \end{equation}
  \end{minipage}
\end{equation}

Typically, multi-headed attention is used, repeating this operation for $h$ heads, each with its own set of keys, queries, and values.
As sequence lengths grow, materializing the dense attention matrix
$\rmP \in \mathbb{R}^{ N_q \times N_{kv}}$ quickly becomes prohibitive due to memory constraints, and when $N_q, N_{kv} \gg d$, the cost of processing this matrix also comes to dominate runtime.
FlashAttention~\cite{dao2022flashattention} (FA) addresses this by tiling the computation into blocks, ensuring that only a small portion of the attention matrix resides in fast memory at any given time. This approach avoids the quadratic memory consumption while also providing significant runtime improvements through better hardware utilization.
Concretely, FA partitions the query and key/value sequences into contiguous blocks of size
$B_q$ and $B_{kv}$, respectively.
Let
\begin{equation}
\label{eq:indices}
\mathcal{I}_r = \{\, i \mid r B_q \le i < (r+1)B_q \,\}, \qquad
\mathcal{J}_c = \{\, j \mid c B_{kv} \le j < (c+1)B_{kv} \,\},
\end{equation}
denote the index sets of the $r$-th query block and $c$-th key/value block.
We denote the corresponding query, key, and value blocks by
${\rmQ_r = \rmQ[\mathcal{I}_r,:]}$,
${\rmK_c = \rmK[\mathcal{J}_c,:]}$,
and
${\rmV_c = \rmV[\mathcal{J}_c,:]}$.
The block-wise attention computation is then given by
\begin{equation}
\tilde{\rmA}_{r,c}
=
\text{softmax}\!\left(
\frac{\rmQ_r \rmK_c^\top}{\sqrt{d}} - m_r
\right) \rmV_c,
\end{equation}
where $m_r$ is a running maximum used for numerical stability. The final output for each query block is obtained by rescaling and accumulating contributions across all key-value blocks using online softmax~\cite{milakov2018online}. For full derivations, see~\cite{dao2022flashattention}.

While FlashAttention improves the runtime of attention computation, the total number of multiplication operations remains unchanged.
To reduce computational cost, many recent methods propose constraining attention to specific structures or patterns that reflect assumptions about token-to-token correlations~\cite{Beltagy2020Longformer, liu2021swin, hassani2023neighborhood, huang2019ccnet, wang2020axial, yang2021focal, zhu2021deformable}.
Due to the restrictive nature of these attention patterns, they are typically chosen during architectural design and can rarely be applied to models trained with dense (standard) attention without fine-tuning.

A complementary line of research exploits the inherent sparsity of typical attention maps, enabling acceleration without architectural modifications.
These methods observe that when tokens are mismatched, the corresponding elements of $\rmP$ are nearly zero, and thus their computation in \cref{eq:pv} can be skipped entirely, eliminating redundant operations.
However, identifying which elements or blocks can be safely skipped a priori is non-trivial, and a variety of approaches have been proposed to address this challenge.
For a detailed survey, see~\cite{zhang2026survey}.

Several approaches~\cite{zhang2025sla, wu2025vmoba, liu2025fpsattention, zhang2025vsa} achieve sparsity by finetuning the model to adapt its attention layers.
While these methods can reach high sparsity levels by explicitly optimizing models to do so, they require access to significant computational resources and high-quality training data which may not be available.

Alternatively, training-free sparsity methods apply sparse attention mechanisms to pre-trained models without additional training.
Some works~\cite{zhang2025fast,liradial} pre-compute attention masks based on positional and structural priors and apply them on $\rmP$ at inference time with minimal runtime overhead.
For instance, Radial Attention~\cite{liradial} employs a static sparse attention mask, where each token attends to a window of spatially nearby tokens, with the window size shrinking exponentially with temporal distance.
In contrast, methods with online decisions~\cite{xi2025sparse,zhang2025spargeattn,shmilovich2025liteattention,xu2025xattentionblocksparseattention,xia2025trainingfreeadaptivesparseattention,ohayon2025blocksparseflashattention} %
identify sparsity patterns during inference by profiling attention characteristics, enabling input-adaptive sparsity at the cost of inference-time overhead. For instance, SpargeAttention~\cite{zhang2025spargeattn} estimates block importance online by computing attention on compressed queries and keys, skipping blocks below a threshold, and introduces skip lists to specify which block intervals to compute. %
LiteAttention~\cite{shmilovich2025liteattention} builds upon SpargeAttention, adapting it into a FlashAttention3-based kernel and reusing skip lists across timesteps. Sparse VideoGen 2~\cite{yang2025sparse} clusters tokens, computes attention among cluster centroids, and computes token-level attention only within high-scoring cluster interactions.

In our work, we focus on the idea of using pre-computed masks to accelerate attention without inference-time overhead.
Instead of assuming fixed decay patterns, we compute masks through model-dependent calibration.
Our calibration algorithm adapts to the inherent attention patterns of each model, diffusion timestep, layer, and attention head, while remaining largely agnostic to the specific calibration data, thereby achieving higher sparsity than methods relying on pre-defined masks.

\input{figures/attn_maps}

\section{Method}
\subsection{Initial Inspection and Observations}

We begin by analyzing the typical behavior of attention layers in text-to-video diffusion models for various input prompts and seeds.
Specifically, we focus on the use case of generating $480p$ videos with Wan 2.1 14B~\cite{wan2025}, and we find that the observations generalize across other settings.
We make four key observations.

\paragraph{\textbf{Observation 1: Many attention maps are sparse}.}
We examine the $N \times N$ post-softmax attention matrix $\rmP$.
In the vast majority of cases, we notice that it focuses on a small number of token-to-token interactions (see \cref{fig:attn_grids_2heads}).
The rest of the tokens receive a negligible amount of attention, rendering their inclusion in the output value summation redundant.
Furthermore, we observe that the sparsity remains very apparent, even when considering $B \times B$ blocks in the attention matrix $\rmP$, where each block represents the query-wise average of the post-softmax attention, summed over keys.
This observation highlights attention as a prime candidate for compute savings at inference time, especially since block-wise sparsity is compatible with hardware-friendly efficient attention kernels~\cite{dao2022flashattention, shah2024flashattention3}.
Similar observations have been made in previous work~\cite{xi2025sparse, liradial, zhang2025vsa}.

\paragraph{\textbf{Observation 2: Attention patterns do not replicate across heads, layers, and timesteps.}}
As can be seen in \cref{fig:attn_grids_2heads}, the attention patterns are inconsistent for different heads in the same layer, different layers in the same timestep, as well as for different timesteps in the same layer and head.
In a successful model training run, different attention layers and heads usually learn to focus on different semantic concepts~\cite{voita2019analyzing}.
Hence, when proposing a sparsity-based acceleration technique, we argue it is preferable to fit a specific attention mask for each combination of layer, head, timestep, rather than use a fixed mask for all of them~\cite{liradial}.
That being said, we note that timesteps are a special case. Later timesteps, corresponding to lower noise levels, often exhibit highly similar attention patterns for a given layer and head, a phenomenon also observed in prior work~\cite{shmilovich2025liteattention}.
We analyze this behavior in \cref{subsec:timestep_mask_sharing}.

\input{figures/data_independence}

\paragraph{\textbf{Observation 3: Attention patterns persist across inputs.}}
We observe that the sparsity patterns of attention maps are persistent for various input prompts and initial diffusion noise vectors.
Despite the different semantic contents generated with different prompts, attention connections tend to focus on the same spatio-temporal areas in most cases, as shown in \cref{fig:cross_prompt_attn}.
This observation motivates the one-time offline calibration of data-independent sparse attention masks, to be used online and alleviate the computational cost of video generation inference with no significant overhead. We explore this in \cref{sec:mask_calibration}.

\paragraph{\textbf{Observation 4: Attention patterns repeat across spatial rows.}} %
We further observe that some attention maps exhibit a repetition structure within frame to frame blocks.
In such cases, attention scores in $\rmP$ corresponding to queries of spatial rows within the same frame exhibit highly similar patterns.
For these maps, computing attention for a single representative row per frame may suffice~\cite{willettedelta}, with remaining rows recoverable via repeating the attention output.
We visualize an example of this behavior in \cref{fig:block_repeat}. 
Interestingly, we find that attention maps exhibiting this repetition structure tend to have low block sparsity, suggesting that spatial replication and block sparsification are complementary acceleration strategies. We elaborate on this in \cref{subsec:spatial-rep}.

\subsection{Attention Mask Calibration}
\label{sec:mask_calibration}

\input{figures/similarity_maps}

We now turn to describing \algoname.
Given a model and an inference configuration, our goal is to produce a binary block mask
$\rmM^{(t,l,h)}$ for each diffusion timestep $t$, transformer layer $l$, and attention head $h$,
whose entries indicate whether each block is computed ($1$) or skipped ($0$). %

\input{figures/method_fig}
\vspace{-5pt}
\paragraph{\textbf{Per-prompt energy-based block selection.}}
At a fixed $(t,l,h)$, we consider the post-softmax attention map $\rmP$ (queries $\times$ keys) of size $N \times N$ defined in \cref{eq:p}.
Motivated by \textbf{Observation 1}, we adopt an \emph{energy} viewpoint: for each query, attention mass is often concentrated on a small subset of keys, while the remaining entries contribute negligibly to the output.

To identify which parts of $\rmP$ can be skipped efficiently in hardware, we operate at block granularity.
We partition $\rmP$ into contiguous blocks of size $B \times B$, assuming equal block sizes for queries and keys/values for simplicity, and ablate on different configurations in \cref{subsec:supp_block_size}. This yields $N_B = N/B$ blocks per dimension.
For each query block-row $r$ and key block-column $c$, we define the block energy as the sum of attention scores over keys in $c$, averaged over queries in $r$:
{\setlength{\abovedisplayskip}{3pt}
\setlength{\belowdisplayskip}{3pt}
\begin{equation}
\label{eq:block_energy}
\rmE_{r,c}  
\;=\;
\frac{1}{B}
\sum_{i \in \mathcal{I}_r}
\sum_{j \in \mathcal{J}_c}
\rmP_{ij},
\end{equation}}

where $\mathcal{I}_r$ and $\mathcal{J}_c$ are the query and key/value index sets defined in
Eq.~\ref{eq:indices}\textcolor{red}{.} %
Intuitively, $\rmE_{r,c}$ measures how much attention mass the queries in block-row $r$ allocate to keys in block-column $c$, averaged across the $B$ queries. Averaging this way ensures that a block-row energy $\rmE_r$ sums up to $1$.

For each query block-row $r$, our goal is to keep the smallest number of key blocks whose cumulative energy reaches a prescribed threshold:
\begin{equation}
\label{eq:row_energy_constraint}
\min_{\mathcal{S}_r \subseteq \{1,\dots,N_B\}} |\mathcal{S}_r|
\quad \text{s.t.} \quad
\sum_{c \in \mathcal{S}_r} \rmE_{r,c} \;\ge\; \epsilon .
\end{equation}

We define the energy threshold to be timestep-dependent via $\epsilon(t)$.
This follows prior observations that applying aggressive sparsification too early in the denoising process can lead to noticeably larger degradation in output quality~\cite{liradial,yang2025sparse}. In practice, we use an exponential schedule over timesteps $t \in \{0,\dots,T-1\}$, where $t=0$ denotes the highest-noise step:
\vspace{-2pt}
\begin{equation}
\label{eq:epsilon_schedule}
\epsilon(t)
\;=\;
A + (C-A)\exp\left(-k t/T\right),
\end{equation}
where $T$ is the total number of diffusion steps, and $A$, $C$, and $k$ are schedule hyperparameters.
In practice, $\epsilon(t)$ remains high throughout, ranging from $0.99$ to $0.84$ in Wan~\cite{wan2025}.
We provide details on the schedule hyperparameter search in the supplementary material \cref{subsec:supp_bayes_search}.

Given $\epsilon(t)$ and a text prompt $p$, we compute $\mathcal{S}_r$ by sorting $\{\rmE_{r,c}\}_c$ in descending order and selecting the smallest prefix whose cumulative energy reaches $\epsilon(t)$, marking the selected blocks as \emph{kept} and the remainder as \emph{skipped}.
This algorithm optimally solves \cref{eq:row_energy_constraint} and yields a per-prompt binary block mask $\rmM_{p}^{(t,l,h)} \in \{0,1\}^{N_B \times N_B}$, where $\big[\rmM_{p}^{(t,l,h)}\big]_{r,c}\!=\!1$ indicates that block $(r,c)$ is computed and $0$ indicates it is skipped. This procedure is illustrated in \cref{fig:method-calib} (top).
\vspace{-5pt}

\paragraph{\textbf{Cross-prompt mask aggregation.}}
Motivated by \textbf{Observation 3}, we aim to identify block connections that can be safely skipped in a data-independent manner. 
While the per-prompt energy selection described above yields a binary mask for a single input, our goal is to compile a single calibrated mask per $(t,l,h)$ that is robust across different prompts and noise initializations.

Concretely, for each $(t,l,h)$ we run the energy-based selection procedure on a set of calibration prompts $\mathcal{D}$, producing prompt-specific masks
$\rmM_{p}^{(t,l,h)}$ for all $p \in \mathcal{D}$.
We aggregate these masks by averaging them elementwise:

\begin{equation}
\label{eq:mask_mean}
\bar{\rmM}^{(t,l,h)} \;=\; \frac{1}{|\mathcal{D}|}\sum_{p \in \mathcal{D}} \rmM_{p}^{(t,l,h)}.
\end{equation}

where each entry $\big[\bar{\rmM}^{(t,l,h)}\big]_{r,c} \in [0,1]$ indicates how frequently block $(r,c)$ was computed across prompts.
In all experiments, we use $|\mathcal{D}|=64$ calibration prompts and ablate its effect in \cref{fig:calib_size_rho}.

As shown in \cref{fig:keep_rate_hist}, the distribution is bimodal: many blocks are either almost always computed or almost always skipped, suggesting that many connections can be pruned reliably.
We then obtain the final calibrated mask $\rmM^{(t,l,h)}$ by thresholding each mask block $\big[\bar{\rmM}^{(t,l,h)}\big]_{r,c}$ with an agreement threshold $\rho$:
\begin{equation}
\label{eq:mask_threshold}
\big[\rmM^{(t,l,h)}\big]_{r,c}
\;=\;
\begin{cases}
1, & \big[\bar{\rmM}^{(t,l,h)}\big]_{r,c} \ge \rho\\[2pt]
0, & \big[\bar{\rmM}^{(t,l,h)}\big]_{r,c} < \rho
\end{cases},
\end{equation}
where repeating this for all $(t,l,h)$ yields a dictionary of calibrated masks $\{\rmM^{(t,l,h)}\}$, as illustrated in \cref{fig:method-calib} (center).
At inference time, we retrieve the appropriate mask by indexing this dictionary with the current diffusion step, layer, and head, as illustrated in \cref{fig:method-infer} (top).
The threshold $\rho$ determines how to handle the non-unanimous blocks, controlling the tradeoff between increased sparsity as $\rho$ increases, at the cost of decreased fidelity to the original model behavior since we skip more blocks.
This threshold is largely independent of the underlying algorithm used in per-prompt mask calibration.
In all experiments, we set $\rho=0.5$, and ablate its effect on quality in Sec.~\ref{sec:ablations}.%

\vspace{-5pt}
 \subsection{Spatial Repetition Detection}
  \label{subsec:spatial-rep}

  Motivated by \textbf{Observation 4}, we identify attention maps
  that exhibit repetitive patterns across spatial query positions
  within each frame, enabling additional computational savings
  beyond block skipping. We examine the attention patterns for tokens within a spatial row
   $i$ of a video frame $f$.
  Due to row-major ordering, these tokens have consecutive indices
  \begin{equation}
  \mathcal{I}^{(f, i)} = \left\{l | fHW + iW \leq l < fHW + \left(
  i+1 \right) W\right\},
  \end{equation}
  where $W$ is the number of tokens in a row, and $H$ is the number
   of rows in a frame.
  Their corresponding post-softmax attention values are denoted as
  ${\rmP^{(f, i)} = \mathrm{flatten}\left(\rmP[\mathcal{I}^{(f, i)}, :]\right)}$, referring to
  $\rmP$ from \cref{eq:p}.

  \input{figures/similarities}
We observe that for some attention maps, attention patterns  
  are highly similar across spatial rows within a frame,           
  \textit{i.e.}, $\rmP^{(f, i)} \approx \rmP^{(f,j)}$ for all $i, j \in \{1, \dots, H\}$.                        
  Thus, in these maps, we can select $k$ equispaced anchor spatial rows per frame,       
  compute their attention against all keys and values, and broadcast 
  each result to its nearest spatial rows within that          
  frame. This reduces the number of computed query tokens per frame from $HW$ to $kW$, inducing a sparsity of $1 - k/H$.

To identify which attention maps exhibit this structure, we compute a spatial similarity score $s^{(t,l,h)}$ at each $(t,l,h)$ as follows:
We compute the cosine similarity between each $\rmP^{(f,i)}$ and its nearest anchor row, then average over $f$, $i$, and the input prompts to obtain $s^{(t,l,h)}$. When $s^{(t,l,h)}$ exceeds a threshold $\gamma$, we mark the corresponding $(t,l,h)$ to be computed as spatially repetitive during inference (\cref{fig:method-calib} bottom). We ablate on the number of anchor rows $k$ and the similarity threshold  
  $\gamma$ in \cref{subsec:anchor_rows}, and choose $k=5$, $\gamma=0.87$ in all experiments.

We find that attention maps with high spatial similarity tend to have lower block sparsity, and vice versa, as shown in \cref{fig:rep_vs_sparsity}. Hence, these two acceleration strategies are complementary.
Additionally, maps exhibiting high spatial similarity do so
  consistently across different input prompts, as shown in
  \cref{fig:spatial_rep_consistency}, allowing us to reliably
  identify repetitive maps using a few prompts.

\subsection{Efficient Implementation}

\paragraph{\textbf{Calibration stage.}}
For block-level attention energies $\rmE_{r,c}$ (Eq.~\ref{eq:block_energy}), we implemented a custom CUDA kernel that operates at block granularity and accumulates the required statistics without materializing the full attention matrix $\rmP$.
Spatial similarity scores $s^{(t,l,h)}$ (\cref{subsec:spatial-rep}) are computed separately using an optimized batched PyTorch implementation.

\vspace{-5pt}
\paragraph{\textbf{Inference stage.}}
Several inference frameworks support block-sparse attention computation~\cite{dong2024flexattentionprogrammingmodel,ye2025flashinferefficientcustomizableattention}.
We opt to build a separate custom CUDA kernel for \algoname, implementing optimizations specific to pre-computed masks that vary per timestep, layer, and head.
Our implementation is based on FlashAttention3~\cite{shah2024flashattention3} and block-sparse kernels proposed in prior work~\cite{zhang2025spargeattn, shmilovich2025liteattention}.
The kernel operates on read-only skip lists, precomputed from the calibrated attention masks as such:
For each calibrated binary block mask $\rmM^{(t,l,h)}$ and each query block-row $r$, we encode the contiguous ranges of key block-columns $c$ for which the corresponding attention blocks should be computed.
All skip lists are precomputed once during calibration and preloaded onto the GPU before inference.
During generation, the appropriate skip list is selected based on $(t,l,h)$ and passed to the
attention kernel at launch time (\cref{fig:method-infer} top).
For attention heads flagged as spatially repetitive, we instead use standard FlashAttention3 with a reduced query set containing only the anchor spatial rows, and broadcast the result to all spatial rows within each frame (\cref{fig:method-infer} bottom).
We provide more details and optimizations in \cref{subsec:skip_list_mem_optimization}.

\input{tables/table_wan}

\section{Experiments}
\subsection{Setup}

\paragraph{\textbf{Models.}}
We evaluate \algoname\ on two different model architectures: (i) Wan2.1 14B~\cite{wan2025}, a 14B-parameter model, 
generating $5$-second videos ($81$ frames) in $T=50$ timesteps in two resolutions: $480p$ and $720p$, and (ii) Mochi 1~\cite{genmo2024mochi}, a 10B-parameter model which generates $480p$ videos of $85$ frames in $T=64$ timesteps.
We also evaluate on LightX2V~\cite{lightx2v}, a $4$-step distilled variant of Wan2.1 14B, using the same resolutions and frame counts.

\paragraph{\textbf{Baselines.}}
We use FlashAttention3 (FA3)~\cite{shah2024flashattention3} %
as a reference dense attention baseline.
We additionally compare against \emph{training-free} attention acceleration methods: 
RadialAttention~\cite{liradial}, %
SparseVideoGen2 (SVG2)~\cite{yang2025sparse},  %
and SpargeAttention~\cite{zhang2025spargeattn}. %
All methods are evaluated using identical pretrained checkpoints and model inference settings, 
and they are all implemented with a FlashAttention3 backend or equivalent.
For more details, please refer to Sec.~\ref{subsec:supp_hparams}.

\paragraph{\textbf{Metrics.}}
We evaluate generation quality as well as text--video alignment using VBench~\cite{huang2023vbench} on its official evaluation prompts and metric suite, reporting \emph{Semantic}, \emph{Quality}, and \emph{Total} scores.
We focus on generation quality and prompt adherence rather than per-seed reconstruction metrics (\textit{e.g.}, PSNR), as our goal is to preserve the overall quality level of generated videos rather than to exactly replicate the dense baseline outputs.
To quantify efficiency, we report \emph{Sparsity}, defined as the fraction of skipped query-key interactions out of all spatiotemporal token pairs in self-attention, averaged over all timesteps, layers, and attention heads.
Additionally, we report end-to-end \emph{Latency}, defined as the wall-clock time of the diffusion process to generate a single video, averaged over all evaluation prompts, and \emph{Speedup} relative to dense FA3. %

\paragraph{\textbf{Implementation details.}}
All evaluations are performed on a single NVIDIA H100 GPU. %
A single calibration trial consists of estimating sparse attention masks using a calibration set of $|\mathcal{D}|=64$ prompts sampled from MovieGenBench~\cite{polyak2025moviegencastmedia} with an agreement threshold of $\rho=0.5$.
All masks are defined at FlashAttention block granularity of $128\times128$.
We choose energy threshold schedule $\epsilon(t)$ hyperparameters that perform reasonably well without model-specific tuning.
Specifically, in low-step regimes, we set $A$, $C$, $k$ to constants, and in high-step regimes, we set $C$, $k$ to constants and $A$ to be linearly dependent on the sequence length.
We provide more details in \cref{subsec:supp_bayes_search}.

\input{tables/table_wan_distilled}

\paragraph{\textbf{Results.}}
We summarize our results on high-timestep regimes in \cref{tab:models_combined}.
\algoname\ achieves the best attention sparsity and latency, while maintaining VBench performance comparable to dense FA3, across different models and different resolutions.
Qualitative examples in \cref{fig:two_prompts_four_rows} show that \algoname\ preserves visual fidelity and temporal coherence relative to dense attention.
Additional qualitative comparisons are provided in \cref{sec:supp_qualitative}.
\algoname\ continues to show state-of-the-art results when evaluated on a distilled 4-step version of Wan 2.1 14B (\cref{tab:wan_distilled_combined}).
Previous training-free acceleration methods may offer lower sparsity, or incur an inference-time overhead that becomes significant when applied on few-step generation, whereas \algoname\ continues to provide gains even in this setting.

\vspace{-6pt}

\subsection{Calibration Stage Ablation}
\label{sec:ablations}

\cref{fig:calib_size_rho} analyzes the sensitivity of \algoname\ to the calibration budget and the cross-prompt agreement threshold $\rho$.
We find that the sparsity--quality tradeoff curves stabilize quickly as the number of calibration prompts increases, indicating that reliable data-independent masks can be obtained with a relatively small calibration set, further validating \textbf{Observation 3}.
Varying $\rho$ exposes the expected tradeoff, where lower values enforce stricter cross-prompt consensus on skipped blocks and yield lower sparsity, while overly aggressive thresholds slightly degrades quality score.
Based on this analysis, we use $\rho{=}0.5$ and $64$ calibration prompts in all experiments, which provides a strong balance between sparsity and quality.
Under a restricted budget, we can use $16$ prompts and reduce the Wan 2.1 14B~\cite{wan2025} $720p$ calibration cost from $89.6$ to $13.7$ H100 GPU-hours with minimal effect on sparsity and quality. We elaborate on this in \cref{subsec:supp_calib_cost}.

\vspace{-6pt}
\section{Limitations}
While \algoname\ achieves impressive sparsity and speedup numbers, our algorithm design has three key inherent limitations.
First, due to \algoname\ being a strictly offline method producing data-independent masks, it may miss out on additional potential runtime gains that are specific to certain prompts or prompt types.
Second, the calibration stage requires investing an initial compute budget. This one-time cost can be considered as amortized, which makes it preferable to alternative methods with inference-time overhead.
Third, because it must store the calibrated masks in memory, \algoname\ increases memory usage at inference time. On Wan2.1 14B at $720p$, the mask memory overhead is $21.5$GB, and it can be reduced to $3.6$GB with minimal effect, as we show in \cref{subsec:skip_list_mem_optimization}. %

\input{figures/n_prompts_calib}

\vspace{-10pt}
\section{Conclusion}
\vspace{-4pt}

We present \algoname, a training-free method for accelerating attention in video diffusion transformers. %
Our approach is grounded in the observation that attention sparsity and repetition patterns, while varying across layers, heads, and timesteps, remain largely consistent across different input prompts.
By performing a one-time offline calibration, our algorithm automatically identifies block-level sparsity patterns and spatially repetitive attention heads, to be leveraged for fast inference. %
We additionally contribute a FlashAttention3-based kernel that supports pre-computed block-sparse skip lists varying per timestep, layer, and head.
Notably, \algoname\ is robust across different model architectures, output resolutions, and diffusion configurations, with no manual tuning of layer- or timestep-specific exclusions.

Several future directions related to our work remain open.
First, the training-free calibration can be extended to identify attention redundancies beyond sparsity and spatial row repetition. For example, while our calibration is performed independently per timestep, exploring correlations among attention maps across consecutive timesteps may reveal additional sparsity patterns and further reduce computation.
Second, the memory footprint of calibrated masks could be reduced through more compact skip list representation.
Third, richer parameterizations of the energy threshold schedule $\epsilon(t)$ and the similarity threshold $\gamma$, including layer-, head-, or timestep-dependent variants, may yield improved sparsity--quality tradeoffs.
Finally, while we focus on video diffusion models, the proposed calibration framework could be generalized to other transformer-based contexts such as image diffusion and language models.

\bibliographystyle{splncs04}
\bibliography{main}

\clearpage
\appendix
\input{appendix_content}

\end{document}

%% file: math_commands.tex
\usepackage{amsmath,amsfonts,bm}
\usepackage{nicefrac}

\def\eqref#1{equation~\ref{#1}}

\def\1{\bm{1}}

\def\rmA{{\mathbf{A}}}

\def\rmE{{\mathbf{E}}}

\def\rmK{{\mathbf{K}}}

\def\rmM{{\mathbf{M}}}

\def\rmP{{\mathbf{P}}}
\def\rmQ{{\mathbf{Q}}}

\def\rmV{{\mathbf{V}}}

\DeclareMathAlphabet{\mathsfit}{\encodingdefault}{\sfdefault}{m}{sl}
\SetMathAlphabet{\mathsfit}{bold}{\encodingdefault}{\sfdefault}{bx}{n}

\newcommand{\R}{\mathbb{R}}

\usepackage{xspace}
\makeatletter
\DeclareRobustCommand\onedot{\futurelet\@let@token\@onedot}
\def\@onedot{\ifx\@let@token.\else.\null\fi\xspace}

\makeatother

%% file: figures/video_frame_comp.tex
\def\promptA{Panda}
\def\promptB{Astro}

\def\marginMetrics{5.25em}
\def\textA{A panda drinking coffee in a cafe in Paris.}
\def\textB{An astronaut flying in space, featuring a steady and smooth perspective.}

\begin{figure}[t]
\centering
\vspace{-1.0em}

\setlength{\fboxsep}{2pt}

\parbox{\linewidth}{%

\makebox[\linewidth]{%
    \llap{\textcolor{red}{sparsity=0\%}\,\,\hspace{\marginMetrics}}
    \makebox[0pt][c]{Dense Attention}%
    \makebox[0pt][l]{\hspace{\marginMetrics}
        \textcolor{red}{latency=20m44s}
    }%
}

\vspace{0.2em}
\includegraphics[width=0.195\linewidth]{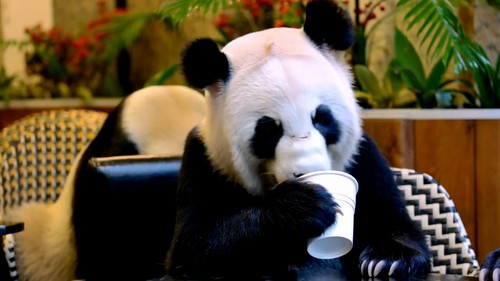}\hfill
\includegraphics[width=0.195\linewidth]{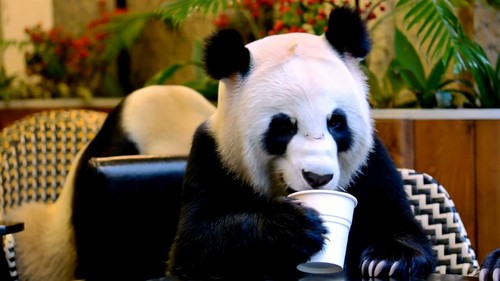}\hfill
\includegraphics[width=0.195\linewidth]{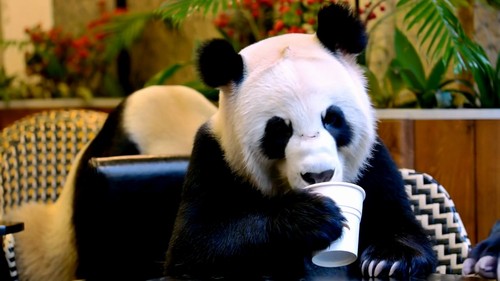}\hfill
\includegraphics[width=0.195\linewidth]{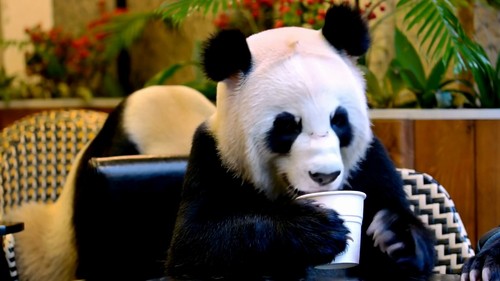}\hfill
\includegraphics[width=0.195\linewidth]{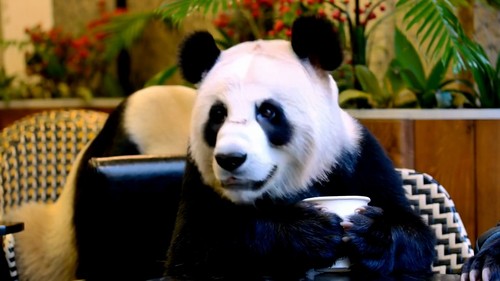}

\vspace{0.4em}

\makebox[\linewidth]{%
    \llap{\textcolor{green!60!black}{sparsity=62\%}\,\,\hspace{\marginMetrics}}
    \makebox[0pt][c]{\algoname}%
    \makebox[0pt][l]{\hspace{\marginMetrics}
        \textcolor{green!60!black}{latency=13m05s}
    }%
}

\vspace{0.2em}
\includegraphics[width=0.195\linewidth]{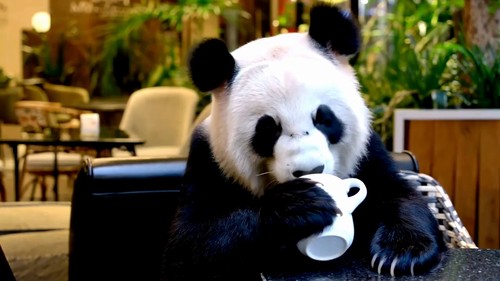}\hfill
\includegraphics[width=0.195\linewidth]{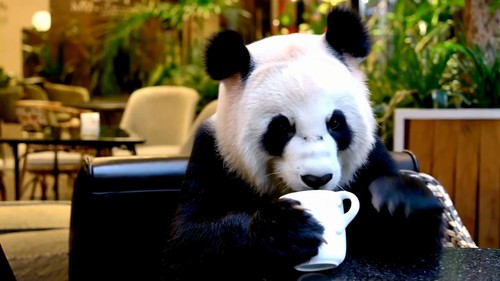}\hfill
\includegraphics[width=0.195\linewidth]{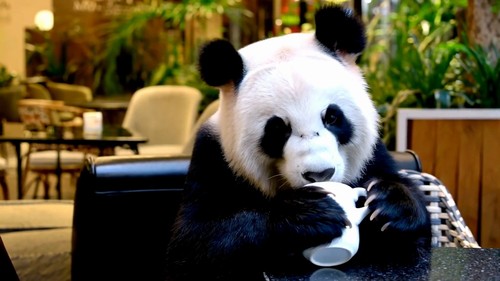}\hfill
\includegraphics[width=0.195\linewidth]{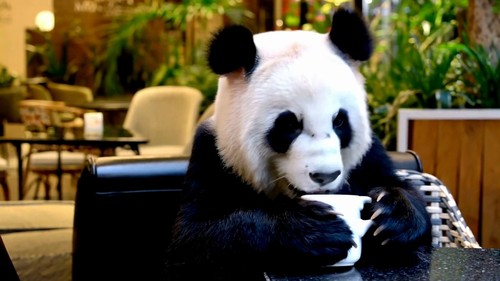}\hfill
\includegraphics[width=0.195\linewidth]{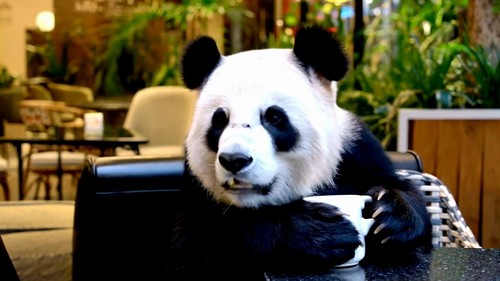}

\parbox{\linewidth}{\centering
\footnotesize \textit{\textcolor{black!70}{``\textA'' (\textbf{720p})}}
}

}

\vspace{1.5em}

\parbox{\linewidth}{%

\makebox[\linewidth]{%
    \llap{\textcolor{red}{sparsity=0\%}\,\,\hspace{\marginMetrics}}
    \makebox[0pt][c]{Dense Attention}%
    \makebox[0pt][l]{\hspace{\marginMetrics}
        \textcolor{red}{latency=6m03s}
    }%
}

\vspace{0.2em}
\includegraphics[width=0.195\linewidth]{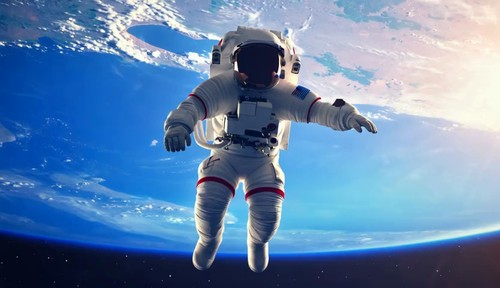}\hfill
\includegraphics[width=0.195\linewidth]{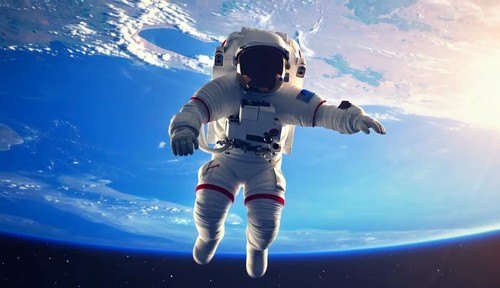}\hfill
\includegraphics[width=0.195\linewidth]{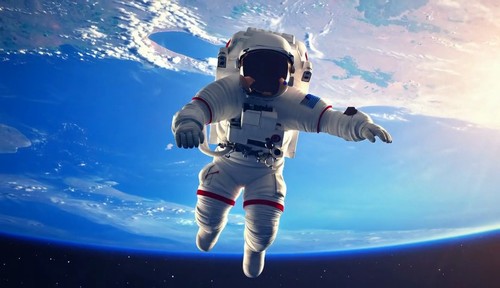}\hfill
\includegraphics[width=0.195\linewidth]{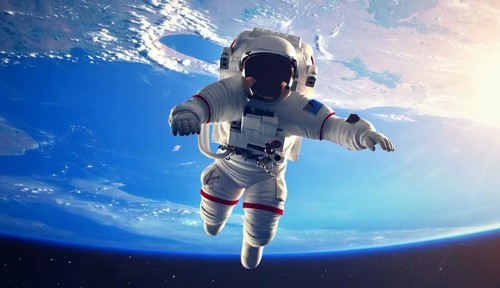}\hfill
\includegraphics[width=0.195\linewidth]{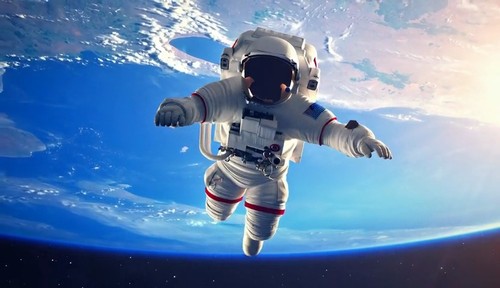}

\vspace{0.4em}

\makebox[\linewidth]{%
    \llap{\textcolor{green!60!black}{sparsity=68\%}\,\,\hspace{\marginMetrics}}
    \makebox[0pt][c]{\algoname}%
    \makebox[0pt][l]{\hspace{\marginMetrics}
        \textcolor{green!60!black}{latency=4m10s}
    }%
}

\vspace{0.2em}
\includegraphics[width=0.195\linewidth]{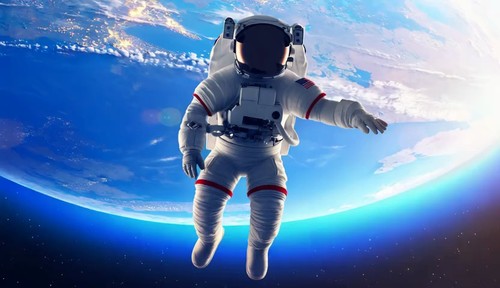}\hfill
\includegraphics[width=0.195\linewidth]{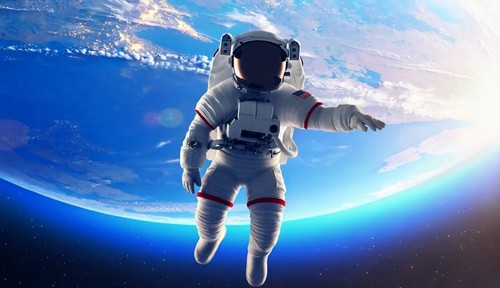}\hfill
\includegraphics[width=0.195\linewidth]{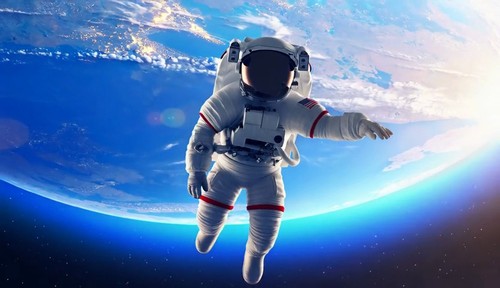}\hfill
\includegraphics[width=0.195\linewidth]{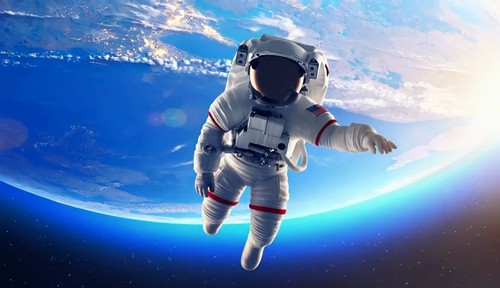}\hfill
\includegraphics[width=0.195\linewidth]{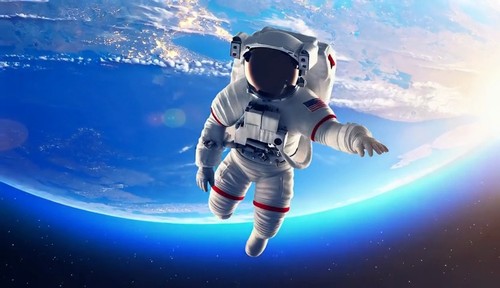}

\parbox{\linewidth}{\centering
\footnotesize \textit{\textcolor{black!70}{``\textB'' (\textbf{480p})}}
}

}

\caption{
Comparison of two prompts and resolutions generated with the same seed on Wan2.1 14B text-to-video.
\algoname\ achieves higher attention sparsity and lower end-to-end latency while maintaining visual quality and prompt alignment.
\vspace{-1em}
}
\label{fig:two_prompts_four_rows}

\end{figure}

%% file: figures/attn_maps.tex
\begin{figure*}[t]
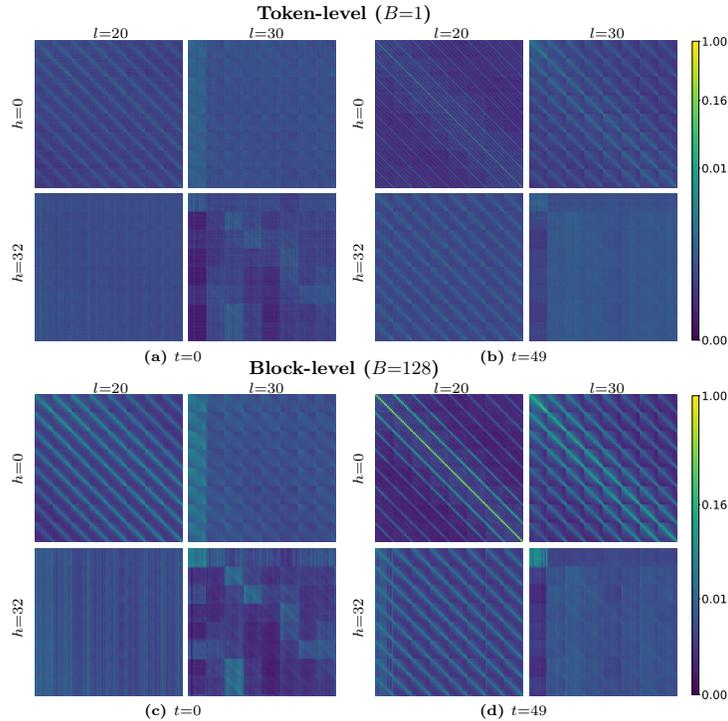

\centering

\def\FigScale{0.80} %

\def\TimeA{0}
\def\TimeB{49}
\def\LayerA{20}
\def\LayerB{30}

\def\DirB{attn_individual_128_scale}
\def\DirC{attn_individual_1_scale}

\def\CapB{{Block-level} ($B{=}128$)}
\def\CapC{{Token-level} ($B{=}1$)}

\def\HeadA{0}
\def\HeadB{32}

\newcommand{\AttnImg}[4]{%
\includegraphics[width=0.44\linewidth]{images/attns_maps/#2/attn_t#3_l#4_h#1_p0.pdf}
}

\newcommand{\AttnCbar}[1]{%
\includegraphics[height=0.56\linewidth]{images/attns_maps/#1/colorbar_attn_token.pdf}
}

\newcommand{\AttnGridAtTime}[2]{%
\begin{minipage}[t]{\linewidth}
\centering

\makebox[\linewidth]{%
    \hspace*{0.095\linewidth}%
    \makebox[0.44\linewidth][c]{\scriptsize $l{=}\LayerA$}%
    \hfill
    \makebox[0.44\linewidth][c]{\scriptsize $l{=}\LayerB$}%
}\\

\makebox[\linewidth]{%
    \makebox[0.085\linewidth][c]{\rotatebox{90}{\scriptsize \hspace{25pt} $h{=}\HeadA$}}%
    \hspace*{0.01\linewidth}%
    \AttnImg{\HeadA}{#2}{#1}{\LayerA}\hfill
    \AttnImg{\HeadA}{#2}{#1}{\LayerB}%
}\\[2pt]

\makebox[\linewidth]{%
    \makebox[0.085\linewidth][c]{\rotatebox{90}{\scriptsize \hspace{25pt} $h{=}\HeadB$}}%
    \hspace*{0.01\linewidth}%
    \AttnImg{\HeadB}{#2}{#1}{\LayerA}\hfill
    \AttnImg{\HeadB}{#2}{#1}{\LayerB}%
}

\end{minipage}
}

\newcommand{\AttnRowForSetting}[2]{%
\vspace{2pt}
\begin{tabular*}{\textwidth}{@{\extracolsep{\fill}}ccc@{}}

\multicolumn{2}{c}{\textbf{#2}} & \\[-2pt]

\begin{subfigure}[t]{0.46\textwidth}
\centering
\AttnGridAtTime{\TimeA}{#1}
\subcaption{$t{=}\TimeA$}
\end{subfigure}
&
\begin{subfigure}[t]{0.46\textwidth}
\centering
\AttnGridAtTime{\TimeB}{#1}
\subcaption{$t{=}\TimeB$}
\end{subfigure}
& \hspace{-125pt}
\begin{subfigure}[t]{0.775\textwidth}
\centering
\vspace{-2pt} %
\AttnCbar{#1}
\end{subfigure}

\end{tabular*}
}

\resizebox{\FigScale\textwidth}{!}{%
\begin{minipage}{\textwidth}

\AttnRowForSetting{\DirC}{\CapC}

\vspace{-8pt}

\AttnRowForSetting{\DirB}{\CapB}

\end{minipage}
}

\vspace{-4pt}
\caption{\textbf{Attention patterns across timesteps ($t$), layers ($l$), and heads ($h$).}
We compare post-softmax attention maps (queries $\times$ keys) for different $t,l,h$ for the same prompt with Wan 2.1 14B~\cite{wan2025}.
Each row fixes the attention granularity (token-level or block-level).
For ease of visualization, we show the first $12544$ tokens out of the full sequence length of ${32760}$.
Notably, the large block structure visible in some of the maps reflects intra-frame token correspondences in the video.
}
\label{fig:attn_grids_2heads}
\vspace{-6pt}
\end{figure*}

%% file: figures/data_independence.tex
\begin{figure}[t]
    \centering
    \vspace{-6pt}
    \begin{subfigure}[t]{0.38\linewidth}
        \centering
        \caption{Attention maps across prompts}
        \label{fig:cross_prompt_attn}
        \setlength{\tabcolsep}{1pt}
        \begin{tabular}{@{}cc@{}}
            \includegraphics[width=0.48\linewidth]{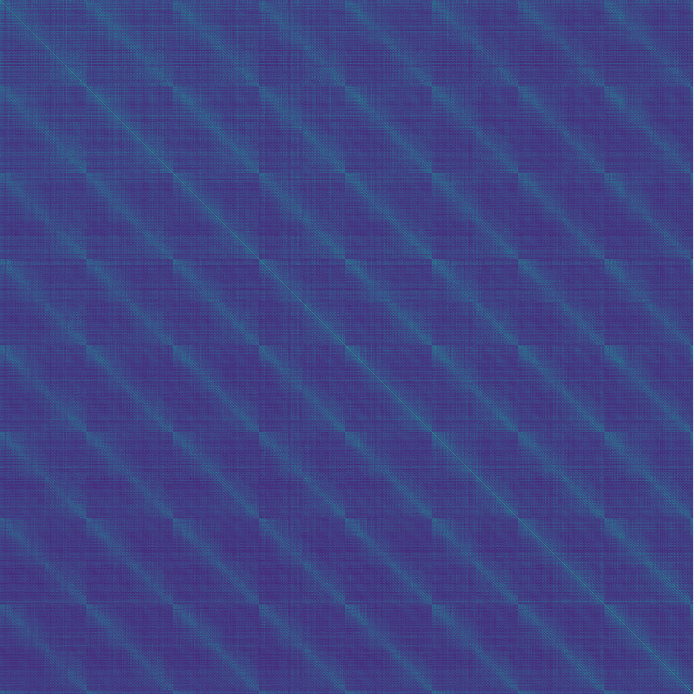} &
            \includegraphics[width=0.48\linewidth]{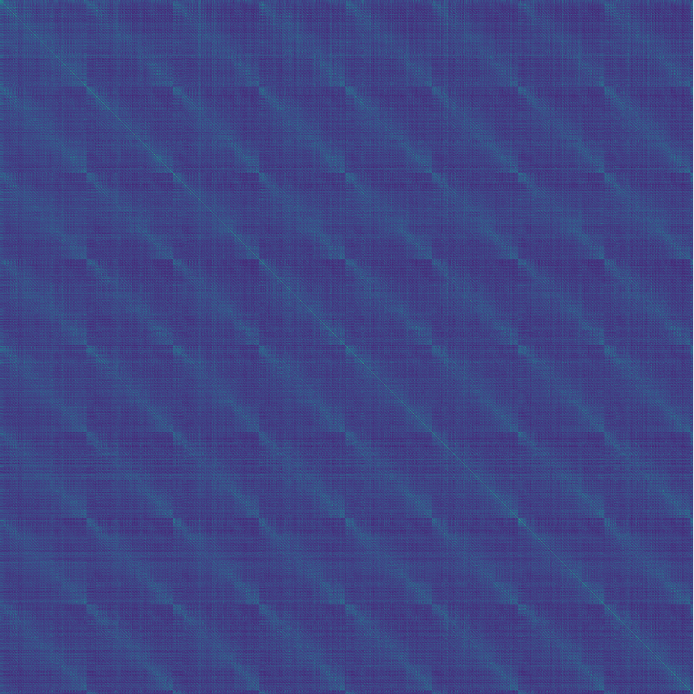} \\
            \includegraphics[width=0.48\linewidth]{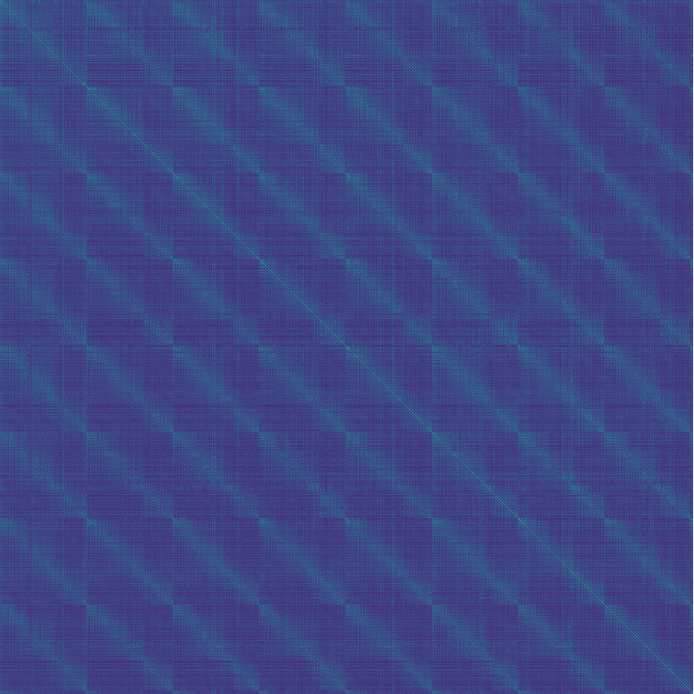} &
            \includegraphics[width=0.48\linewidth]{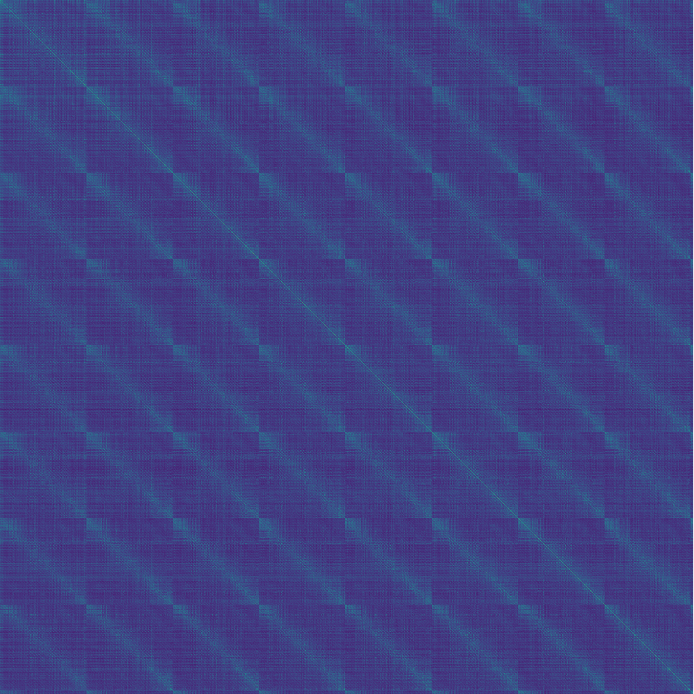} \\
        \end{tabular}
    \end{subfigure}
    \hfill
    \begin{subfigure}[t]{0.58\linewidth}
        \centering
        \caption{Block keep-rate histogram}
        \label{fig:keep_rate_hist}
        \includegraphics[width=\linewidth]{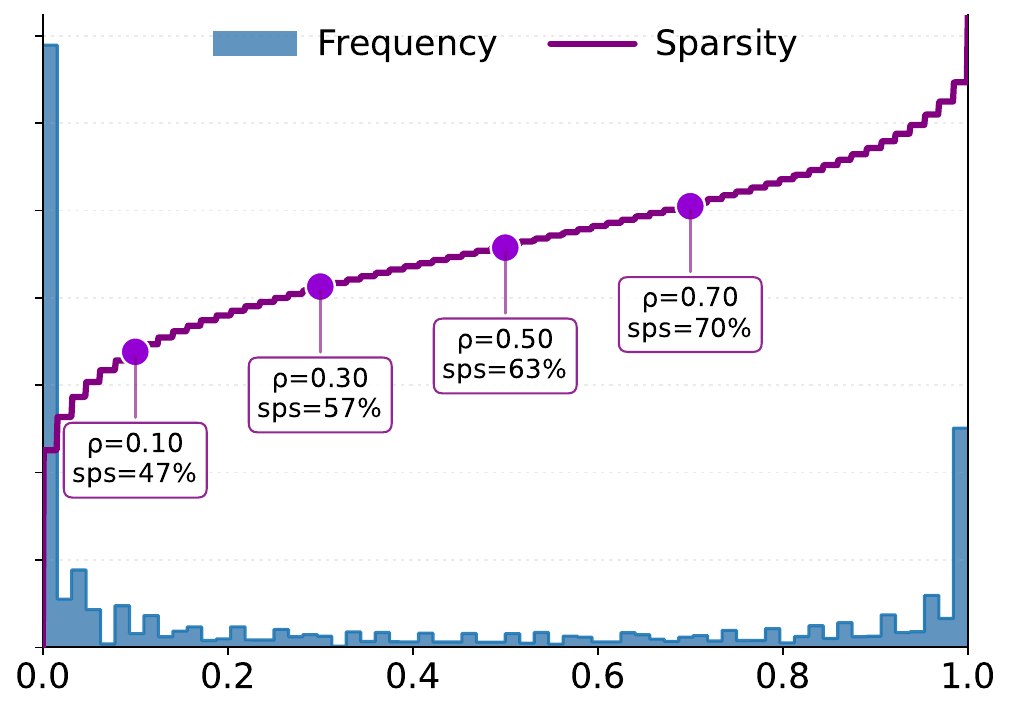}
    \end{subfigure}
    \vspace{-6pt}
    \caption{\textbf{Data-independence of block sparsity.}
    \textbf{(a)} Attention maps from layer 20, head 24, at timestep 10 across four different prompts, showing consistent sparsity patterns.
    \textbf{(b)} Histogram showing how often each block is marked to be kept across calibration prompts. A value of 0 means the block is skipped for all prompts, while 1 means it is always computed.
    Many blocks cluster near 0 or 1, indicating a largely data-independent sparsity pattern.
    The curve (purple) shows the cumulative fraction of blocks skipped under different agreement thresholds.}
    \vspace{-8pt}
\end{figure}

%% file: figures/similarity_maps.tex
\begin{figure}[t]
    \centering
    \begin{minipage}[c]{0.44\textwidth}
        \centering
        \textbf{(a)}\\[2pt]
        \includegraphics[width=\textwidth]{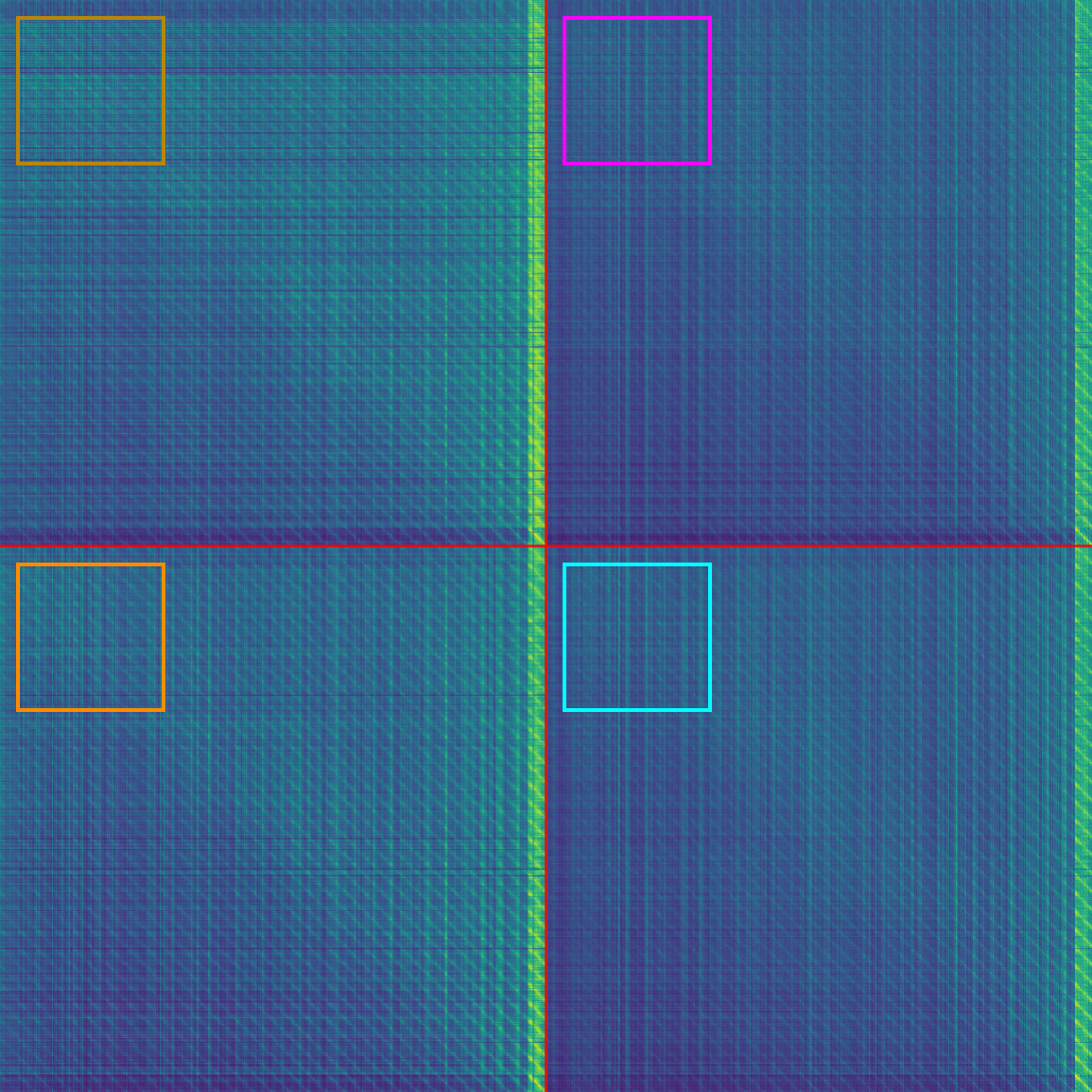}
    \end{minipage}%
    \hfill
    \begin{minipage}[c]{0.445\textwidth}
        \centering
        \textbf{(b)}\\[2pt]
        \includegraphics[width=\textwidth]{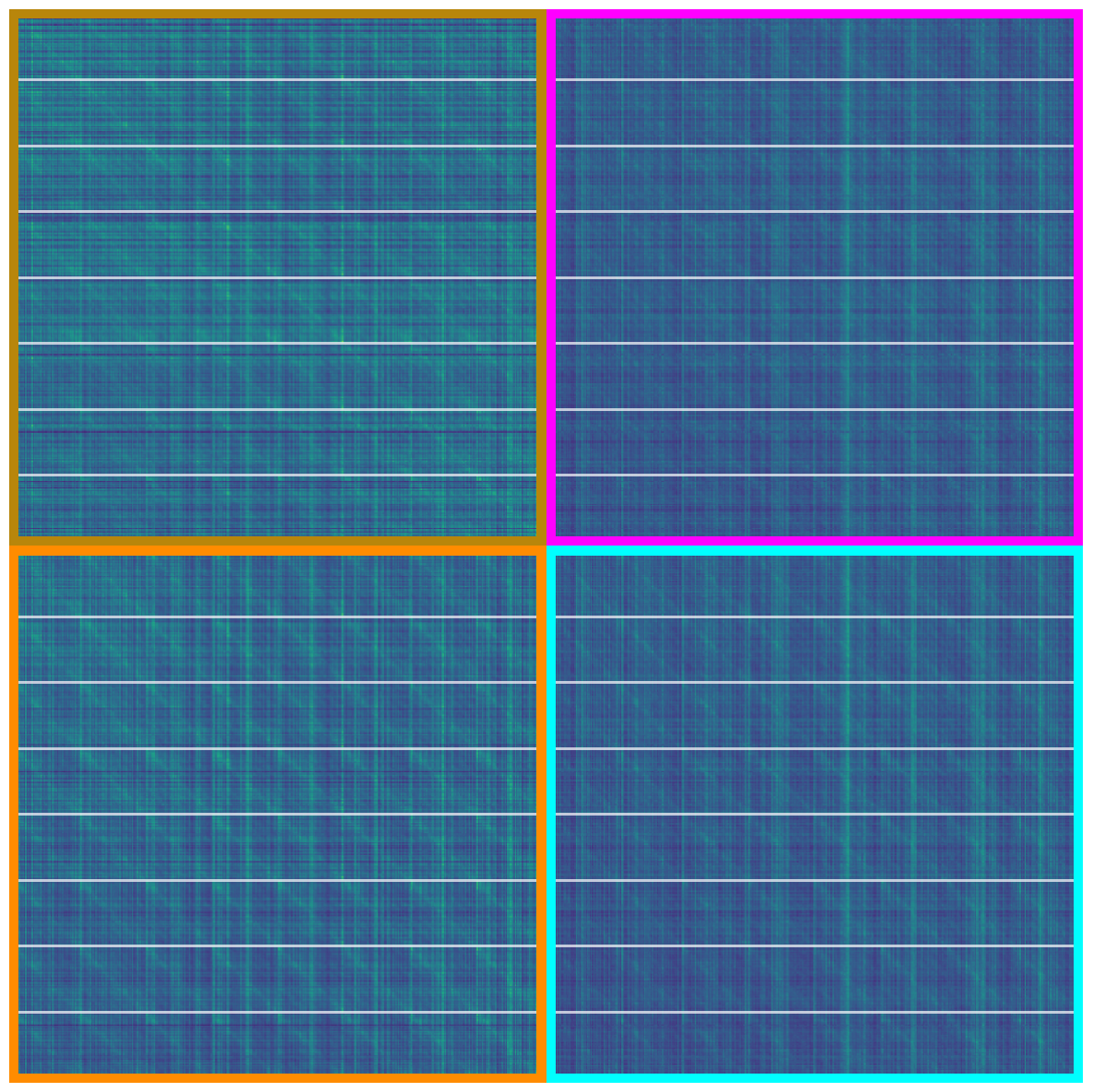}
    \end{minipage}
    \begin{minipage}[c]{0.05\textwidth}
        \centering
        \vspace{14pt} %
        \includegraphics[height=9.15\linewidth]{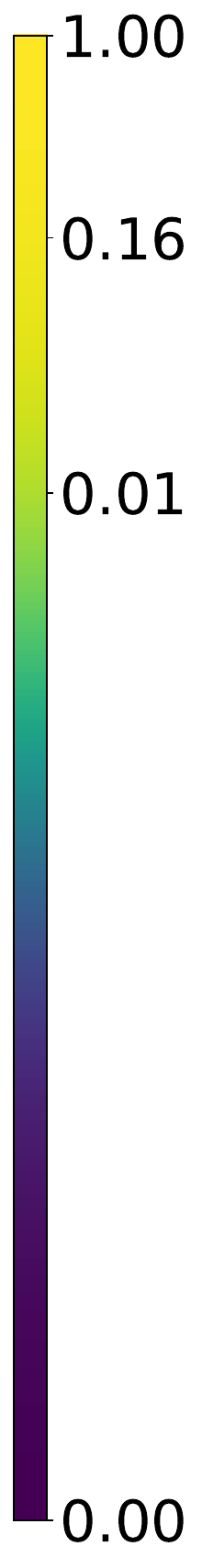}
    \end{minipage}
    \caption{\textbf{Spatial repetition within frames.} \textbf{(a)} Token-level attention map at layer 30, head 24, timestep 0, showing two frames ($3120$ of $32760$ tokens). Frame boundaries are marked in red.
    \textbf{(b)} Zoomed-in slices from each frame-to-frame block with white grid lines separating spatial rows. The attention pattern repeats across spatial rows within each query frame.}
    \label{fig:block_repeat}
\end{figure}

%% file: figures/method_fig.tex
\begin{figure}[t]
    \centering
    \begin{subfigure}[b]{0.713\textwidth}
        \centering
        \includegraphics[width=\textwidth]{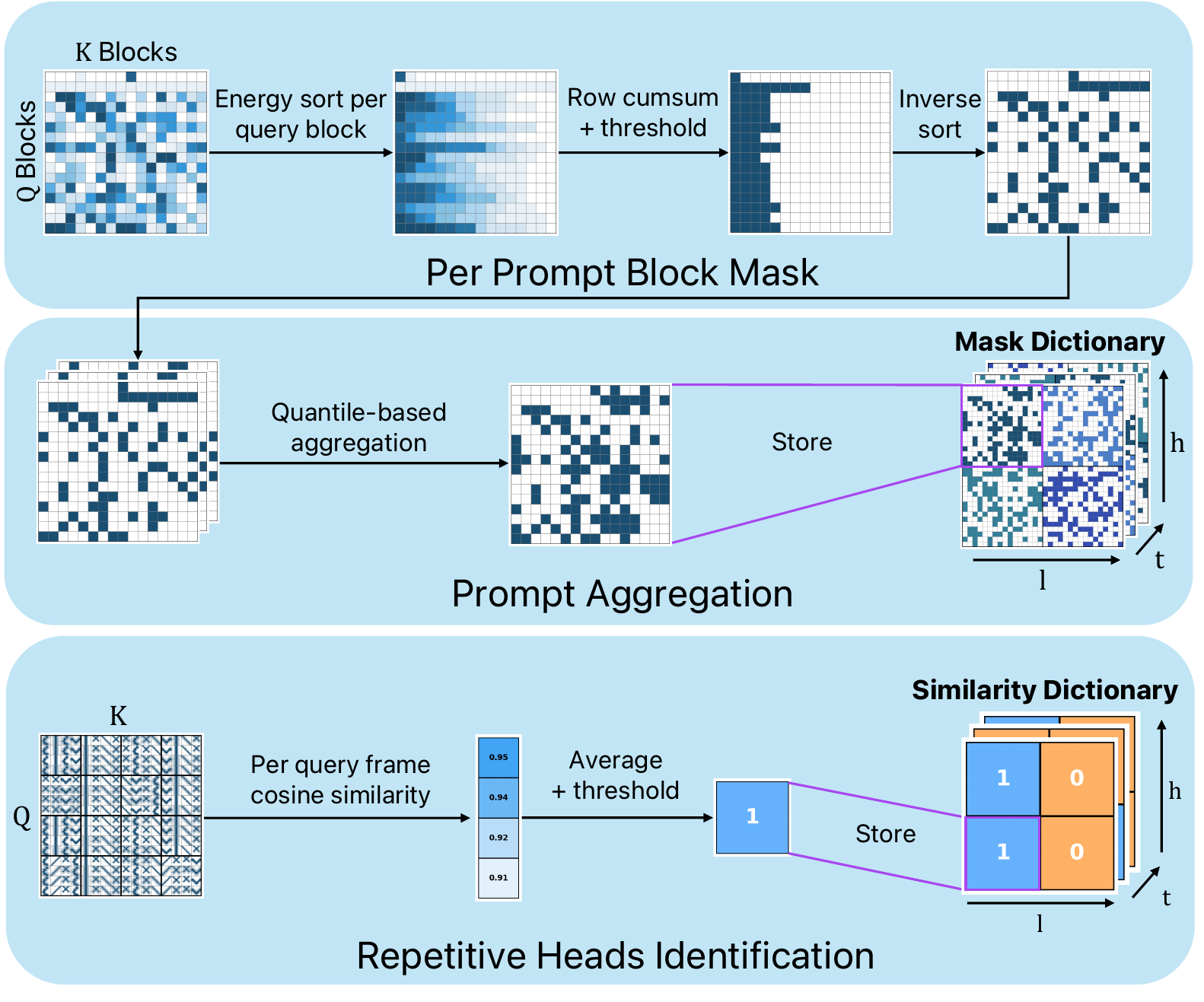}
        \caption{Offline calibration per timestep/layer/head}
        \label{fig:method-calib}
    \end{subfigure}
    \hspace{-5.0pt}
    \begin{subfigure}[b]{0.277\textwidth}
        \centering
        \includegraphics[width=\textwidth]{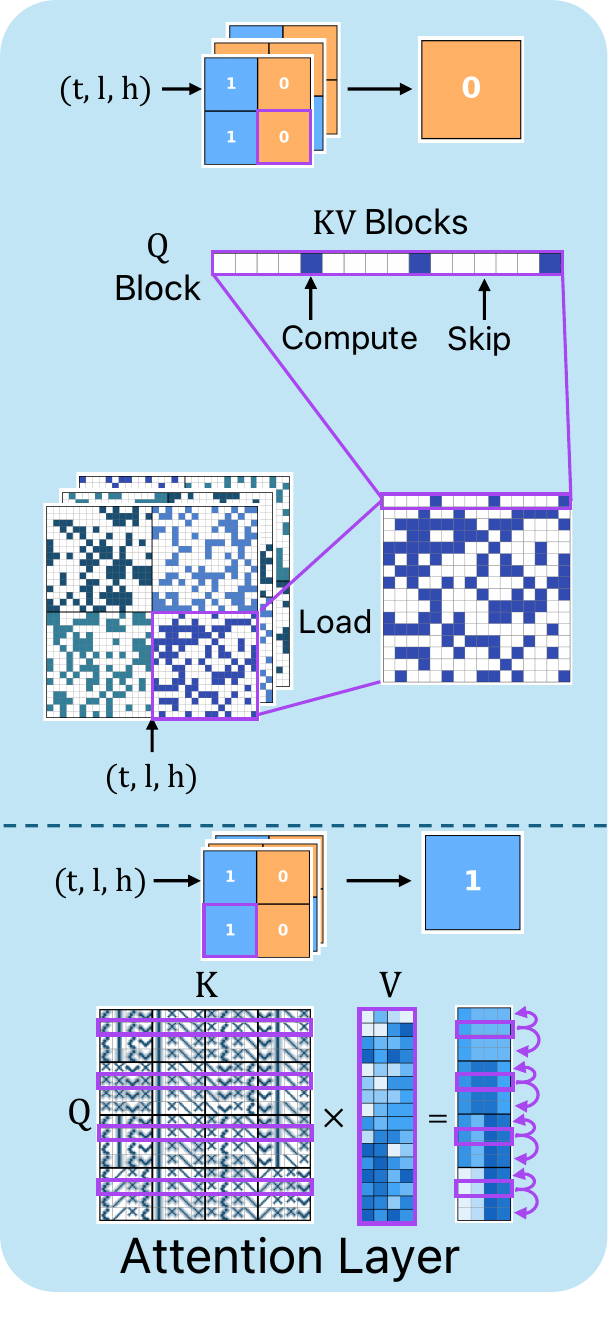}
        \caption{Inference}
        \label{fig:method-infer}
    \end{subfigure}
    \vspace{-5pt}
    \caption{\textbf{Schematic description of \algoname.} \textbf{(a)} We threshold the top key blocks per query block. %
    Then, we aggregate the resulting masks across prompts %
    and store them in a mask dictionary. In addition, we identify attention heads that exhibit spatial row repetition and store them in a dictionary. \textbf{(b)} At inference time, for non-repetitive heads ({top}), we load block masks into memory and skip the computation of the unset blocks accordingly. For heads flagged as spatially repetitive ({bottom}), we compute attention only for selected anchor rows per frame and broadcast the outputs to neighboring rows.}
    \label{fig:method}
     \vspace{-10pt}
\end{figure}

%% file: figures/similarities.tex
\begin{figure}[t]
    \centering
    \begin{subfigure}[t]{0.48\linewidth}
        \centering
        \includegraphics[width=\linewidth]{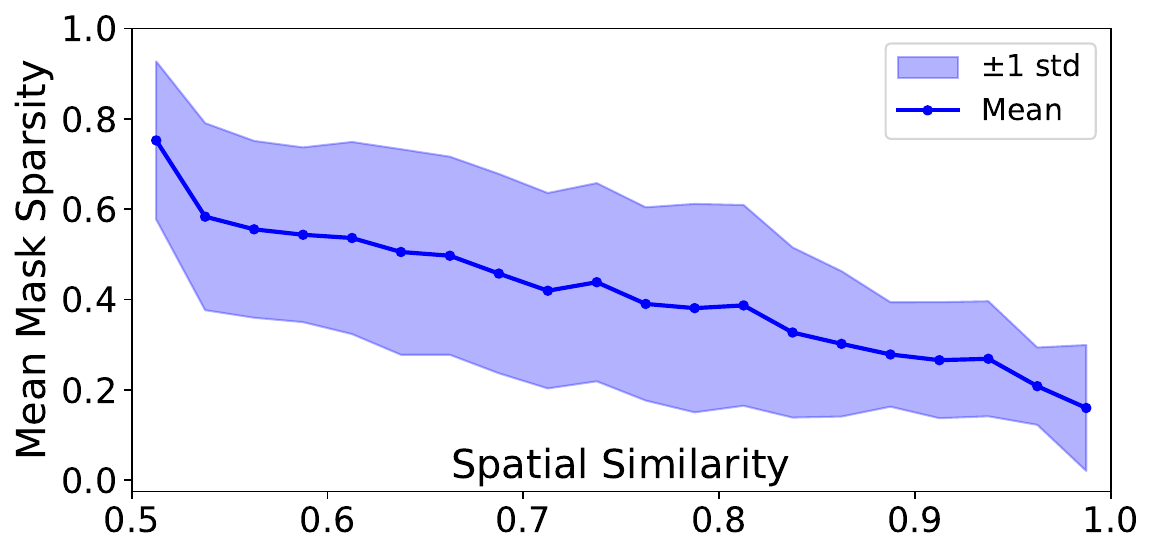}
        \caption{}
        \label{fig:rep_vs_sparsity}
    \end{subfigure}
    \hfill
    \begin{subfigure}[t]{0.48\linewidth}
        \centering
        \includegraphics[width=\linewidth]{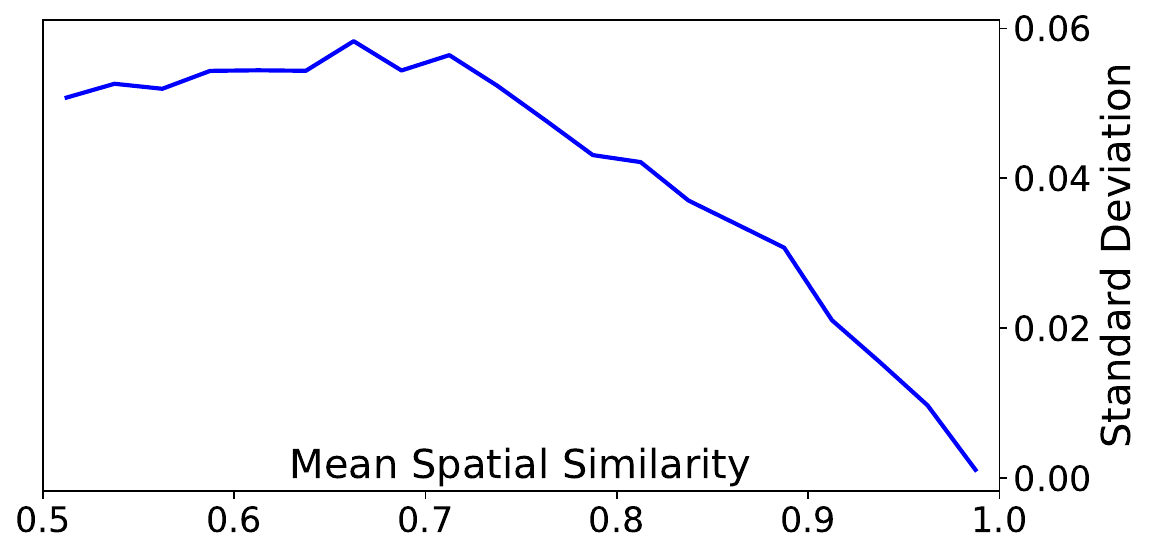}
        \caption{}
        \label{fig:spatial_rep_consistency}
    \end{subfigure}
    \vspace{-6pt}
    \caption{\textbf{Spatial repetition analysis.}
    \textbf{(a) Complementarity with block sparsity}: Spatial similarity vs. calibrated mask sparsity for each attention map. The negative correlation indicates that the two strategies target different maps.
    \textbf{(b) Cross-prompt consistency}: Standard deviation of spatial similarity across calibration prompts. Low variance for high-similarity maps enables their reliable identification from few samples.}
    \label{fig:spatial_rep_analysis}
    \vspace{-10pt}
\end{figure}

%% file: tables/table_wan.tex
\begin{table}[tb]
  \caption{VBench scores, averaged attention sparsity, and end-to-end latency for high-timestep different text-to-video generation models and different resolutions.}
  \label{tab:models_combined}
  \centering
  \begin{tabular}{@{}l@{\hspace{6pt}}l|ccc|ccc@{}}
    \toprule
    \multirow{2}{*}{Model} & \multirow{2}{*}{Method} & \multicolumn{3}{c|}{VBench $\uparrow$} & \multirow{2}{*}{Sparsity $\uparrow$} & \multirow{2}{*}{Latency $\downarrow$} & \multirow{2}{*}{Speedup $\uparrow$} \\
     & & Semantic & Quality & Total & & & \\
    \midrule
      \multirowcell{5}{\textbf{Wan 2.1} \\ \textbf{14B} \\ ${480p}$ \\ $81$ frames}
      & Dense (FA3) & $72.71$ & $82.19$ & $80.29$ & $0\%$ & $363$s & $1.00\times$\\
      \cmidrule(l){2-8}
      & SpargeAttention & $72.87$ & $82.03$ & $80.20$ & $49.5\%$ & $317$s & $1.15\times$\\
      & RadialAttention & $76.69$ & $81.24$ & $80.33$ & $49.0\%$ & $302$s & $1.20\times$\\
      & SVG2 & $67.84$ & $82.27$ & $79.38$ & $53.4\%$ & $291$s & $1.25\times$\\
      & \algoname\ (Ours) & $72.80$ & $82.30$ & $80.40$ & $\textbf{68.1}\%$ & $\textbf{250}$s & $\textbf{1.45}\times$\\
    \midrule
      \multirowcell{5}{\textbf{Wan 2.1} \\ \textbf{14B} \\ ${720p}$ \\ $81$ frames}
      & Dense (FA3) & $71.65$ & $81.27$ & $79.35$ & $0\%$ & $1244$s & $1.00\times$\\
      \cmidrule(l){2-8}
      & SpargeAttention & $71.12$ & $80.78$ & $78.85$ & $49.9\%$ & $930$s & $1.33\times$\\
      & RadialAttention & $72.44$ & $79.33$ & $77.95$ & $54.7\%$ & $936$s & $1.32\times$\\
      & SVG2 & $70.03$ & $81.61$ & $79.30$ & $46.3\%$ & $846$s & $1.47\times$\\
      & \algoname\ (Ours) & $72.81$ & $81.41$ & $79.69$ & $\textbf{62.5}\%$ & $\textbf{785}$s & $\textbf{1.58}\times$\\
    \midrule
      \multirowcell{5}{\textbf{Mochi 1} \\ ${480p}$ \\ $85$ frames}
      & Dense (FA3) & $67.02$ & $76.42$ & $74.54$ & $0\%$ & $188$s & $1.00\times$\\
      \cmidrule(l){2-8}
      & SpargeAttention & $69.03$ & $75.64$ & $74.32$ & $49.6\%$ & $178$s & $1.05\times$\\
      & RadialAttention & $66.16$ & $75.47$ & $73.61$ & $37.8\%$ & $175$s & $1.07\times$\\
      & SVG2 & $65.24$ & $74.89$ & $72.96$ & $55.6\%$ & $237$s & $0.79\times$\\
      & \algoname\ (Ours) & $67.34$ & $76.38$ & $74.57$ & $\textbf{69.1}\%$ & $\textbf{161}$s & $\textbf{1.16}\times$\\
    \bottomrule
  \end{tabular}
\end{table}

%% file: tables/table_wan_distilled.tex
\begin{table}[tb]
  \caption{VBench scores, averaged attention sparsity, and end-to-end latency on $4$-step distilled LightX2V~\cite{lightx2v} at $480p$ and $720p$ resolutions with 81 frames.}
  \label{tab:wan_distilled_combined}
  \centering
  \begin{tabular}{@{}ll|ccc|ccc@{}}
    \toprule
    \multirow{2}{*}{Res.} & \multirow{2}{*}{Method} & \multicolumn{3}{c|}{VBench $\uparrow$} & \multirow{2}{*}{Sparsity $\uparrow$} & \multirow{2}{*}{Latency $\downarrow$} & \multirow{2}{*}{Speedup $\uparrow$} \\
     & & Semantic & Quality & Total & & & \\
    \midrule
    \multirow{5}{*}{$\mathbf{480p}$}
      & Dense (FA3) & $76.06$ & $82.28$ & $81.04$ & $0\%$ & $14.5$s & $1.00\times$\\
      \cmidrule(l){2-8}
      & SpargeAttention & $76.40$ & $82.08$ & $80.94$ & $49.6\%$ & $13.5$s & $1.07\times$\\
      & RadialAttention & $77.02$ & $82.46$ & $81.37$ & $48.4\%$ & $12.7$s & $1.14\times$\\
      & SVG2 & $76.19$ & $82.64$ & $81.35$ & $62.7\%$ & $24.0$s & $0.60\times$\\
      & \algoname\ (Ours) & $77.17$ & $82.29$ & $81.26$ & $\textbf{70.9}\%$ & $\textbf{11.2}$s & $\textbf{1.29}\times$\\
    \midrule
    \multirow{5}{*}{$\mathbf{720p}$}                 
      & Dense (FA3) & $73.17$ & $78.74$ & $77.63$ & $0\%$ & $48.3$s & $1.00\times$\\
      \cmidrule(l){2-8}
      & SpargeAttention & $73.47$ & $78.75$ & $77.69$ & $49.9\%$ & $39.9$s & $1.21\times$\\
      & RadialAttention & $73.24$ & $78.68$ & $77.60$ & $52.5\%$ & $37.0$s & $1.30\times$\\
      & SVG2 & $73.19$ & $78.83$ & $77.70$ & $61.9\%$ & $47.3$s & $1.02\times$\\
      & \algoname\ (Ours) & $74.04$ & $78.80$ & $77.84$ & $\textbf{73.9}\%$ & $\textbf{30.6}$s & $\textbf{1.57}\times$\\
    \bottomrule
  \end{tabular}
  \vspace{-8pt}
\end{table}

%% file: figures/n_prompts_calib.tex
\begin{figure}[t]
    \centering
    \vspace{-7pt}
    \includegraphics[width=0.76\linewidth]{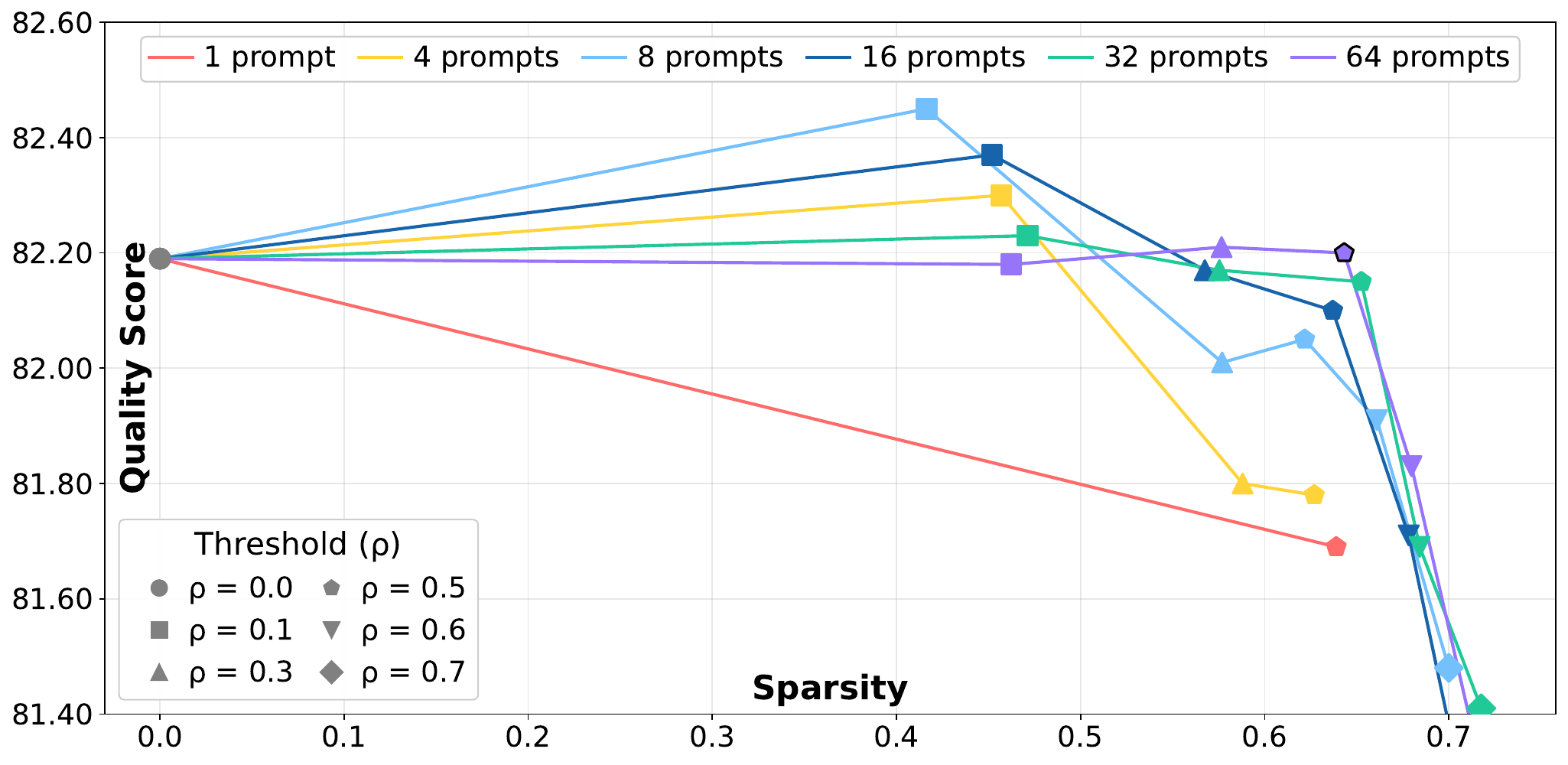}
    \vspace{-9pt}
    \caption{\textbf{Calibration set size and agreement threshold.}
    VBench Quality Score over sparsity for different calibration set sizes and agreement thresholds \(\rho\).
    }
    
    \label{fig:calib_size_rho}
    \vspace{-11pt}
\end{figure}

%% file: appendix_content.tex
\title{Accelerating Video Generation with Calibrated Sparse Attention\\Supplementary Material} 
\titlerunning{\algoname: Supplementary Material}

\renewcommand{\theHsection}{\Alph{section}}
\renewcommand{\theHsubsection}{\theHsection.\arabic{subsection}}
\renewcommand{\thefigure}{S\arabic{figure}}
\renewcommand{\thetable}{S\arabic{table}}
\renewcommand{\theequation}{S\arabic{equation}}

\author{Shai Yehezkel\inst{1,2} \and
Shahar Yadin\inst{1} \and
Noam Elata\inst{1} \and
Yaron Ostrovsky-Berman\inst{1} \and
Bahjat Kawar\inst{1}}

\authorrunning{S.~Yehezkel et al.}

\institute{Apple \and
Tel Aviv University}

\maketitle

\input{sections/sup/implementation}

\input{sections/sup/add_exps}

\input{sections/sup/add_qual}

%% file: sections/sup/implementation.tex
\section{Additional Implementation Details}
\label{sec:supp_impl_details}

\noindent

This section provides additional implementation details for \algoname.
\cref{subsec:supp_bayes_search} describes the hyperparameter search procedure for the energy threshold schedule.
\cref{subsec:skip_list_mem_optimization} details memory optimizations for skip-list storage.
\cref{subsec:timestep_mask_sharing} explains how masks can be shared across similar timesteps to further reduce memory.
\cref{subsec:supp_hparams} provides implementation details for the baseline methods.
When using classifier-free guidance (CFG), we calibrate the sparse masks and spatial similarity scores using only the conditional branch. At inference time, the same calibrated mask for each timestep, layer, and head is applied to both the conditional and unconditional branches.

\subsection{Hyperparameter search for energy threshold schedule}
\label{subsec:supp_bayes_search}

We tune the timestep-dependent energy threshold schedule $\epsilon(t)$ in \cref{eq:epsilon_schedule} using hyperparameter optimization with Optuna~\cite{akiba2019optunanextgenerationhyperparameteroptimization}.
For each model and inference configuration reported in the paper, we run an independent search over the schedule parameters $A$, $C$, and $k$, which control the exponential profile.
Each trial evaluates a candidate schedule by calibrating masks using $|\mathcal{D}|{=}64$ prompts and measuring the resulting VBench score and sparsity on the calibration set.

Across settings, we find that the fitted schedules exhibit simple regularities.
In the low-step (distilled) regime, the optimal parameters are highly consistent across resolutions, and we therefore use the same schedule for both $480p$ and $720p$ distilled LightX2V: $A{=}0.763$, $C{=}0.863$, and $k{=}5.64$.
In the high-step regime, we find that the schedule is dominated by the \emph{base level} $A$, while $C$ and $k$ remain relatively stable.
In practice, we use
\begin{equation}
A(N) \;=\; 0.796 \;+\; \left(1.41 \cdot 10^{-6}\right)\,N,\qquad
C \;=\; 0.99,\qquad
k \;=\; 16,
\end{equation}
where $N$ is the attention sequence length for the given model and configuration.
\cref{fig:keep_ratio_comparison} plots the energy threshold $\epsilon$ as a function of the normalized timestep $t/T$ for different models and resolutions.
\input{figures/keep_ratio}

\subsection{Skip-List Memory Optimization}
\label{subsec:skip_list_mem_optimization}
At inference, each calibrated block mask $\rmM^{(t,l,h)}$ is converted into a skip-list representation
that specifies, at block granularity, which attention blocks should be computed.
\input{figures/row_length_freq}
\noindent
For a sequence of length $N$ and block size $B$, each mask is stored as a matrix of size
$N_B \times N_B + 1$.
Each row corresponds to a query block-row, while the columns encode a variable-length list of
contiguous intervals along the key dimension.
The additional column stores the number of valid intervals for each query block-row.
In the worst case, an alternating compute--skip pattern would require storing $N_B/2$ single-block
intervals per row (each represented by start/end indices), forcing the skip-list matrix to retain its
full width (Fig.~\ref{fig:skip_list_full}).
Although such patterns are theoretically possible, we do not observe them in practice
(Fig.~\ref{fig:row_length_freq}).
\\
Motivated by this observation, we implement skip-list matrices with mask-dependent column lengths in
our kernel.
Concretely, instead of allocating a fixed-width matrix of size $N_B$ columns for every mask, we
trim each skip-list matrix to the maximum interval index populated by any query block-row
(Fig.~\ref{fig:skip_list_trimmed}), preserving a dense layout shared across rows while significantly
reducing padding.
For Wan-T2V 14B at $720p$, the full set of untrimmed skip-list matrices totals $52$\,GB.
Per-layer trimming alone reduces this to $21.5$\,GB with no effect on correctness or sparsity which we use for the $720p$ results reported in Tab.~\ref{tab:models_combined}.
\\
Tab.~\ref{tab:skip_list_memory} further reports the effect of progressively merging nearby intervals,
which intentionally marks a small number of additional blocks for computation in exchange for reduced
padding (Fig.~\ref{fig:skip_list_merged}).
In particular, interval merging reduces the footprint to $6.3$\,GB with neglible sparsity loss.
\\
Finally, a more compact alternative stores all intervals in a flat one-dimensional array and
maintains a per-row offset and length for each query block-row, eliminating padding entirely.
Tab.~\ref{tab:skip_list_memory} shows that this representation alone reduces the footprint to
$4.0$\,GB without any merging. The same interval merging strategy applies, yielding identical merged intervals for each percentile. We leave the efficient kernel implementation of the 1D representation to future work and report only VBench scores and sparsity levels.

\input{figures/skip_list_viz}
\input{tables/skip_list_mem}

\input{figures/t_mask_sim}
\subsection{Timestep Mask Sharing}
\label{subsec:timestep_mask_sharing}

We observe that calibrated masks become increasingly similar at later timesteps. To quantify this, we compute pairwise Intersection over Union (IoU):
\begin{equation}
\label{eq:timestep_iou}
\mathrm{IoU}(t_1,t_2)
\;=\;
\frac{\left| S^{(t_1)} \cap S^{(t_2)} \right|}{\left| S^{(t_1)} \cup S^{(t_2)} \right|},
\end{equation}
where $S^{(t)}$ denotes the set of skipped blocks at timestep $t$, averaged over layers and heads.
As shown in \cref{fig:timestep_similarity_heatmap}, later timesteps exhibit high pairwise IoU.
This suggests that masks can be shared across timesteps, reducing memory footprint without affecting inference.
For each (layer, head) pair, we greedily cluster timesteps into cliques: a timestep joins a cluster only if its IoU with \emph{all} existing cluster members exceeds a threshold $\tau$. All timesteps in a cluster share a single mask, obtained by taking the logical OR over the \emph{kept-block} masks of all cluster members.
Combined with the skip-list optimizations described in \cref{subsec:skip_list_mem_optimization}, timestep sharing further reduces the inference-time footprint to $3.6$GB for 2D skip lists and $2.2$GB for 1D skip lists. \cref{tab:skip_list_memory} reports results for $\tau \in \{0.97, 0.98\}$, showing the trade-off between memory reduction and sparsity loss.

\subsection{Baselines Implementation Details}
\label{subsec:supp_hparams}
\noindent
For RadialAttention~\cite{liradial}, we use the authors' official implementation.
We apply the default configuration reported for Wan2.1-14B in the repository for both $480p$ and $720p$ settings, including a FlashAttention block size of $128{\times}128$.
For the dense/warmup portions of RadialAttention, we set the attention backend to FlashAttention3 (FA3) to match our dense baseline.

\noindent
For Sparse VideoGen2 (SVG2)~\cite{yang2025sparse}, we use the authors' official implementation.
For evaluation at $480p$, we adopt the hyperparameters provided in the official implementation for Wan2.1 14B Image-to-Video, and for $720p$ Text-to-Video we use the hyperparameters reported in the SVG2 paper for Wan2.1 14B.
SVG2 uses customized sparse attention kernels built on FlashInfer~\cite{ye2025flashinferefficientcustomizableattention} supporting FA3, including support for varied block sizes.
For the dense/warmup portions of SVG2, we set the attention backend to FA3 to match our dense baseline.
SVG2 performs k-means clustering on query and key tokens at each attention layer to enable semantic-aware permutation. While a centroid caching mechanism reduces this overhead across denoising steps~\cite{yang2025sparse}, an initialization cost remains at the start of inference. In distilled models with a small number of sampling timesteps (\cref{tab:wan_distilled_combined}), this initialization overhead constitutes a larger fraction of the total runtime, limiting the effective speedup. For other models, the per-step clustering overhead alongside initialization may outweigh the speedup gained, as observed with Mochi (\cref{tab:models_combined}).

\noindent
For SpargeAttention~\cite{zhang2025spargeattn}, we use the authors' official implementation.
Following the repository recommendation, we use the SageAttention2~\cite{zhang2025sageattention2efficientattentionthorough} backend for sparse attention.

%% file: figures/keep_ratio.tex
\begin{figure}[t]                                        
    \centering                                           
        \includegraphics[width=0.8\linewidth]{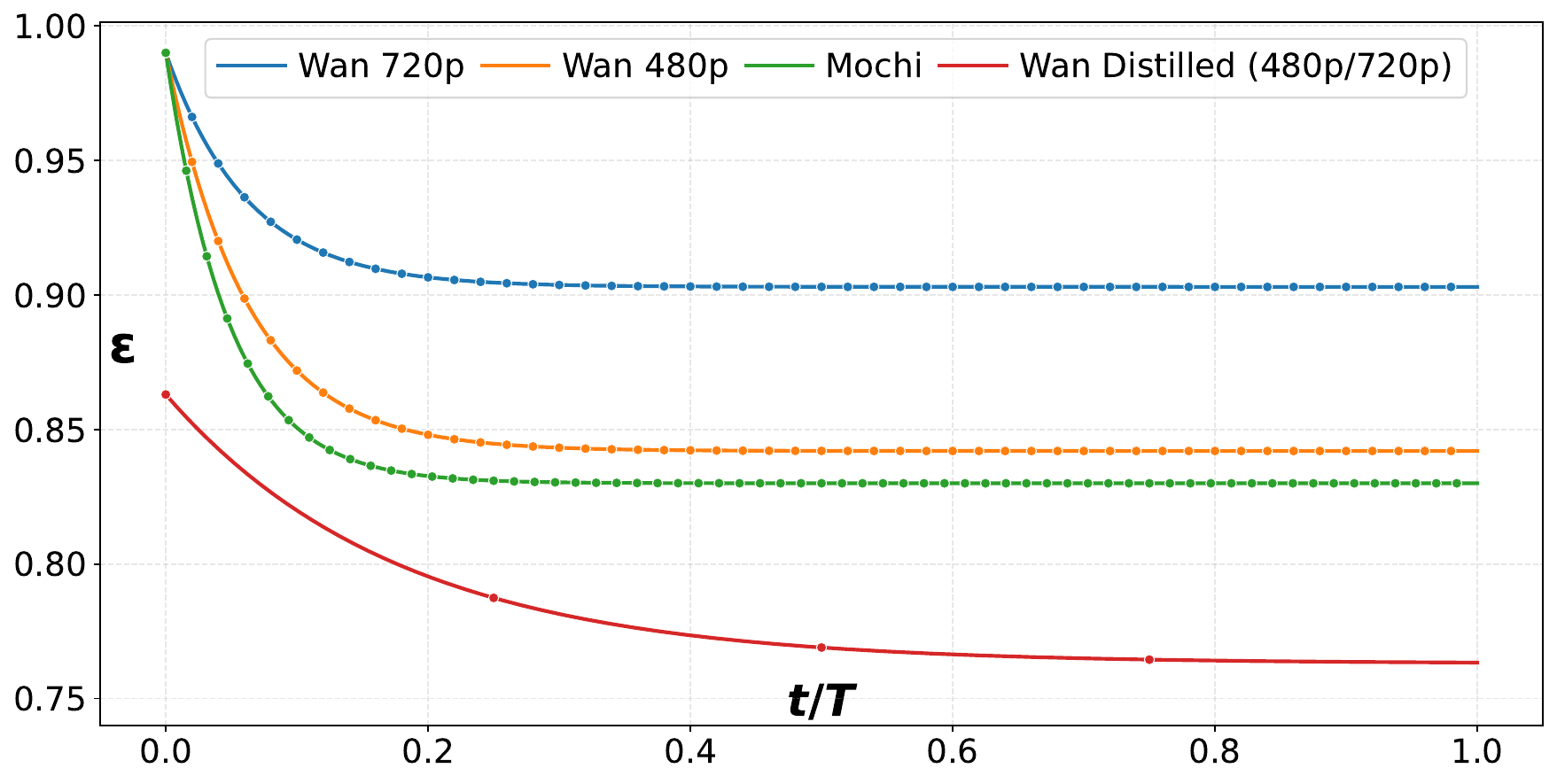}                                         
    \caption{\textbf{Energy threshold schedule.} $\epsilon$ as a function of the normalized timestep $t/T$ for different models and resolutions.}
    \vspace{-5pt}
    \label{fig:keep_ratio_comparison}
\end{figure}

%% file: figures/row_length_freq.tex
\begin{wrapfigure}[16]{r}{0.45\linewidth}
    \vspace{-20pt}
    \centering
    \includegraphics[width=\linewidth]{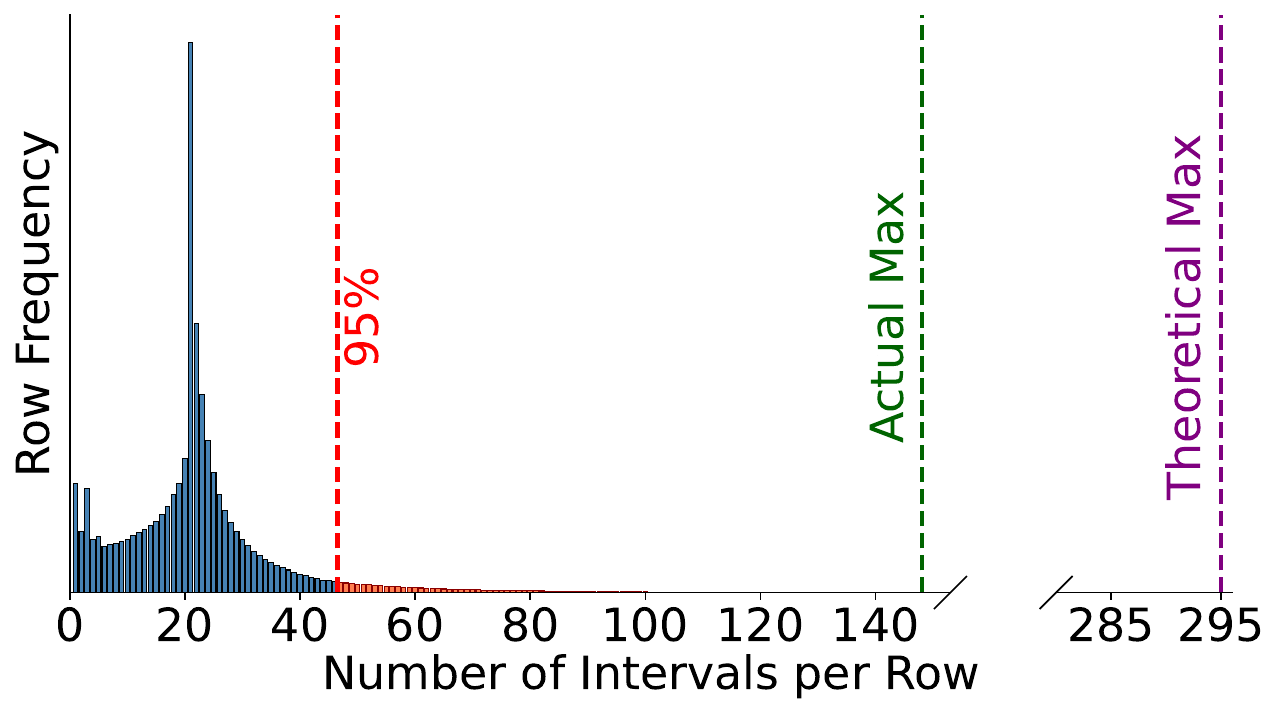}
    \vspace{-20pt}
    \caption{\textbf{Skip-list row length distribution.}
    Histogram of the number of intervals per skip-list row, measured on skip lists calibrated for \textbf{720p} generation on Wan2.1-14B text-to-video with $N{=}75{,}600$.
    \textbf{95\%} of query rows require at most $\leq 50$ intervals, out of a theoretical maximum of $295$.}
    \label{fig:row_length_freq}
\end{wrapfigure}

%% file: figures/skip_list_viz.tex
\begin{figure}[t]
  \centering
  \begin{subfigure}[t]{0.32\textwidth}
      \centering
      \caption{Full allocation}
      \label{fig:skip_list_full}
      \vspace{2pt}
      \includegraphics[width=\textwidth]{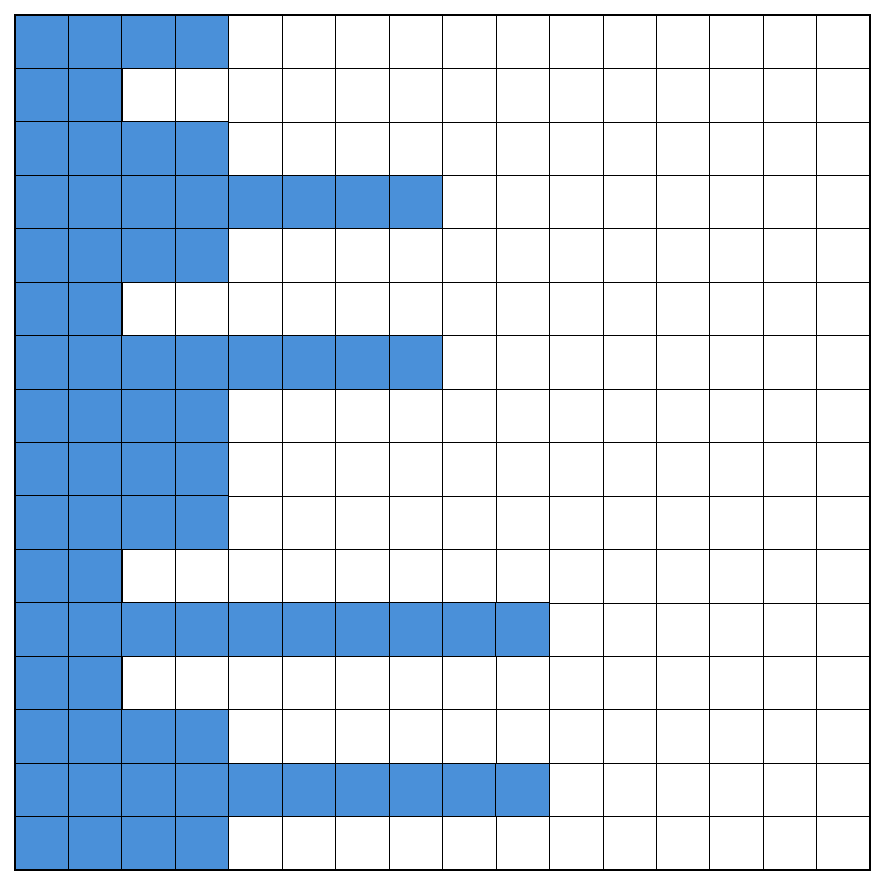}
  \end{subfigure}
  \hfill
  \begin{subfigure}[t]{0.32\textwidth}
      \centering
      \caption{Trimmed}
      \label{fig:skip_list_trimmed}
      \vspace{2pt}
      \includegraphics[width=\textwidth]{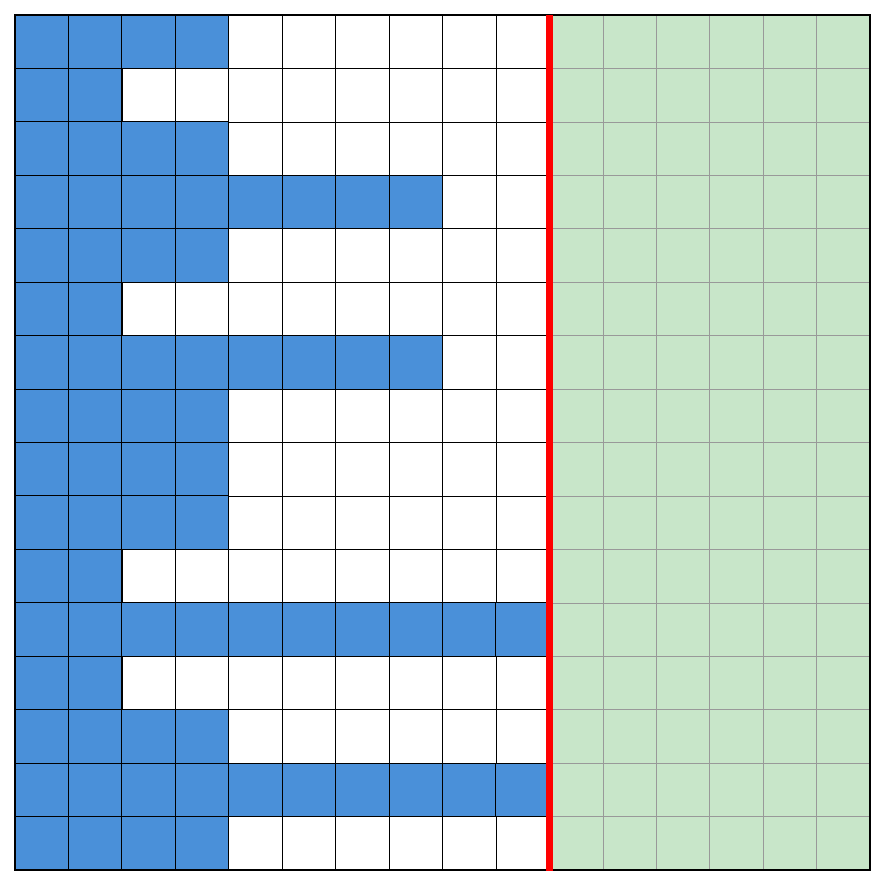}
  \end{subfigure}
  \hfill
  \begin{subfigure}[t]{0.32\textwidth}
      \centering
      \caption{Interval merging}
      \label{fig:skip_list_merged}
      \vspace{2pt}
      \includegraphics[width=\textwidth]{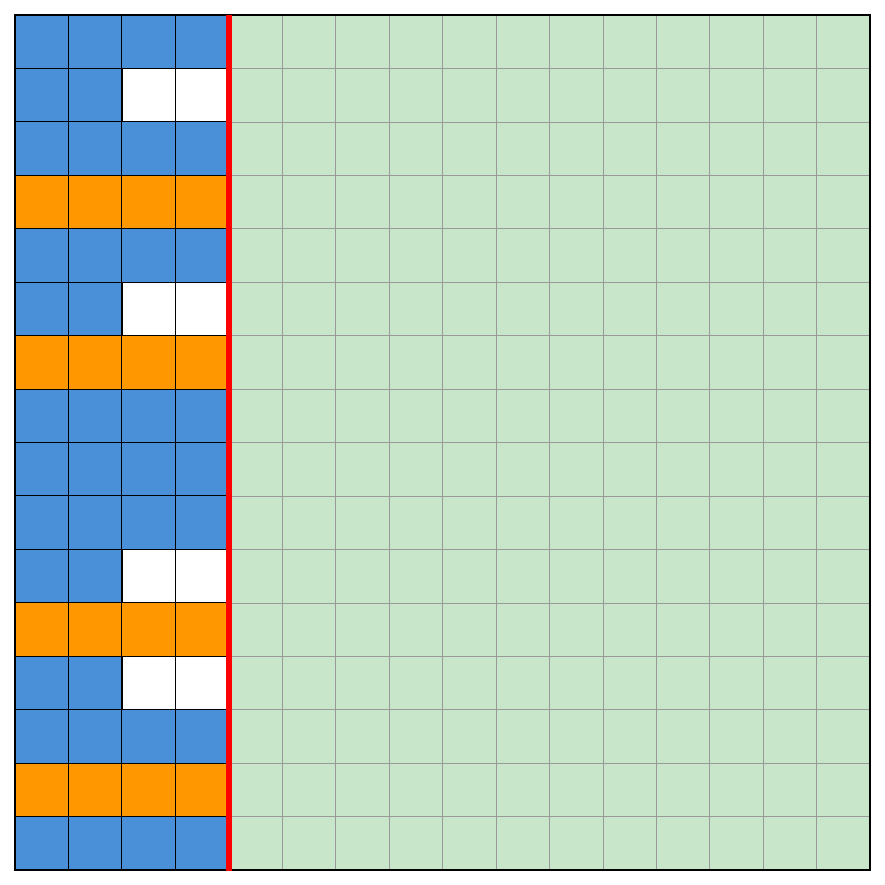}
  \end{subfigure}

  \vspace{2pt}
  \includegraphics[width=0.7\textwidth]{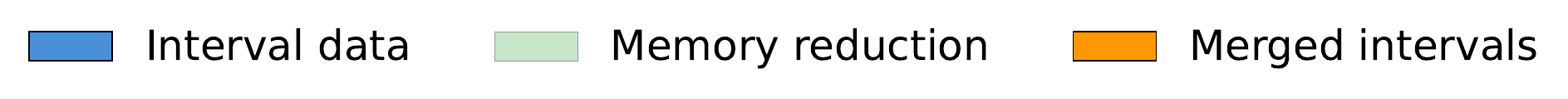}

  \caption{\textbf{Skip-list memory optimization}. Each row stores interval pairs
  for a (timestep, head, query-block) entry within a layer; columns beyond the data are padding. 
  \textbf{(a)}~Full allocation to theoretical maximum width.
  \textbf{(b)}~Per-layer trimming to the widest row used across all entries in the layer (red line).
  \textbf{(c)}~Merging outlier rows (orange) reduces the layer's maximum width, enabling further trimming.}
  \label{fig:skip_list_optimization}
  \vspace{-10pt}
\end{figure}

%% file: tables/skip_list_mem.tex
\begin{table}[t]
  \centering
  \caption{Skip-list memory footprint for Wan-T2V 14B at 720p. Reductions are relative to the per-layer trimmed 2D baseline (21.5\,GB). Timestep sharing clusters masks with pair-wise IoU $\geq \tau$.}
  \label{tab:skip_list_memory}
  \setlength{\tabcolsep}{3pt}
  \renewcommand{\arraystretch}{1.08}
  \begin{tabular}{@{}lc|cc|ccc|ccc@{}}
    \toprule
    \multirow{2}{*}{Merge \%} & \multirow{2}{*}{$\tau$} & Size & Red. & \multicolumn{3}{c|}{VBench $\uparrow$} & \multirow{2}{*}{Sparsity $\uparrow$} & Latency & Speedup \\
    & & (GB) $\downarrow$ & $\uparrow$ & Qual. & Sem. & Total & & $\downarrow$ & $\uparrow$ \\
    \midrule
    Dense (FA3) & -- & -- & -- & $81.27$ & $71.65$ & $79.35$ & $0\%$ & $1244$s & $1.00\times$ \\
    \midrule
    \multicolumn{10}{c}{\textbf{2D skip lists}} \\
    \midrule
    100 (no merge) & --   & $21.5$ & $0\%$  & $81.41$ & $72.81$ & $79.69$ & $62.50\%$ & $785$s & $1.58\times$ \\
    99             & --   & $11.8$ & $45\%$ & $81.41$ & $72.81$ & $79.69$ & $62.50\%$ & $785$s & $1.58\times$ \\
    95             & --   & $8.1$  & $63\%$ & $81.38$ & $72.76$ & $79.65$ & $62.41\%$ & $786$s & $1.58\times$ \\
    90             & --   & $6.3$  & $71\%$ & $81.35$ & $72.86$ & $79.65$ & $62.23\%$ & $788$s & $1.57\times$ \\
    90             & 0.98 & $4.7$  & $78\%$ & $81.25$ & $72.55$ & $79.51$ & $62.00\%$ & $790$s & $1.57\times$ \\
    90             & 0.97 & $3.6$  & $83\%$ & $81.40$ & $72.77$ & $79.67$ & $61.74\%$ & $793$s & $1.56\times$ \\
    \midrule
    \multicolumn{10}{c}{\textbf{1D skip lists}} \\
    \midrule
100 (no merge) & --   & $4.0$ & $81\%$ & $81.41$ & $72.81$ & $79.69$ & $62.50\%$ & -- & -- \\
    95             & --   & $3.9$ & $82\%$ & $81.38$ & $72.76$ & $79.65$ & $62.41\%$ & -- & -- \\
    90             & --   & $3.8$ & $82\%$ & $81.35$ & $72.86$ & $79.65$ & $62.23\%$ & -- & -- \\
    90             & 0.98 & $2.8$ & $87\%$ & $81.25$ & $72.55$ & $79.51$ & $62.00\%$ & -- & -- \\
    90             & 0.97 & $2.2$ & $90\%$ & $81.40$ & $72.77$ & $79.67$ & $61.74\%$ & -- & -- \\
    \bottomrule
  \end{tabular}
\end{table}

%% file: figures/t_mask_sim.tex
\begin{figure}[t]
    \centering
    \vspace{-4pt}
    \includegraphics[width=0.67\linewidth]{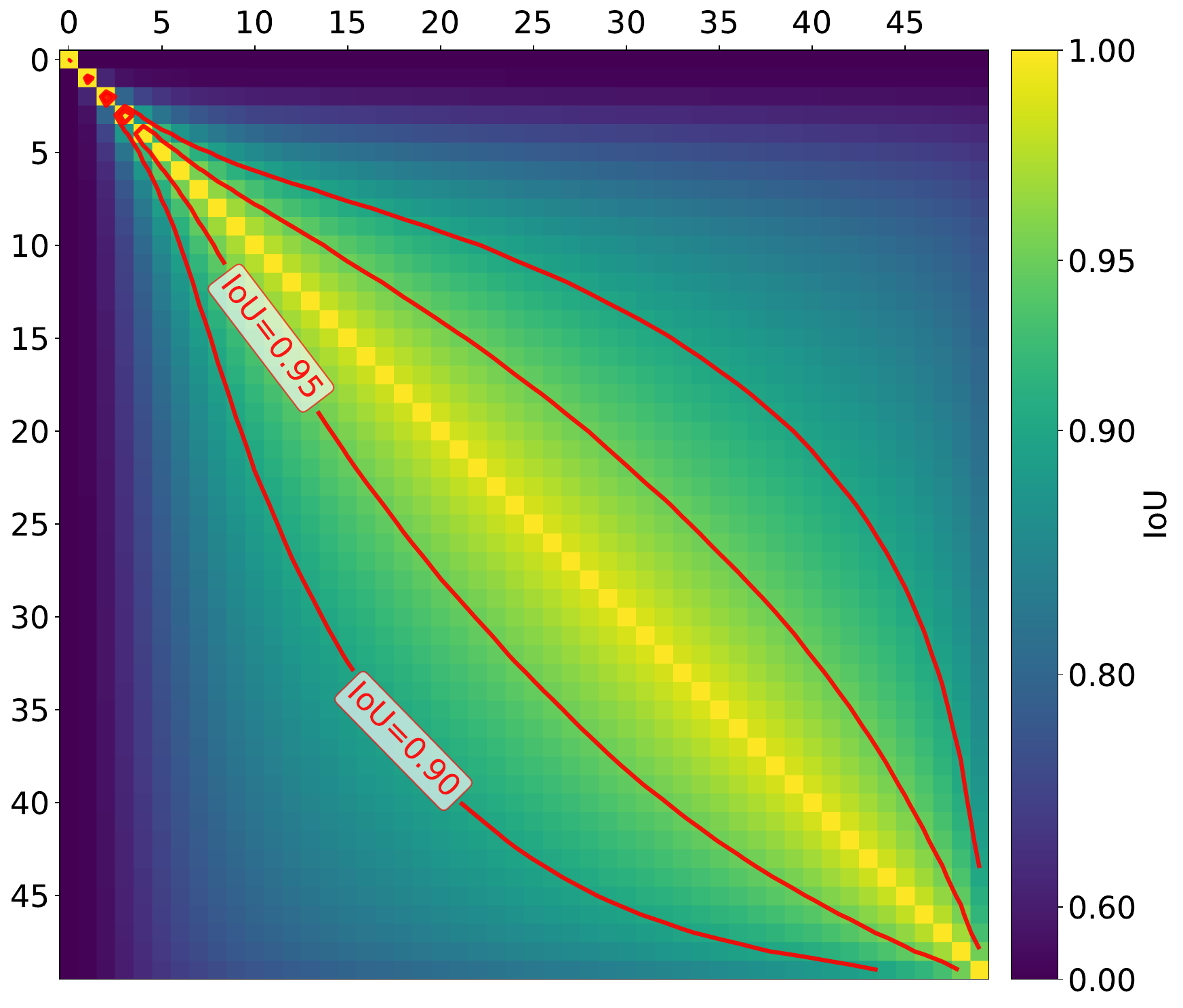}
    \vspace{-8pt}
    \caption{\textbf{Timestep similarity of calibrated masks.}
    Each cell $(t_1,t_2)$ reports the intersection-over-union (IoU) between the \emph{skipped-block} masks at timesteps $t_1$ and $t_2$, averaged over all layers $l$ and heads $h$.
    Later timesteps show higher cross-timestep similarity (brighter off-diagonal region), suggesting that masks can be shared across consecutive late denoising steps, reducing the memory footprint of storing per-timestep masks.
    Red contour lines mark constant-similarity levels.
}
    \label{fig:timestep_similarity_heatmap}
    \vspace{-10pt}
\end{figure}

%% file: sections/sup/add_exps.tex
\section{Additional Experiments}
\label{sec:supp_ablations}

\noindent
This section presents additional experiments and ablations.
\cref{subsec:anchor_rows} analyzes the spatial repetition mechanism, studying the effect of anchor row count and similarity threshold on speedup and approximation quality.
\cref{subsec:supp_calib_cost} reports the compute cost of the calibration stage and explores trade-offs between calibration budget and output quality.
\cref{subsec:supp_block_size} ablates the effect of FlashAttention block size on achievable sparsity and end-to-end speedup.

\input{figures/anchor_speedup}
\subsection{Spatial Repetition Anchor Rows}                                           
\label{subsec:anchor_rows}                                                             
  We study the effect of the number of anchor rows $k$ and the                          
  similarity threshold $\gamma$ on the spatial repetition mechanism
  introduced in \cref{subsec:spatial-rep}, evaluated on
  Wan~2.1 14B T2V at 480p resolution. Each generated video
  consists of 21 frames with $1{,}560$ spatial tokens per frame
  ($H{=}30$ spatial rows $\times$ $W{=}52$ spatial columns), yielding a total
  sequence length of $32{,}760$ tokens.
\\
  We first benchmark the raw kernel-level speedup of the anchor row
  attention against full Flash Attention~3
  (\cref{tab:anchor_bench}) corresponding to selecting all rows as anchors ($k{=}30$). By using a single anchor row kernel achieves a $20.5\times$ speedup by reducing
  the query count from $HW{=}1{,}560$ to $W{=}52$ tokens per
  frame.
\input{tables/table_anchor_speedup}
\noindent
To evaluate the quality--speedup trade-off, we run inference on
our calibration set of 64 prompts using different combinations of
$\gamma$ and $k$. For each configuration, a different number of
attention heads are marked as spatially repetitive: lower
thresholds $\gamma$ mark more heads, while higher thresholds are
more conservative. For each head at every $(t,l,h)$, we compute
the relative error between the sparse and full attention outputs
as $\lVert \mathbf{A}_{\text{sparse}} -
\mathbf{A} \rVert_2 \,/\,
\lVert \mathbf{A} \rVert_2$, where $\mathbf{A}$ is defined in \cref{eq:pv}, then average over all
$(t,l,h)$ and all prompts. We report the overall sparsity induced
by spatial repetition at each $(\gamma, k)$ configuration in \cref{fig:sparsity_tradeoff} and the effective attention speedup, weighted by the fraction of heads marked repetitive at each $\gamma$ in \cref{fig:speedup_tradeoff}.
As shown,
using multiple anchor rows at a lower threshold almost always
achieves the same or lower relative error compared to using a
  single anchor row at a more conservative threshold while
  attaining higher sparsity and speedup. This suggests that
  increasing spatial coverage through additional anchors is more
  effective than restricting the set of repetitive heads. Based on
  these results, we set $\gamma{=}0.87$ and $k{=}5$ in all
  experiments. 
  
  \noindent
  To visualize where repetitive heads occur, \cref{fig:similarity_heatmap_tau_0p87} plots the percentage of heads exceeding $\tau{=}0.87$ per timestep and layer. Repetitive heads cluster in the first and last layers throughout the denoising trajectory, while the earliest timesteps exhibit slightly higher repetition overall.

\input{figures/repetitive_heads}

\subsection{Calibration Stage Compute Cost}
\label{subsec:supp_calib_cost}

The calibration stage consists of two main components: computing block energies for mask calibration (\cref{sec:mask_calibration}), for which we implemented a custom CUDA kernel, and computing spatial similarity scores for repetition detection (\cref{subsec:spatial-rep}), which uses an optimized PyTorch implementation. Both computations run over a set of calibration prompts, but can use different numbers of prompts independently.
As shown in \cref{fig:spatial_rep_consistency}, the standard deviation of spatial similarity scores across prompts is low for high-similarity maps, suggesting that fewer prompts suffice for reliable repetition detection.
Similarly, \cref{fig:calib_size_rho} shows that for our selected threshold $\rho = 0.5$, quality remains stable across different calibration set sizes beyond a minimal number of prompts, motivating the use of fewer prompts for mask calibration as well.
We ablate on different numbers of prompts for each component in \cref{tab:calib_budget} for Wan 2.1 14B at $720p$ with $50$ timesteps, reporting the total calibration cost in H100 GPU-hours. Since similarity computation has lower cross-prompt variance, we can use as few as 1 prompt for similarity while using more prompts for mask calibration. This allows trading off calibration time against quality, enabling faster calibration under restricted compute budgets.
In particular, using 16 prompts for mask calibration and just 1 prompt for similarity computation reduces calibration time to 13.7 GPU-hours while maintaining quality and semantic scores comparable to the dense FA3 baseline.
The calibration cost could be reduced further through a more efficient implementation of the similarity computation, or by fusing the similarity and block energy computations into a single kernel. We leave these optimizations for future work.

\input{tables/table_calibs}

\vspace{-15pt}
\subsection{Kernel Block Size}
\label{subsec:supp_block_size}
\vspace{-5pt}

\input{figures/kernel_size_sparse}
FlashAttention~\cite{dao2022flashattention} partitions the attention computation into blocks of size $B_q \times B_{kv}$, where the optimal sizes depend on the head dimension $d$ and available shared memory (SHMEM) on the GPU. For a given head dimension, larger block sizes improve hardware utilization up to the SHMEM limit. For $d=128$ as in Wan~\cite{wan2025}, FlashAttention uses $128 \times 176$ as its native configuration on H100 GPUs.
Our energy-based block selection in \cref{eq:row_energy_constraint} is largely invariant to block size since the energy threshold $\epsilon$ specifies the fraction of attention mass to retain and the same spatial regions are selected regardless of how they are partitioned into blocks.
However, larger block sizes can reduce achievable sparsity. When a block is selected, all its entries are computed, even if only a subset contributes significant attention mass. Thus, coarser blocks may include more redundant computation, or fewer blocks may be marked as skippable. We illustrate this effect in \cref{fig:block_size_ablation}.
This reveals a tradeoff where larger blocks are faster per block but yield lower sparsity, while smaller blocks achieve higher sparsity but with reduced per-block efficiency. We evaluate this in \cref{tab:block_size_ablation} on the VBench evaluation suite using block sizes supported by FlashAttention, using the same calibrated $\epsilon$ schedule throughout and reporting mask sparsity only (excluding spatial repetition from \ref{subsec:spatial-rep}) to isolate the effect, with dense FA3 included as reference. As expected, smaller blocks yield higher sparsity. However, the improved hardware utilization of larger blocks compensates for the reduced sparsity, and these two effects balance out across a wide range of configurations. From $128 \times 96$ onward, all block sizes achieve comparable speedups of $1.41$--$1.42\times$ while maintaining comparable VBench scores, demonstrating that our method is robust to block size choice and compatible with the native FA3 configuration.

\input{tables/kernel_size}

%% file: figures/anchor_speedup.tex
\begin{figure}[t]                                                                     
      \centering                                                                        
      \begin{subfigure}[t]{0.48\linewidth}                                              
        \centering
        \caption{Sparsity vs.\ relative error.}
        \label{fig:sparsity_tradeoff}
        \includegraphics[width=\linewidth]{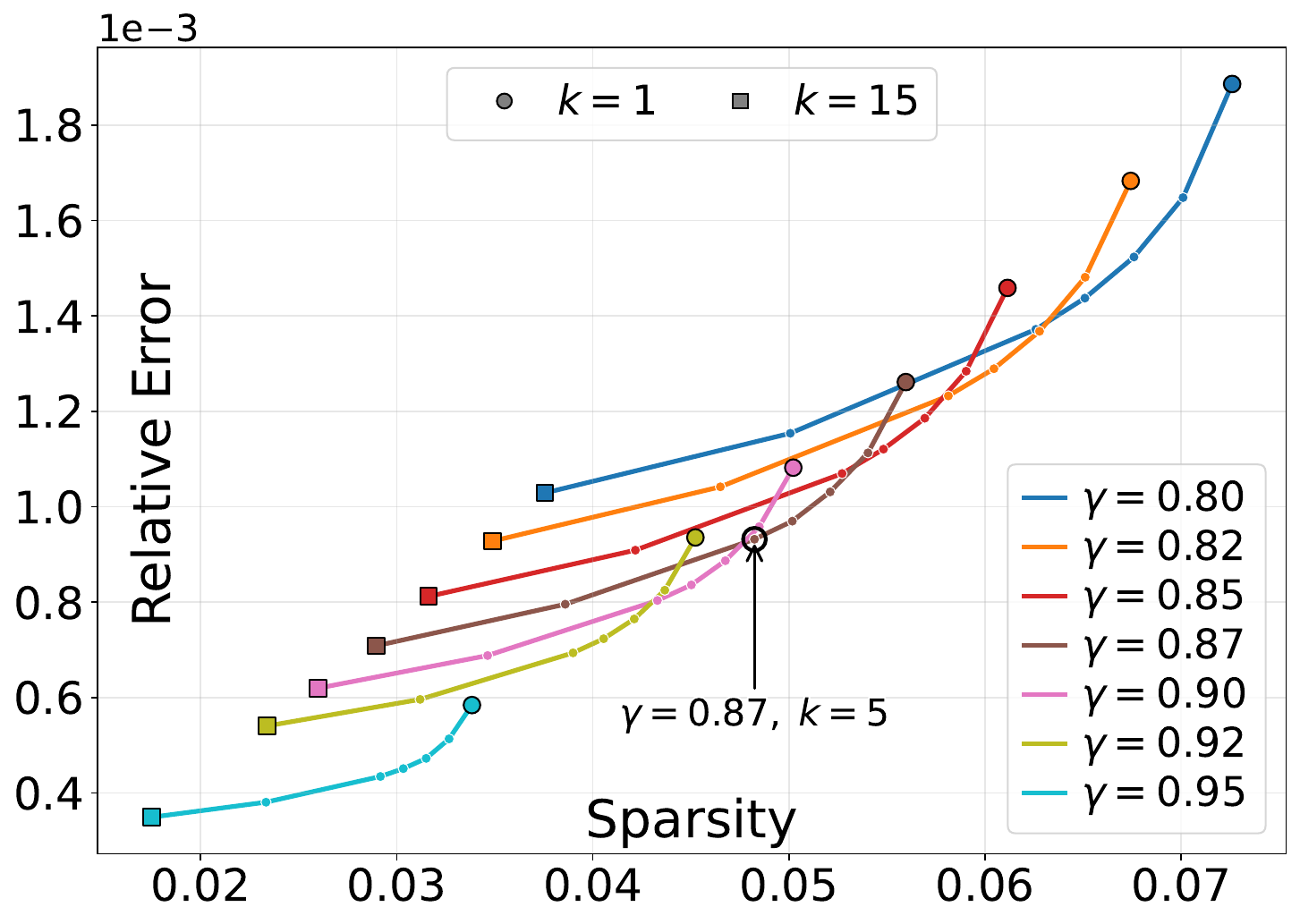}
      \end{subfigure}
      \hfill
      \begin{subfigure}[t]{0.48\linewidth}
        \centering
        \caption{Attention speedup vs.\ relative error.}
        \label{fig:speedup_tradeoff}
        \includegraphics[width=\linewidth]{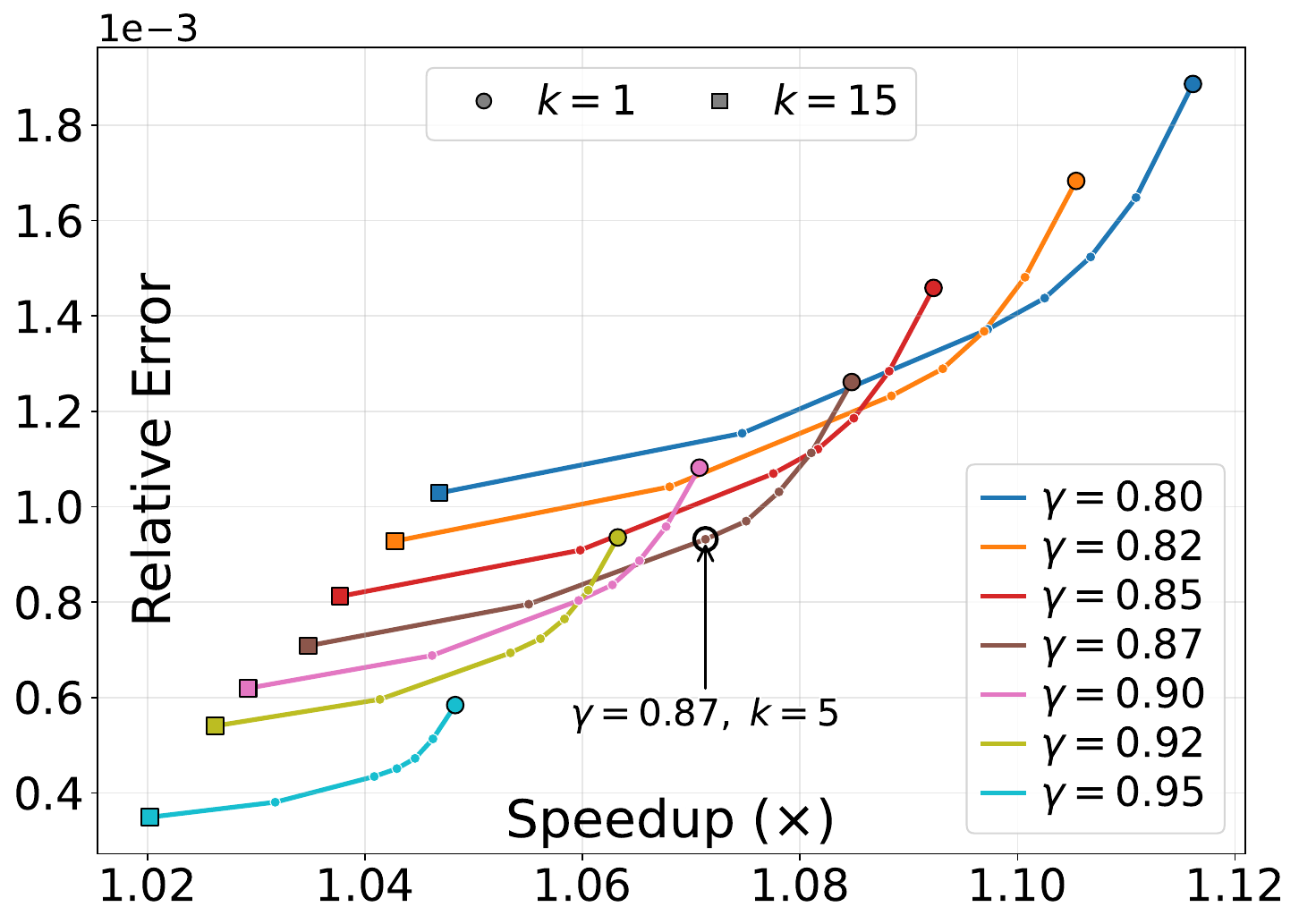}
      \end{subfigure}
      \vspace{-5pt}
      \caption{Trade-off between anchor row count $k$ and approximation
        quality for different similarity thresholds $\gamma$.
        Each curve sweeps $k$ from $1$ to $15$; circle and square markers highlight the $k{=}1$ and $k{=}15$ endpoints.
        (a)~Sparsity vs.\ relative error
        (lower-right is better). (b)~Weighted attention speedup vs.\ relative
        error, accounting for the fraction of repetitive heads at each
        $\gamma$ (lower-right is better).}
        \vspace{-12pt}
      \label{fig:anchor_tradeoff}
    \end{figure}

%% file: tables/table_anchor_speedup.tex
\begin{wraptable}[19]{r}{0.48\linewidth}
    \vspace{-25pt}
    \centering
    \caption{Spatial repetition attention benchmark. Speedup of                         
      computing attention over $k$ anchor query rows versus full                        
      attention, measured with FA3                    
      on a single H100 GPU (batch~1, 40 heads, head dim~128,                            
      sequence length~32{,}760). Sparsity is the fraction of
      query rows skipped per attention map.}
    \label{tab:anchor_bench}
    \small
    \setlength{\tabcolsep}{4pt}
    \begin{tabular}{@{}l|c|cc@{}}
      \toprule
      $k$ & ms $\downarrow$ & Speedup $\uparrow$ & Sps. $\uparrow$ \\
      \midrule
      30 (baseline) & $31.62$ & $1.0\times$ & $0\%$ \\
      \midrule
      15 & $18.60$ & $1.7\times$ & $50.0\%$ \\
      10 & $11.52$ & $2.8\times$ & $66.7\%$ \\
      5  & $5.99$  & $5.3\times$ & $83.3\%$ \\
      4  & $4.79$  & $6.6\times$ & $86.7\%$ \\
      3  & $3.72$  & $8.5\times$ & $90.0\%$ \\
      2  & $2.77$  & $11.4\times$ & $93.3\%$ \\
      1  & $1.54$  & $20.5\times$ & $96.7\%$ \\
      \bottomrule
    \end{tabular}
    \vspace{-5pt}
\end{wraptable}

%% file: figures/repetitive_heads.tex
\begin{figure}[t]
    \centering
    \includegraphics[width=0.6\linewidth]{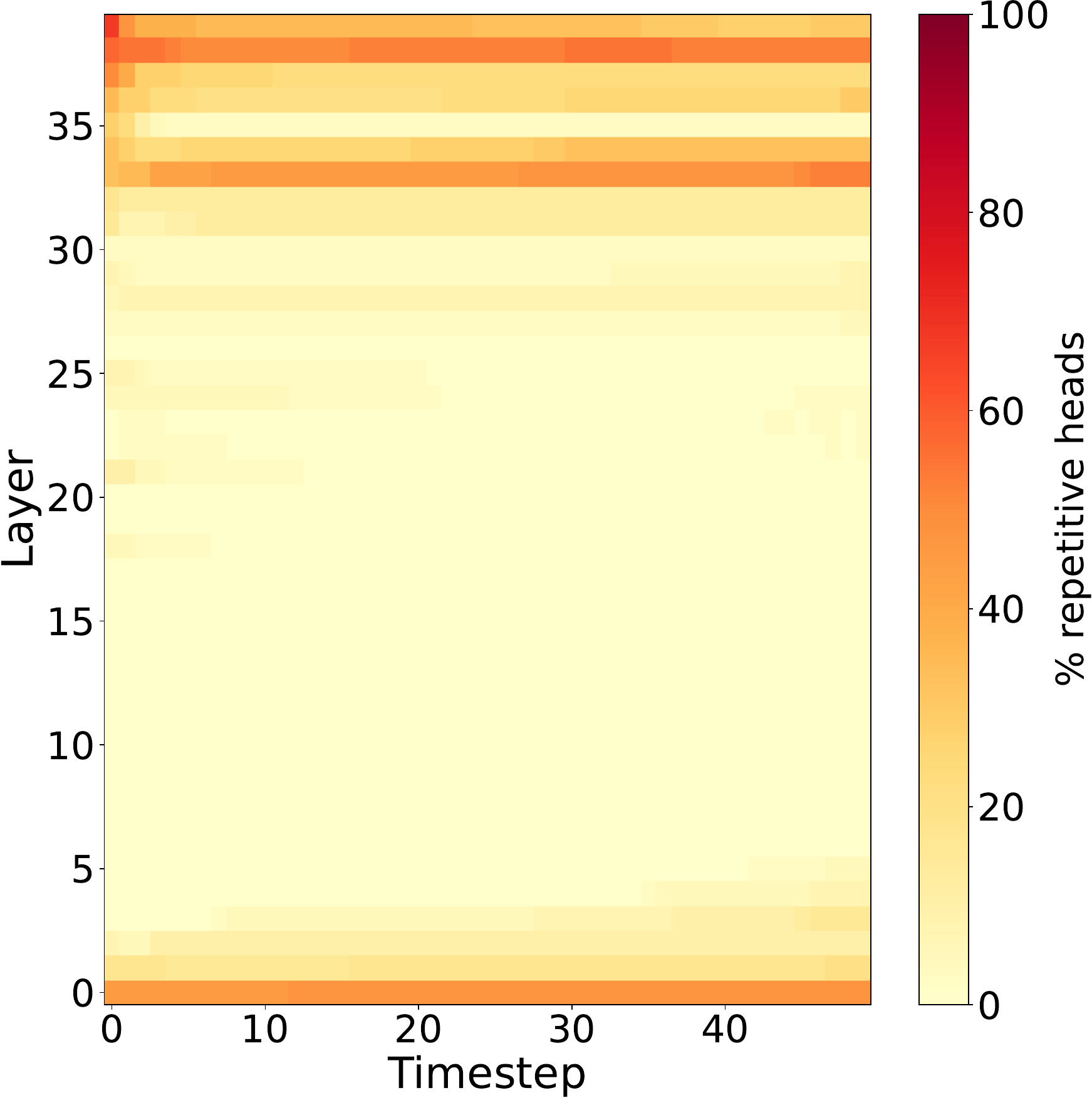}
    \caption{\textbf{Spatial repetition heads count at $\tau=0.87$.} Percentage of attention heads whose spatial-similarity score exceeds $\tau$, per timestep and layer for Wan~2.1 14B 480p.}
    \label{fig:similarity_heatmap_tau_0p87}
    \vspace{-10pt}
\end{figure}

%% file: tables/table_calibs.tex
\begin{table}[t]
  \centering
  \vspace{-1em}
  \caption{Impact of calibration budget on VBench scores for Wan 2.1 14B at 720p.
  Block energies are computed on all calibration prompts; spatial similarity uses a subset.
  GPU-hours on a single H100. Dense FA3 as reference.}
  \label{tab:calib_budget}
  \vspace{-8pt}
  \begin{tabular}{@{}cc|ccc|ccc|c@{}}
  \toprule
  Mask. & Sim. & \multirow{2}{*}{Quality $\uparrow$} & \multirow{2}{*}{Semantic $\uparrow$} & \multirow{2}{*}{Total $\uparrow$} & \multirow{2}{*}{Sparsity $\uparrow$} & \multirow{2}{*}{Latency $\downarrow$} & \multirow{2}{*}{Speedup $\uparrow$} & GPU \\
  Prompts & Prompts & & & & & & & hrs $\downarrow$ \\
  \midrule
  \multicolumn{2}{c|}{Dense (FA3)} & $81.27$ & $71.65$ & $79.35$ & $0\%$ & $1244$s & $1.00\times$ & -- \\
  \midrule
  64 & 64 & $81.41$ & $72.81$ & $79.69$ & $62.5\%$ & $785$s & $1.58\times$ & $89.6$ \\
  64 & 1 & $81.62$ & $71.24$ & $79.55$ & $63.4\%$ & $772$s & $1.61\times$ & $53.0$ \\
  32 & 1 & $81.47$ & $72.47$ & $79.67$ & $62.0\%$ & $796$s & $1.56\times$ & $27.0$ \\
  16 & 1 & $81.36$ & $71.81$ & $79.45$ & $61.3\%$ & $801$s & $1.55\times$ & $13.7$ \\
  8 & 1  & $80.41$ & $71.02$ & $78.53$ & $62.7$\%     & $784$s     & $1.58\times$           & $7.1$ \\
  \bottomrule
  \end{tabular}
  \vspace{-12pt}
\end{table}

%% file: figures/kernel_size_sparse.tex
\begin{wrapfigure}[25]{r}{0.34\linewidth}
    \vspace{-40pt}
    \centering
    \begin{subfigure}{\linewidth}
        \centering
        \caption{Fine block size}
        \label{fig:block_size_small}
        \vspace{-1pt}
        \includegraphics[width=0.9\linewidth]{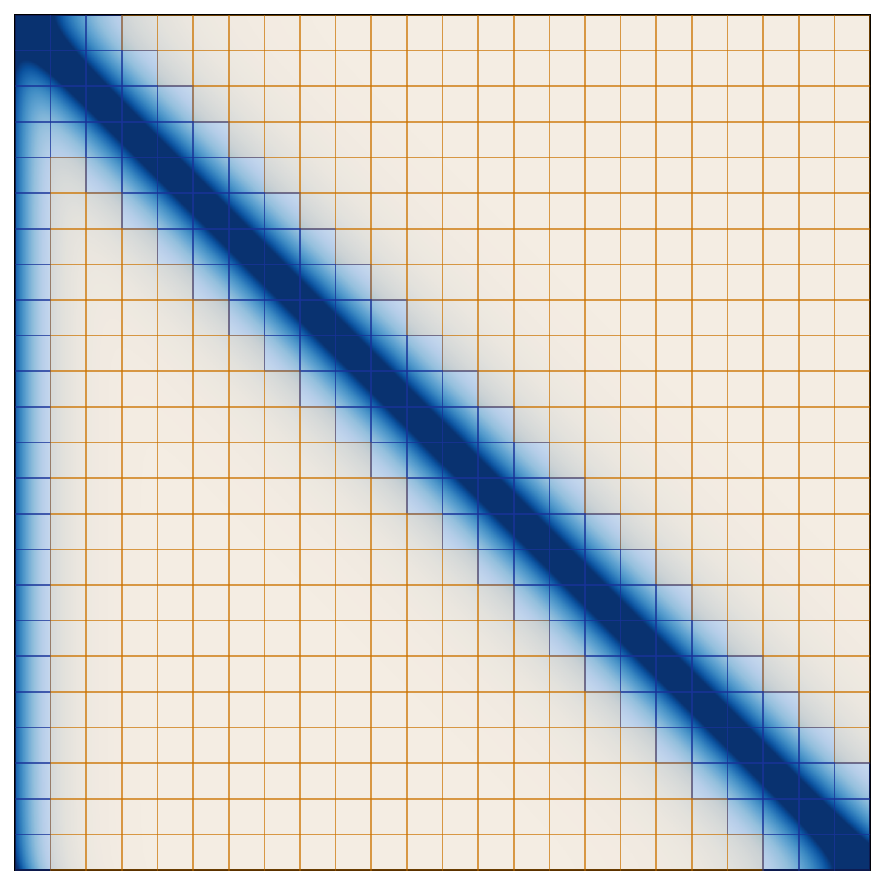}
    \end{subfigure}\\[-3pt]
    \begin{subfigure}{\linewidth}
        \centering
        \caption{Large block size}
        \label{fig:block_size_large}
        \vspace{-1pt}
        \includegraphics[width=0.9\linewidth]{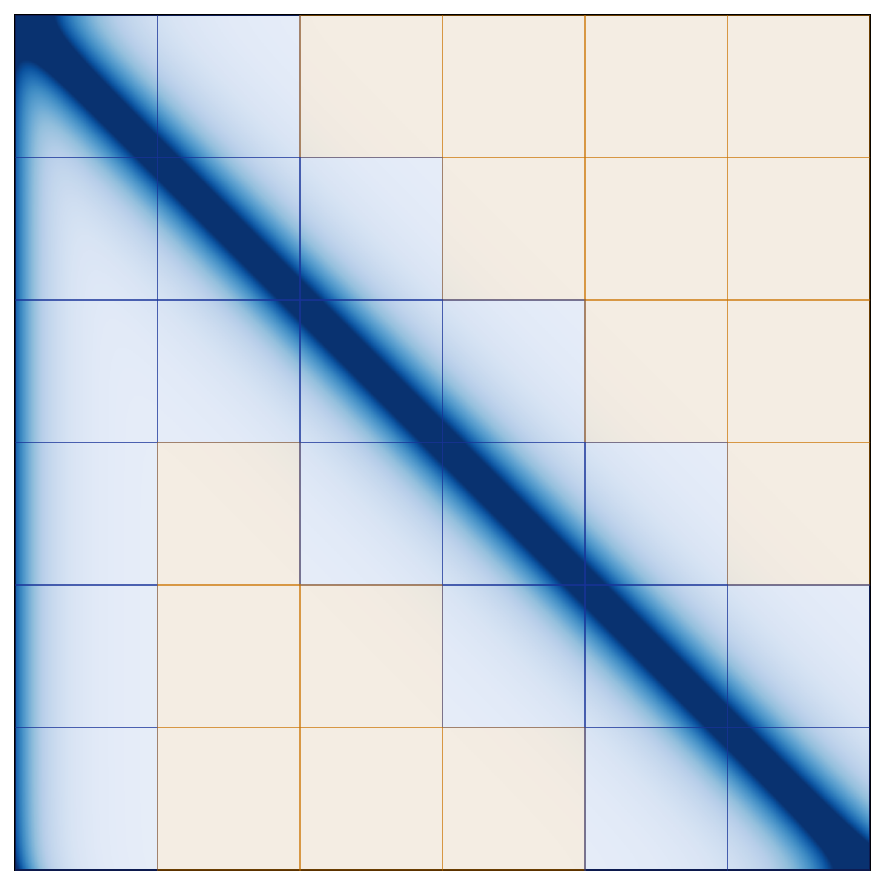}
    \end{subfigure}\\[-2pt]
    \includegraphics[height=1.3em]{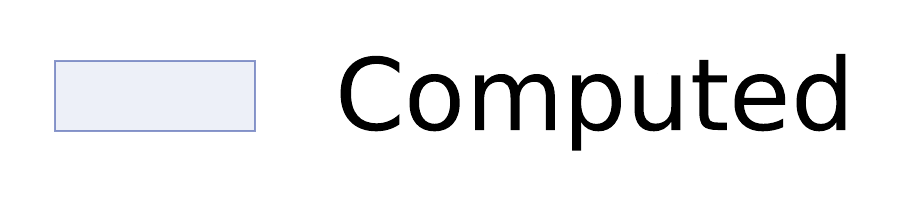}\hspace{0pt}
    \includegraphics[height=1.3em]{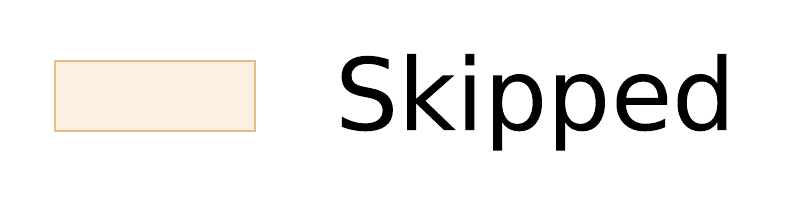}\\[-3pt]
    \includegraphics[height=1.3em]{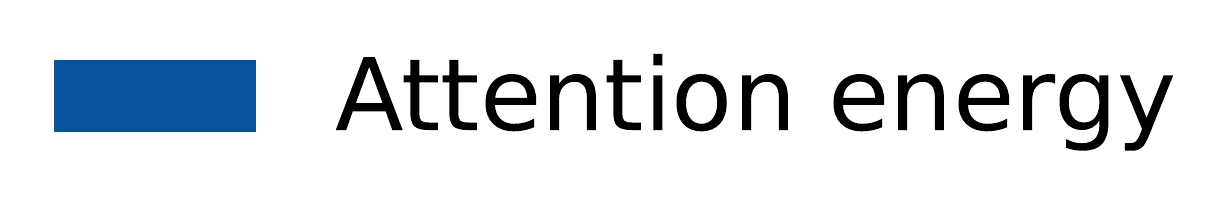}
    \vspace{-9pt}
    \caption{Effect of block size on achievable sparsity. (a) Fine blocks enable finer-grained selection. (b) Large blocks reduce sparsity.}
    \label{fig:block_size_ablation}
    \vspace{-8pt}
\end{wrapfigure}

%% file: tables/kernel_size.tex
\begin{table}[t]                                                
  \centering                                                    
  \caption{Effect of kernel block size on sparsity and      
  quality for Wan 2.1 14B at 480p. Speedup is relative to dense
  FlashAttention-3 baseline. Latency is per-prompt wall-clock time
   on a single H100 GPU.}
  \label{tab:block_size_ablation}
  \setlength{\tabcolsep}{1pt}
  \begin{tabular}{@{}lc|ccc|ccc@{}}                                 
  \toprule                                                        
  Method & $B_q \times B_{kv}$ & Quality $\uparrow$ & Semantic $\uparrow$ & Total $\uparrow$ & Sparsity $\uparrow$ & Latency $\downarrow$ & Speedup $\uparrow$ \\
  \midrule
  Dense (FA3) & $128 \times 176$ & $82.19$ & $72.71$ & $80.29$ & $0\%$ & $363$s & $1.00\times$ \\
  \midrule
  \multirow{7}{*}{\algoname\ (Ours)}
   & $128 \times \ 64$  & $82.16$ & $72.91$ & $80.31$ & $64.9\%$ & $265$s & $1.37\times$ \\
   & $128 \times \ 80$  & $82.08$ & $73.12$ & $80.29$ & $64.6\%$ & $259$s & $1.40\times$ \\
   & $128 \times \ 96$  & $81.92$ & $74.00$ & $80.34$ & $64.1\%$ & $258$s & $1.41\times$ \\
   & $128 \times 128$ & $82.23$ & $72.91$ & $80.37$ & $63.4\%$ & $257$s & $1.41\times$ \\
   & $128 \times 144$ & $82.18$ & $73.05$ & $80.35$ & $63.1\%$ & $258$s & $1.41\times$ \\
   & $128 \times 176$ & $82.27$ & $72.68$ & $80.35$ & $62.5\%$ & $255$s & $1.42\times$ \\
   & $128 \times 192$ & $82.34$ & $72.47$ & $80.37$ & $62.1\%$ & $257$s & $1.41\times$ \\
  \bottomrule
  \end{tabular}
\end{table}

%% file: sections/sup/add_qual.tex
\FloatBarrier

\section{Additional Qualitative Results}
\label{sec:supp_qualitative}

\noindent
\cref{fig:sup_two_prompts_four_rows_480p,fig:sup_two_prompts_four_rows_720p} provide additional qualitative comparisons between \algoname\ and FlashAttention3~\cite{shah2024flashattention3} dense attention on Wan2.1~\cite{wan2025} 14B at 480p and 720p.
\cref{fig:sup_two_prompts_four_rows_mochi} shows comparisons on Mochi~1~\cite{genmo2024mochi} at 480p.
\cref{fig:sup_two_prompts_four_rows_distilled_480p,fig:sup_two_prompts_four_rows_distilled_720p} show comparisons on the distilled LightX2V~\cite{lightx2v} model at 480p and 720p.
Across diverse prompts, \algoname\ preserves visual fidelity, temporal coherence, and prompt alignment while substantially increasing attention sparsity.
\input{figures/video_frame_comp_sup_480p}
\input{figures/video_frame_comp_sup_720p}
\input{figures/video_frame_comp_sup_mochi}
\input{figures/video_frame_comp_sup_480p_dist}
\input{figures/video_frame_comp_sup_720p_dist}

%% file: figures/video_frame_comp_sup_480p.tex
\def\promptA{Dog}
\def\promptB{Cat}
\def\promptC{Koala}
\def\promptD{Jellyfish}

\def\textA{A cute happy Corgi playing in park, sunset.}
\def\textB{A cat eating food out of a bowl.}
\def\textC{A koala bear playing piano in the forest.}
\def\textD{A jellyfish floating through the ocean, with bioluminescent tentacles.}

\def\marginMetrics{5.25em}

\begin{figure}[!h]
\centering
\vspace{-1.0em}

\setlength{\fboxsep}{2pt}

\parbox{\linewidth}{%

\makebox[\linewidth]{%
    \llap{\textcolor{red}{sparsity=0\%}\,\,\hspace{\marginMetrics}}
    \makebox[0pt][c]{Dense Attention}%
    \makebox[0pt][l]{\hspace{\marginMetrics}
        \textcolor{red}{latency=6m03s}
    }%
}

\vspace{0.2em}
\includegraphics[width=0.195\linewidth]{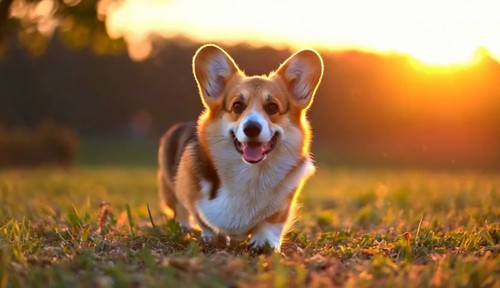}\hfill
\includegraphics[width=0.195\linewidth]{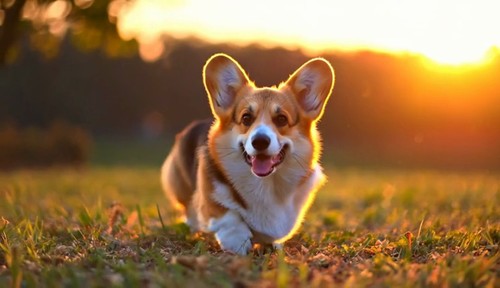}\hfill
\includegraphics[width=0.195\linewidth]{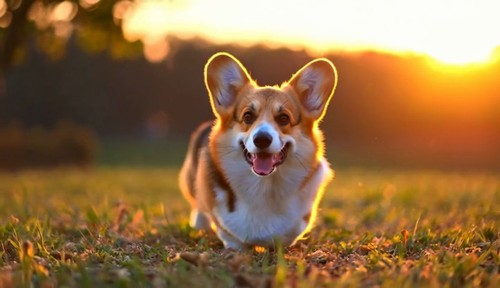}\hfill
\includegraphics[width=0.195\linewidth]{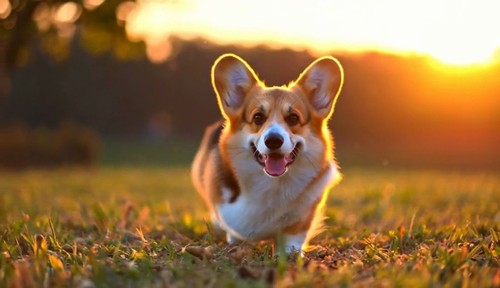}\hfill
\includegraphics[width=0.195\linewidth]{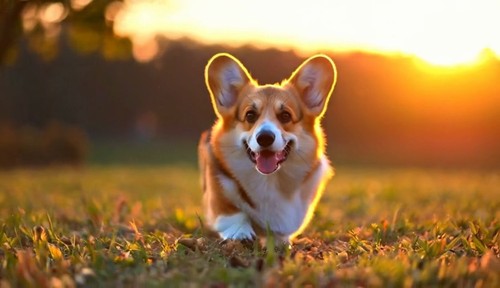}

\vspace{0.4em}

\makebox[\linewidth]{%
    \llap{\textcolor{green!60!black}{sparsity=68\%}\,\,\hspace{\marginMetrics}}
    \makebox[0pt][c]{\algoname}%
    \makebox[0pt][l]{\hspace{\marginMetrics}
        \textcolor{green!60!black}{latency=4m10s}
    }%
}

\vspace{0.2em}
\includegraphics[width=0.195\linewidth]{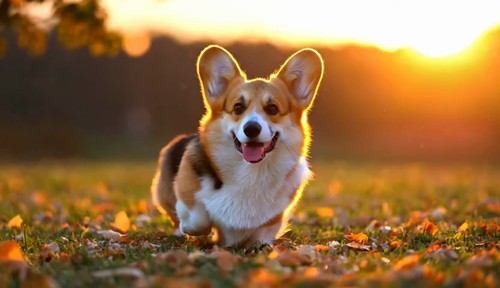}\hfill
\includegraphics[width=0.195\linewidth]{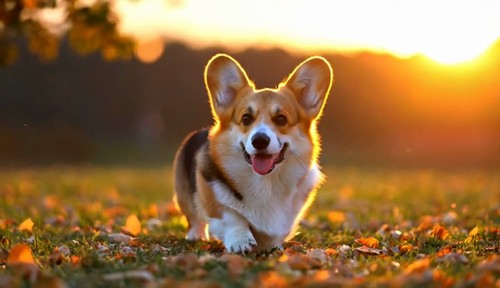}\hfill
\includegraphics[width=0.195\linewidth]{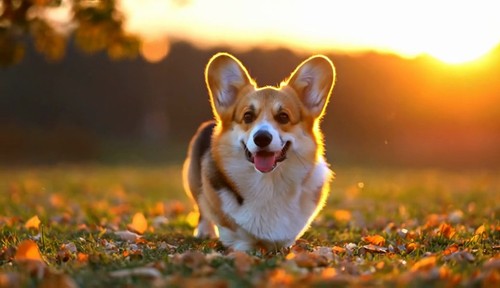}\hfill
\includegraphics[width=0.195\linewidth]{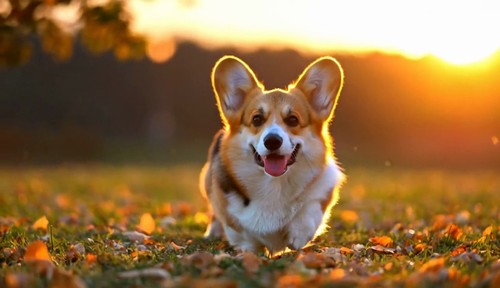}\hfill
\includegraphics[width=0.195\linewidth]{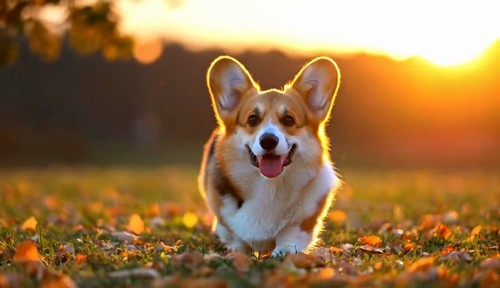}

\parbox{\linewidth}{\centering
\footnotesize \textit{\textcolor{black!70}{``\textA''}}
}

}

\vspace{1.0em}

\parbox{\linewidth}{%

\makebox[\linewidth]{%
    \llap{\textcolor{red}{sparsity=0\%}\,\,\hspace{\marginMetrics}}
    \makebox[0pt][c]{Dense Attention}%
    \makebox[0pt][l]{\hspace{\marginMetrics}
        \textcolor{red}{latency=6m03s}
    }%
}

\vspace{0.2em}
\includegraphics[width=0.195\linewidth]{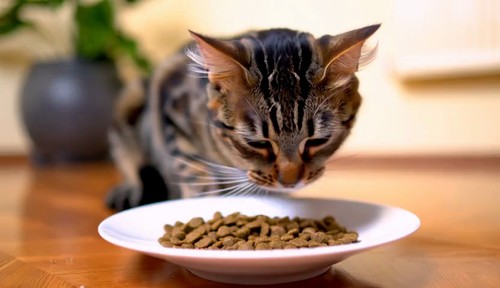}\hfill
\includegraphics[width=0.195\linewidth]{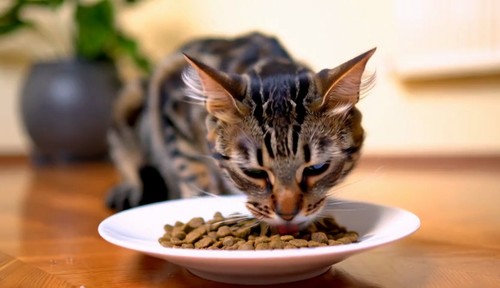}\hfill
\includegraphics[width=0.195\linewidth]{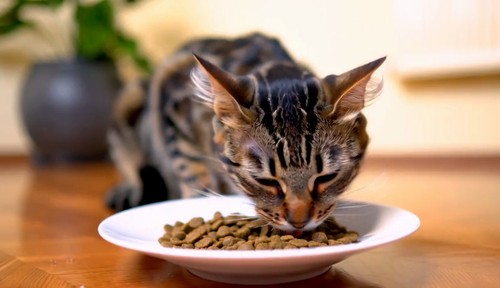}\hfill
\includegraphics[width=0.195\linewidth]{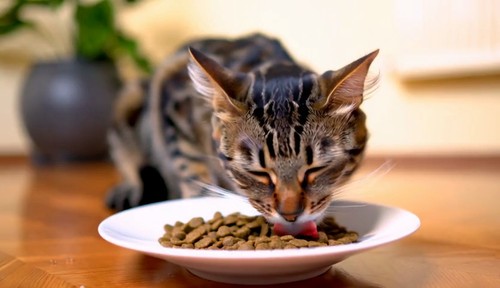}\hfill
\includegraphics[width=0.195\linewidth]{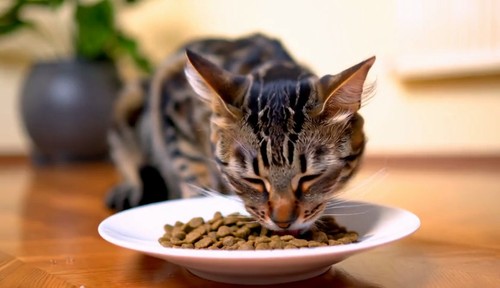}

\vspace{0.4em}

\makebox[\linewidth]{%
    \llap{\textcolor{green!60!black}{sparsity=68\%}\,\,\hspace{\marginMetrics}}
    \makebox[0pt][c]{\algoname}%
    \makebox[0pt][l]{\hspace{\marginMetrics}
        \textcolor{green!60!black}{latency=4m10s}
    }%
}

\vspace{0.2em}
\includegraphics[width=0.195\linewidth]{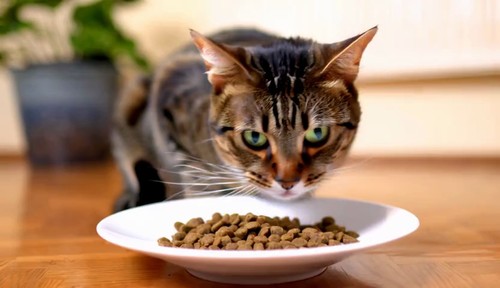}\hfill
\includegraphics[width=0.195\linewidth]{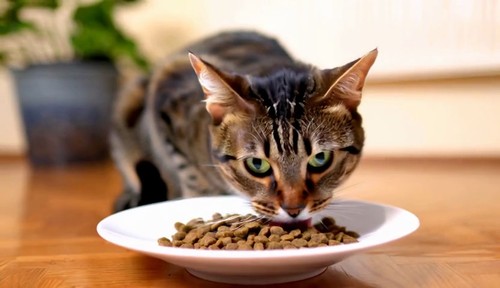}\hfill
\includegraphics[width=0.195\linewidth]{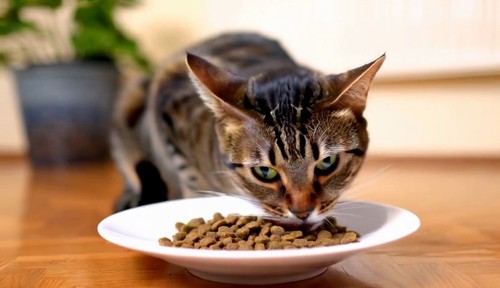}\hfill
\includegraphics[width=0.195\linewidth]{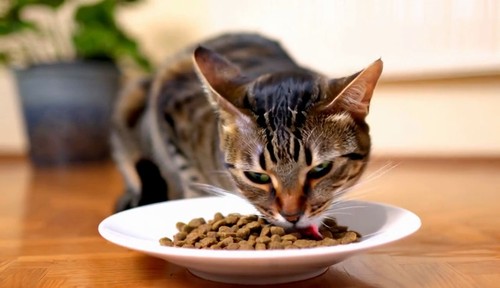}\hfill
\includegraphics[width=0.195\linewidth]{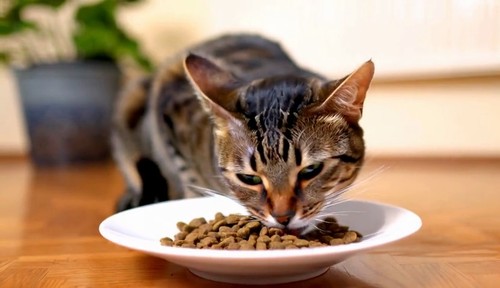}

\parbox{\linewidth}{\centering
\footnotesize \textit{\textcolor{black!70}{``\textB''}}
}

}

\vspace{1.0em}

\parbox{\linewidth}{%

\makebox[\linewidth]{%
    \llap{\textcolor{red}{sparsity=0\%}\,\,\hspace{\marginMetrics}}
    \makebox[0pt][c]{Dense Attention}%
    \makebox[0pt][l]{\hspace{\marginMetrics}
        \textcolor{red}{latency=6m03s}
    }%
}

\vspace{0.2em}
\includegraphics[width=0.195\linewidth]{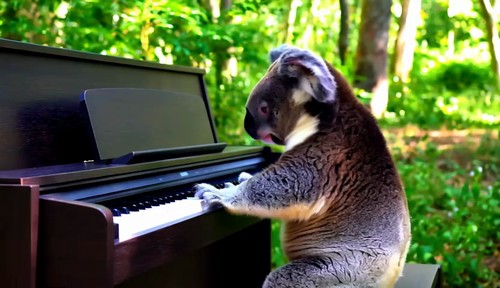}\hfill
\includegraphics[width=0.195\linewidth]{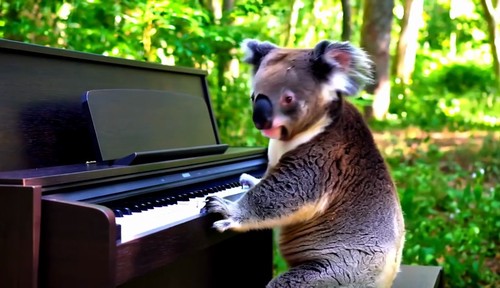}\hfill
\includegraphics[width=0.195\linewidth]{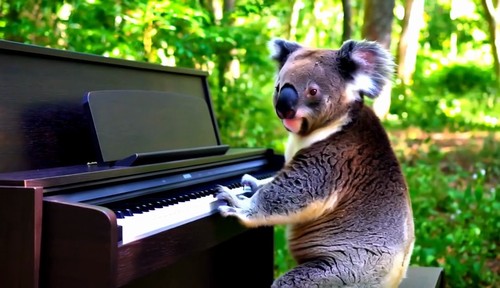}\hfill
\includegraphics[width=0.195\linewidth]{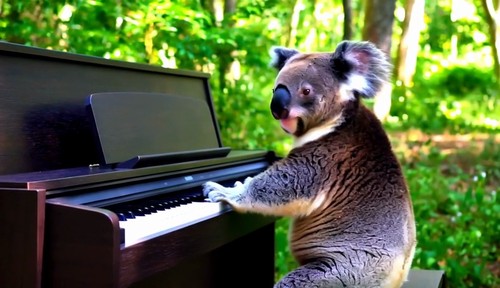}\hfill
\includegraphics[width=0.195\linewidth]{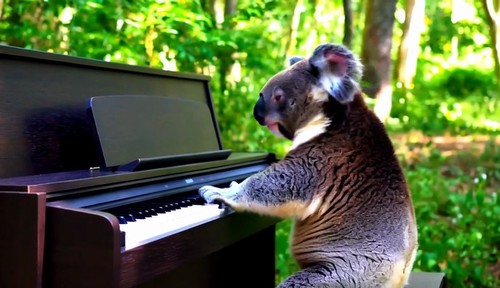}

\vspace{0.4em}

\makebox[\linewidth]{%
    \llap{\textcolor{green!60!black}{sparsity=68\%}\,\,\hspace{\marginMetrics}}
    \makebox[0pt][c]{\algoname}%
    \makebox[0pt][l]{\hspace{\marginMetrics}
        \textcolor{green!60!black}{latency=4m10s}
    }%
}

\vspace{0.2em}
\includegraphics[width=0.195\linewidth]{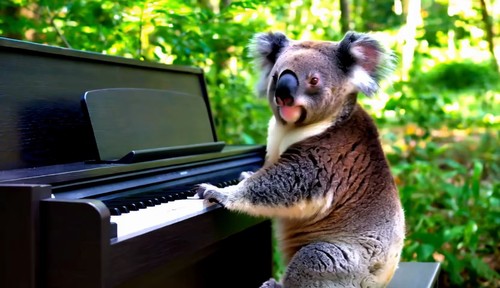}\hfill
\includegraphics[width=0.195\linewidth]{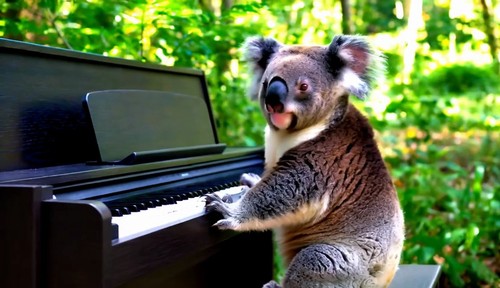}\hfill
\includegraphics[width=0.195\linewidth]{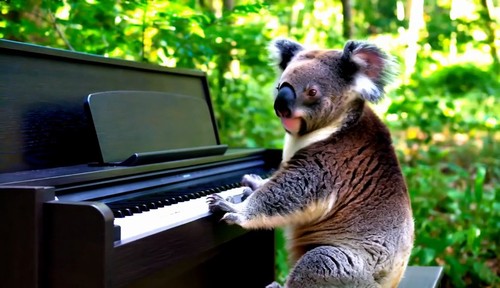}\hfill
\includegraphics[width=0.195\linewidth]{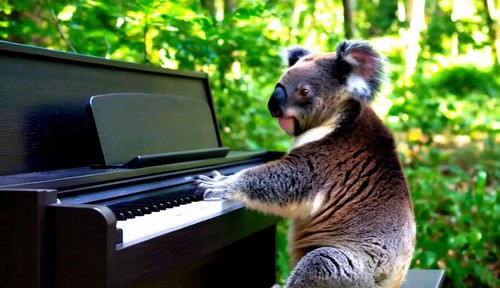}\hfill
\includegraphics[width=0.195\linewidth]{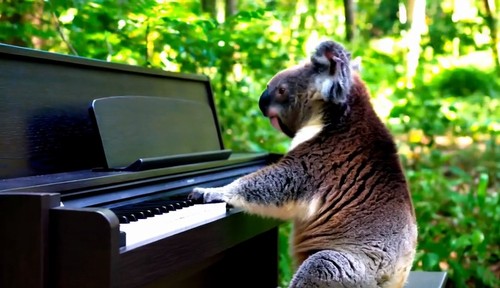}

\parbox{\linewidth}{\centering
\footnotesize \textit{\textcolor{black!70}{``\textC''}}
}

}

\vspace{1.0em}

\parbox{\linewidth}{%

\makebox[\linewidth]{%
    \llap{\textcolor{red}{sparsity=0\%}\,\,\hspace{\marginMetrics}}
    \makebox[0pt][c]{Dense Attention}%
    \makebox[0pt][l]{\hspace{\marginMetrics}
        \textcolor{red}{latency=6m03s}
    }%
}

\vspace{0.2em}
\includegraphics[width=0.195\linewidth]{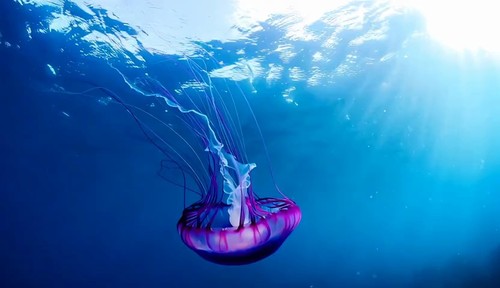}\hfill
\includegraphics[width=0.195\linewidth]{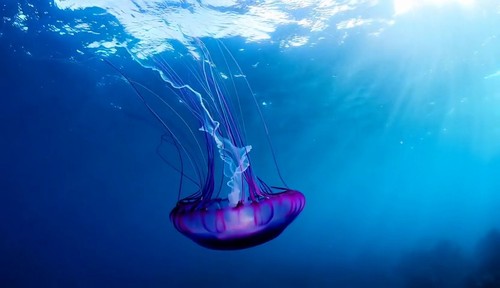}\hfill
\includegraphics[width=0.195\linewidth]{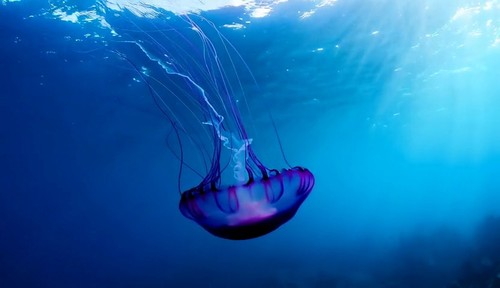}\hfill
\includegraphics[width=0.195\linewidth]{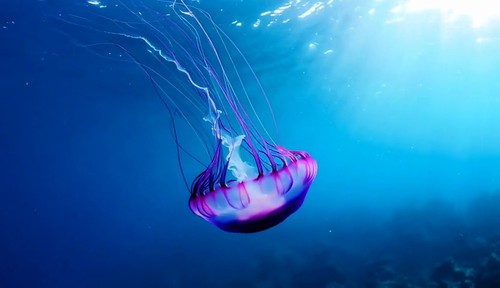}\hfill
\includegraphics[width=0.195\linewidth]{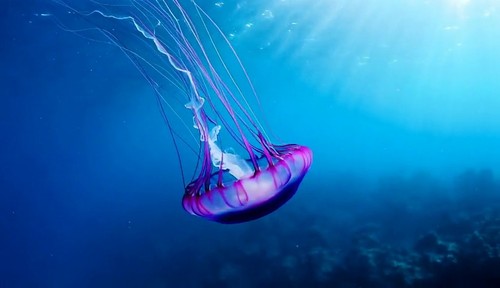}

\vspace{0.4em}

\makebox[\linewidth]{%
    \llap{\textcolor{green!60!black}{sparsity=68\%}\,\,\hspace{\marginMetrics}}
    \makebox[0pt][c]{\algoname}%
    \makebox[0pt][l]{\hspace{\marginMetrics}
        \textcolor{green!60!black}{latency=4m10s}
    }%
}

\vspace{0.2em}
\includegraphics[width=0.195\linewidth]{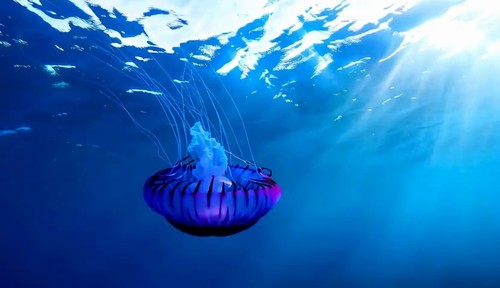}\hfill
\includegraphics[width=0.195\linewidth]{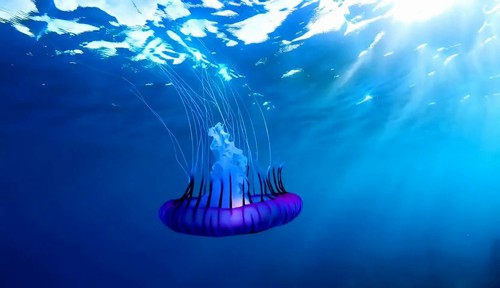}\hfill
\includegraphics[width=0.195\linewidth]{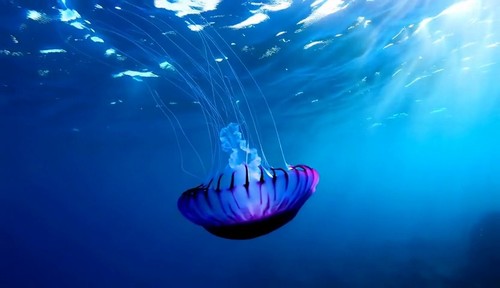}\hfill
\includegraphics[width=0.195\linewidth]{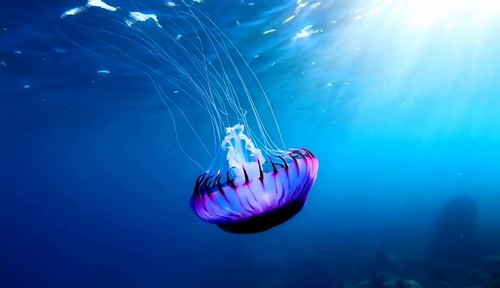}\hfill
\includegraphics[width=0.195\linewidth]{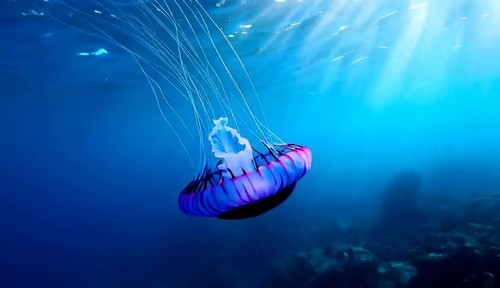}

\parbox{\linewidth}{\centering
\footnotesize \textit{\textcolor{black!70}{``\textD''}}
}

}

\caption{
Comparison of four prompts generated with the same seed on Wan2.1 14B 480p text-to-video.
\algoname\ achieves higher attention sparsity and lower end-to-end latency while maintaining visual quality and prompt alignment.
}
\label{fig:sup_two_prompts_four_rows_480p}

\end{figure}

%% file: figures/video_frame_comp_sup_720p.tex
\def\promptA{Beach}
\def\promptB{Eiffel}
\def\promptC{Person}
\def\promptD{Mountain}

\def\textA{A beautiful coastal beach in spring, waves lapping on sand.}
\def\textB{A boat sailing leisurely along the Seine River with the Eiffel Tower in background, animated style.}
\def\textC{A person is cleaning windows.}
\def\textD{Snow rocky mountains peaks canyon... through the high elevated mountain peaks.}

\def\marginMetrics{5.25em}

\begin{figure}[!h]
\centering
\vspace{-1.0em}

\setlength{\fboxsep}{2pt}

\parbox{\linewidth}{%

\makebox[\linewidth]{%
    \llap{\textcolor{red}{sparsity=0\%}\,\,\hspace{\marginMetrics}}
    \makebox[0pt][c]{Dense Attention}%
    \makebox[0pt][l]{\hspace{\marginMetrics}
        \textcolor{red}{latency=20m44s}
    }%
}

\vspace{0.2em}
\includegraphics[width=0.195\linewidth]{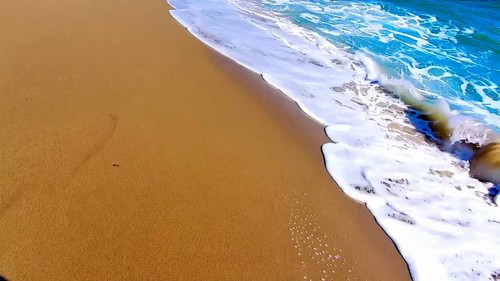}\hfill
\includegraphics[width=0.195\linewidth]{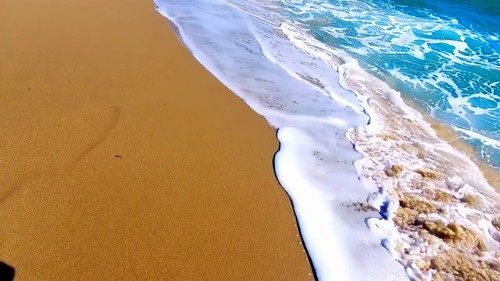}\hfill
\includegraphics[width=0.195\linewidth]{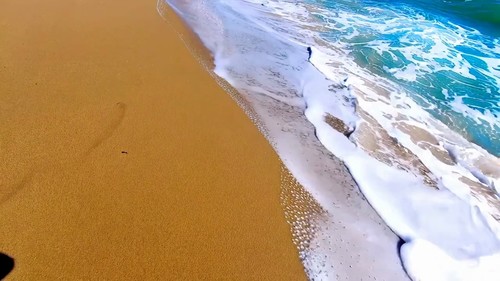}\hfill
\includegraphics[width=0.195\linewidth]{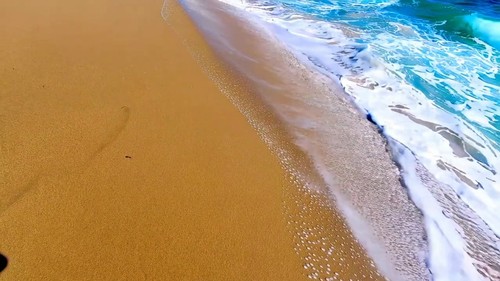}\hfill
\includegraphics[width=0.195\linewidth]{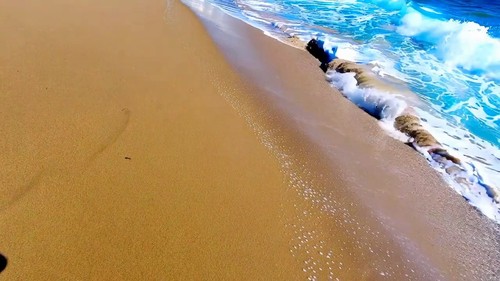}

\vspace{0.4em}

\makebox[\linewidth]{%
    \llap{\textcolor{green!60!black}{sparsity=62\%}\,\,\hspace{\marginMetrics}}
    \makebox[0pt][c]{\algoname}%
    \makebox[0pt][l]{\hspace{\marginMetrics}
        \textcolor{green!60!black}{latency=13m05s}
    }%
}

\vspace{0.2em}
\includegraphics[width=0.195\linewidth]{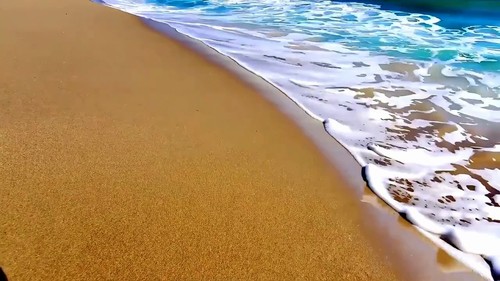}\hfill
\includegraphics[width=0.195\linewidth]{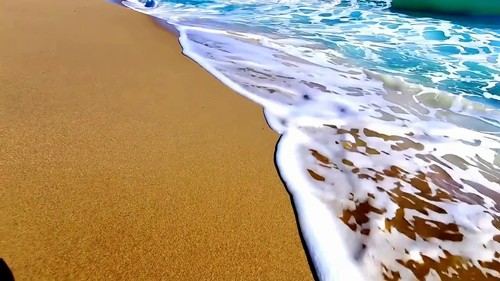}\hfill
\includegraphics[width=0.195\linewidth]{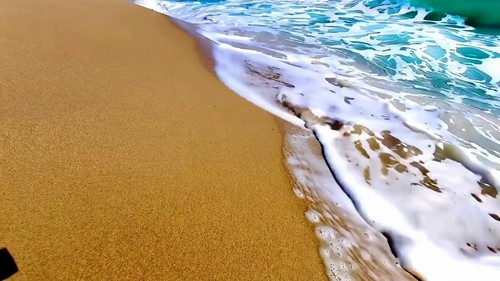}\hfill
\includegraphics[width=0.195\linewidth]{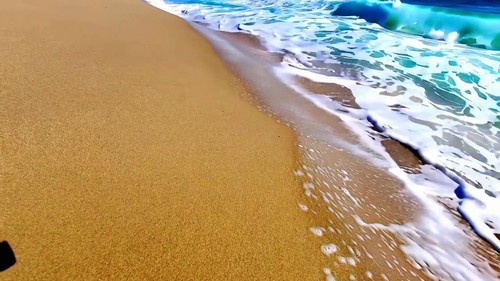}\hfill
\includegraphics[width=0.195\linewidth]{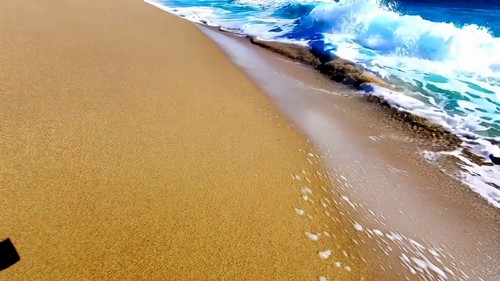}

\parbox{\linewidth}{\centering
\footnotesize \textit{\textcolor{black!70}{``\textA''}}
}

}

\vspace{1.0em}

\parbox{\linewidth}{%

\makebox[\linewidth]{%
    \llap{\textcolor{red}{sparsity=0\%}\,\,\hspace{\marginMetrics}}
    \makebox[0pt][c]{Dense Attention}%
    \makebox[0pt][l]{\hspace{\marginMetrics}
        \textcolor{red}{latency=20m44s}
    }%
}

\vspace{0.2em}
\includegraphics[width=0.195\linewidth]{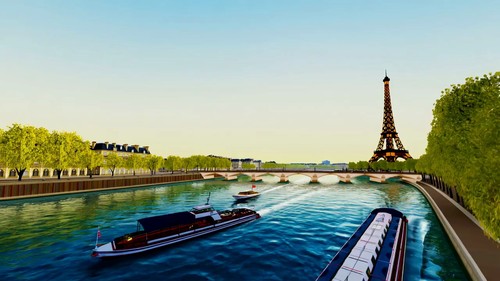}\hfill
\includegraphics[width=0.195\linewidth]{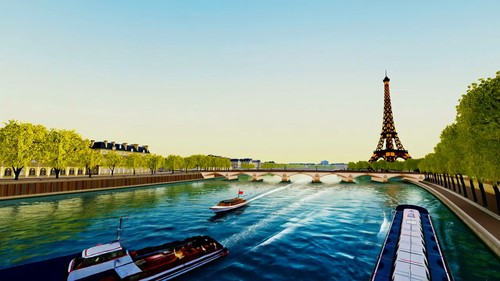}\hfill
\includegraphics[width=0.195\linewidth]{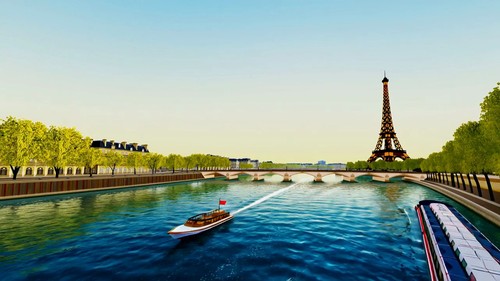}\hfill
\includegraphics[width=0.195\linewidth]{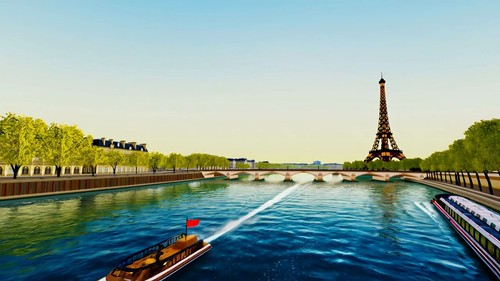}\hfill
\includegraphics[width=0.195\linewidth]{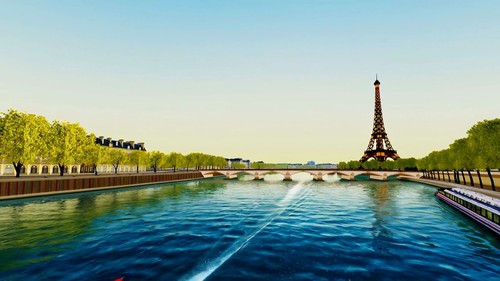}

\vspace{0.4em}

\makebox[\linewidth]{%
    \llap{\textcolor{green!60!black}{sparsity=62\%}\,\,\hspace{\marginMetrics}}
    \makebox[0pt][c]{\algoname}%
    \makebox[0pt][l]{\hspace{\marginMetrics}
        \textcolor{green!60!black}{latency=13m05s}
    }%
}

\vspace{0.2em}
\includegraphics[width=0.195\linewidth]{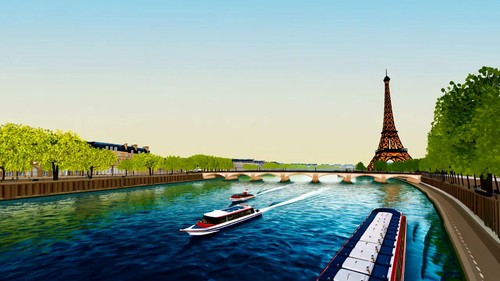}\hfill
\includegraphics[width=0.195\linewidth]{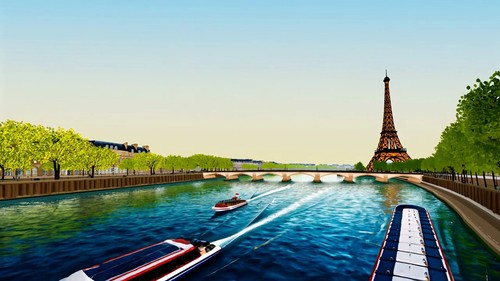}\hfill
\includegraphics[width=0.195\linewidth]{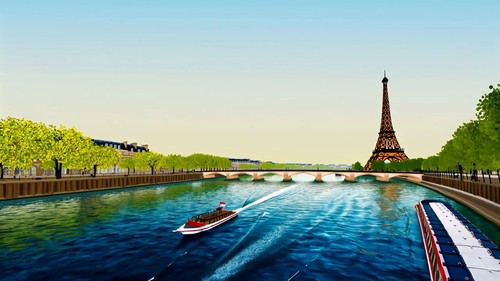}\hfill
\includegraphics[width=0.195\linewidth]{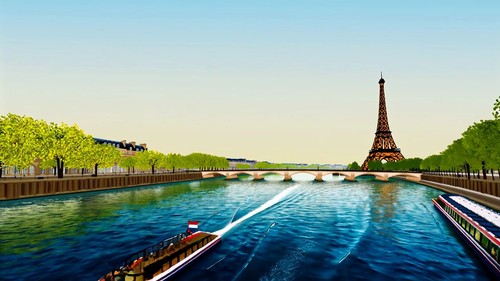}\hfill
\includegraphics[width=0.195\linewidth]{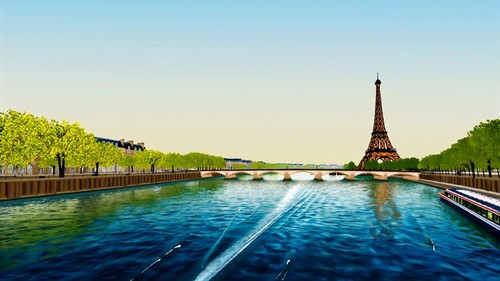}

\parbox{\linewidth}{\centering
\footnotesize \textit{\textcolor{black!70}{``\textB''}}
}

}

\vspace{1.0em}

\parbox{\linewidth}{%

\makebox[\linewidth]{%
    \llap{\textcolor{red}{sparsity=0\%}\,\,\hspace{\marginMetrics}}
    \makebox[0pt][c]{Dense Attention}%
    \makebox[0pt][l]{\hspace{\marginMetrics}
        \textcolor{red}{latency=20m44s}
    }%
}

\vspace{0.2em}
\includegraphics[width=0.195\linewidth]{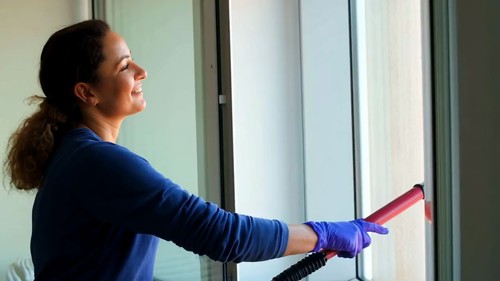}\hfill
\includegraphics[width=0.195\linewidth]{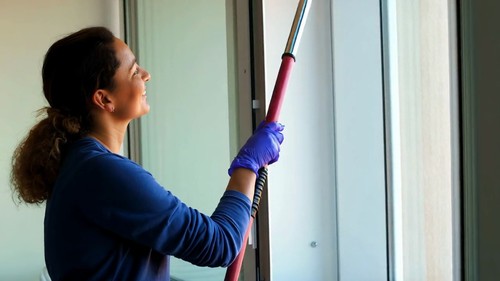}\hfill
\includegraphics[width=0.195\linewidth]{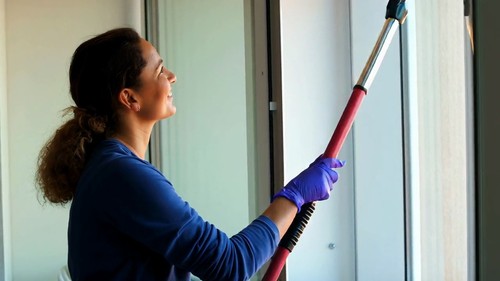}\hfill
\includegraphics[width=0.195\linewidth]{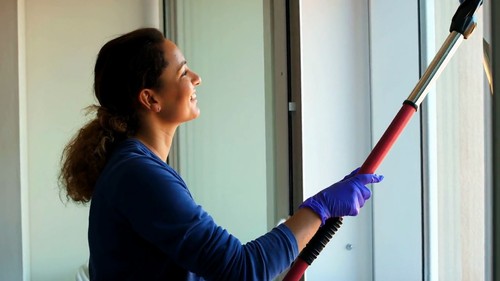}\hfill
\includegraphics[width=0.195\linewidth]{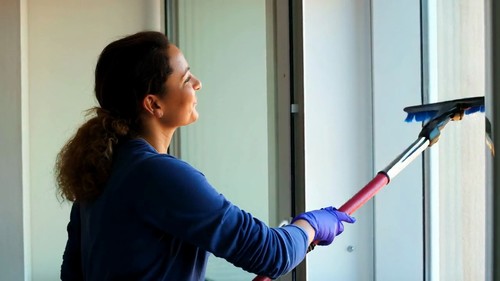}

\vspace{0.4em}

\makebox[\linewidth]{%
    \llap{\textcolor{green!60!black}{sparsity=62\%}\,\,\hspace{\marginMetrics}}
    \makebox[0pt][c]{\algoname}%
    \makebox[0pt][l]{\hspace{\marginMetrics}
        \textcolor{green!60!black}{latency=13m05s}
    }%
}

\vspace{0.2em}
\includegraphics[width=0.195\linewidth]{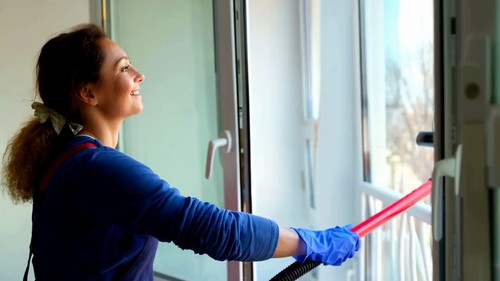}\hfill
\includegraphics[width=0.195\linewidth]{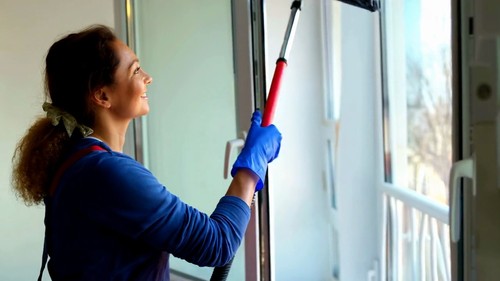}\hfill
\includegraphics[width=0.195\linewidth]{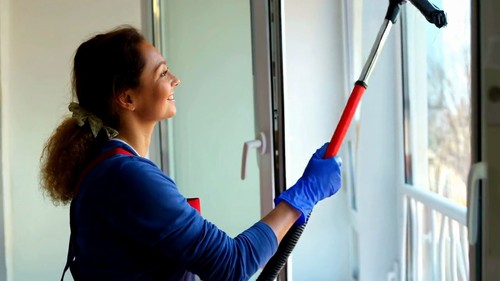}\hfill
\includegraphics[width=0.195\linewidth]{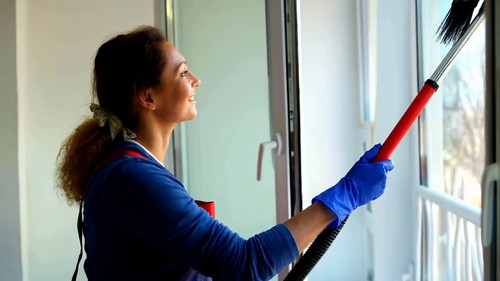}\hfill
\includegraphics[width=0.195\linewidth]{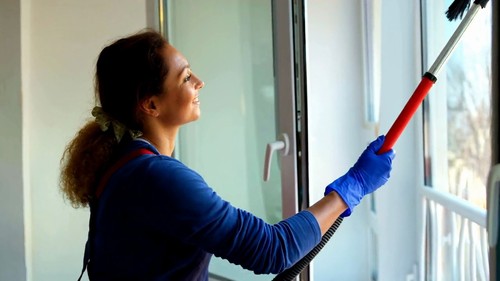}

\parbox{\linewidth}{\centering
\footnotesize \textit{\textcolor{black!70}{``\textC''}}
}

}

\vspace{1.0em}

\parbox{\linewidth}{%

\makebox[\linewidth]{%
    \llap{\textcolor{red}{sparsity=0\%}\,\,\hspace{\marginMetrics}}
    \makebox[0pt][c]{Dense Attention}%
    \makebox[0pt][l]{\hspace{\marginMetrics}
        \textcolor{red}{latency=20m44s}
    }%
}

\vspace{0.2em}
\includegraphics[width=0.195\linewidth]{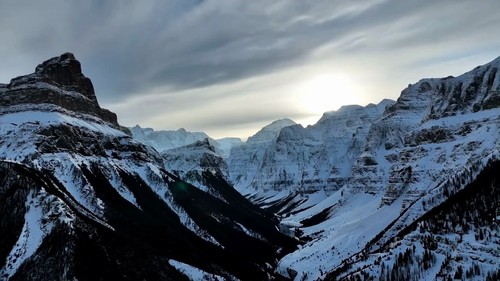}\hfill
\includegraphics[width=0.195\linewidth]{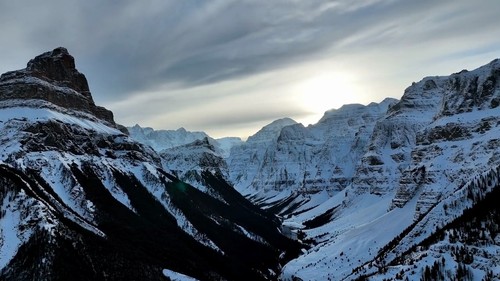}\hfill
\includegraphics[width=0.195\linewidth]{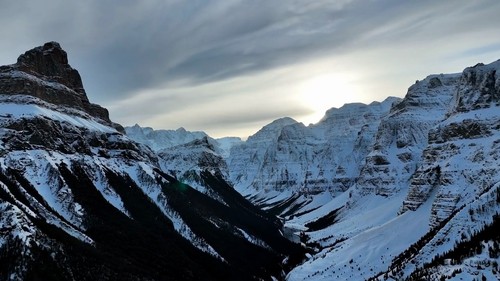}\hfill
\includegraphics[width=0.195\linewidth]{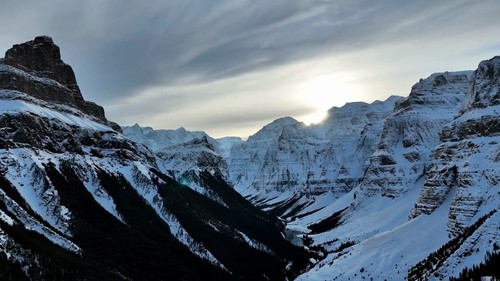}\hfill
\includegraphics[width=0.195\linewidth]{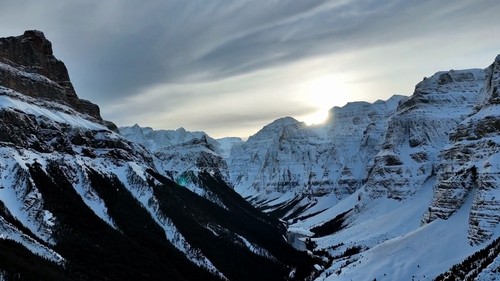}

\vspace{0.4em}

\makebox[\linewidth]{%
    \llap{\textcolor{green!60!black}{sparsity=62\%}\,\,\hspace{\marginMetrics}}
    \makebox[0pt][c]{\algoname}%
    \makebox[0pt][l]{\hspace{\marginMetrics}
        \textcolor{green!60!black}{latency=13m05s}
    }%
}

\vspace{0.2em}
\includegraphics[width=0.195\linewidth]{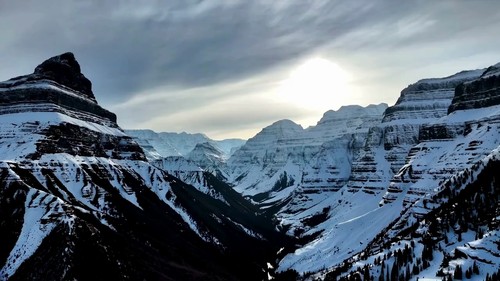}\hfill
\includegraphics[width=0.195\linewidth]{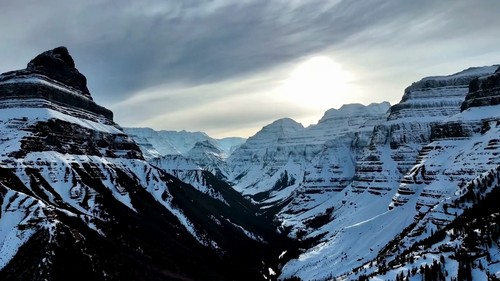}\hfill
\includegraphics[width=0.195\linewidth]{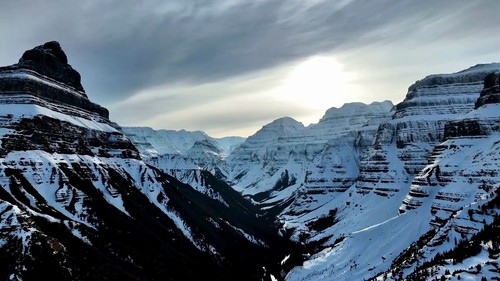}\hfill
\includegraphics[width=0.195\linewidth]{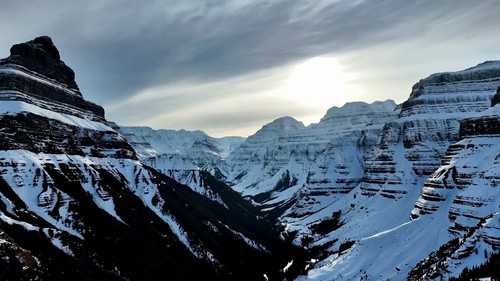}\hfill
\includegraphics[width=0.195\linewidth]{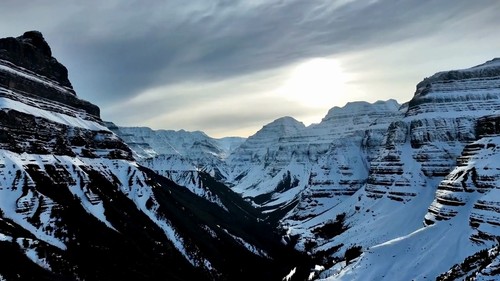}

\parbox{\linewidth}{\centering
\footnotesize \textit{\textcolor{black!70}{``\textD''}}
}

}

\caption{
Comparison of four prompts generated with the same seed on Wan2.1 14B 720p text-to-video.
\algoname\ achieves higher attention sparsity and lower end-to-end latency while maintaining visual quality and prompt alignment.
}
\label{fig:sup_two_prompts_four_rows_720p}

\end{figure}

%% file: figures/video_frame_comp_sup_mochi.tex
\def\promptA{Christmas}
\def\promptB{Plums}
\def\promptC{Cliff}
\def\promptD{Cabin}

\def\textA{A drone view of celebration with Christmas tree and fireworks, starry sky.}
\def\textB{Few big purple plums rotating on the turntable... isolated on the white background.}
\def\textC{A tranquil tableau of cliff.}
\def\textD{A tranquil tableau of a tranquil lakeside cabin nestled... in the calm water.}

\def\marginMetrics{5.25em}

\begin{figure}[!h]
\centering
\vspace{-1.0em}

\setlength{\fboxsep}{2pt}

\parbox{\linewidth}{%

\makebox[\linewidth]{%
    \llap{\textcolor{red}{sparsity=0\%}\,\,\hspace{\marginMetrics}}
    \makebox[0pt][c]{Dense Attention}%
    \makebox[0pt][l]{\hspace{\marginMetrics}
        \textcolor{red}{latency=3m08s}
    }%
}

\vspace{0.2em}
\includegraphics[width=0.195\linewidth]{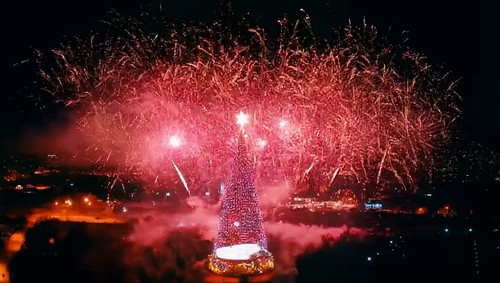}\hfill
\includegraphics[width=0.195\linewidth]{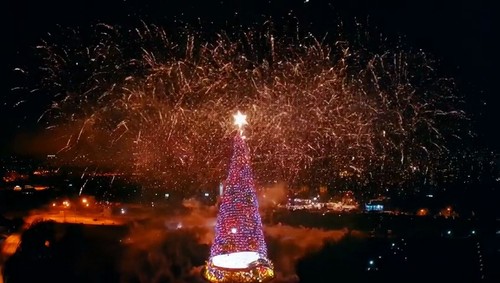}\hfill
\includegraphics[width=0.195\linewidth]{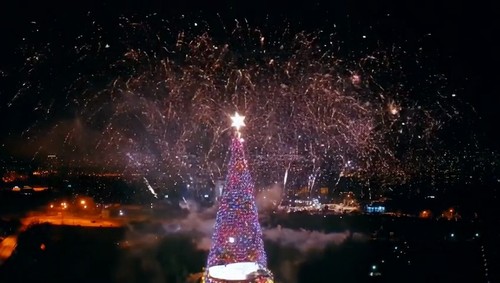}\hfill
\includegraphics[width=0.195\linewidth]{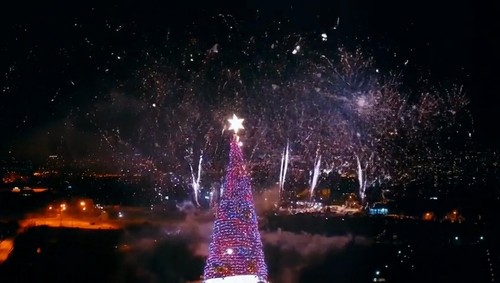}\hfill
\includegraphics[width=0.195\linewidth]{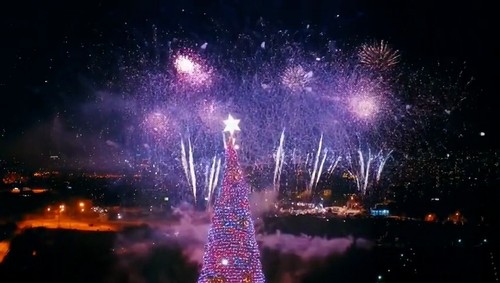}

\vspace{0.4em}

\makebox[\linewidth]{%
    \llap{\textcolor{green!60!black}{sparsity=69\%}\,\,\hspace{\marginMetrics}}
    \makebox[0pt][c]{\algoname}%
    \makebox[0pt][l]{\hspace{\marginMetrics}
        \textcolor{green!60!black}{latency=2m41s}
    }%
}

\vspace{0.2em}
\includegraphics[width=0.195\linewidth]{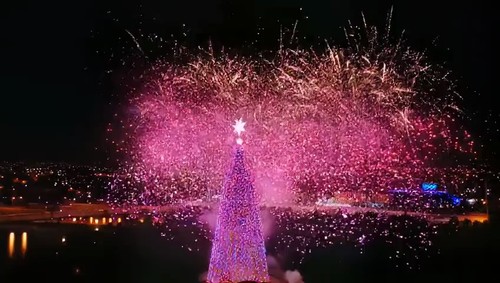}\hfill
\includegraphics[width=0.195\linewidth]{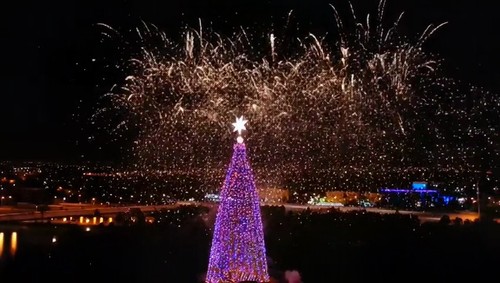}\hfill
\includegraphics[width=0.195\linewidth]{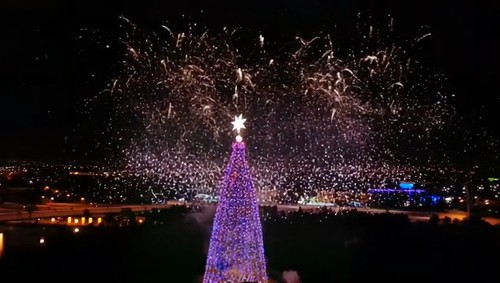}\hfill
\includegraphics[width=0.195\linewidth]{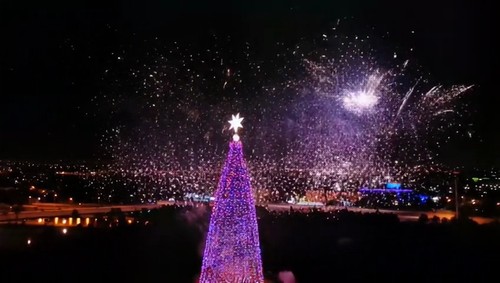}\hfill
\includegraphics[width=0.195\linewidth]{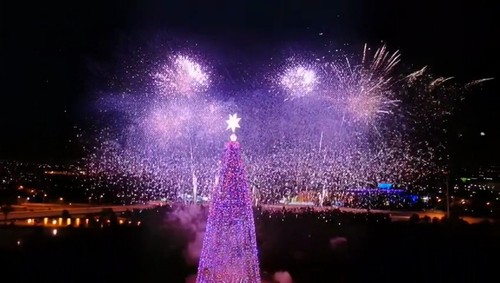}

\parbox{\linewidth}{\centering
\footnotesize \textit{\textcolor{black!70}{``\textA''}}
}

}

\vspace{1.0em}

\parbox{\linewidth}{%

\makebox[\linewidth]{%
    \llap{\textcolor{red}{sparsity=0\%}\,\,\hspace{\marginMetrics}}
    \makebox[0pt][c]{Dense Attention}%
    \makebox[0pt][l]{\hspace{\marginMetrics}
        \textcolor{red}{latency=3m08s}
    }%
}

\vspace{0.2em}
\includegraphics[width=0.195\linewidth]{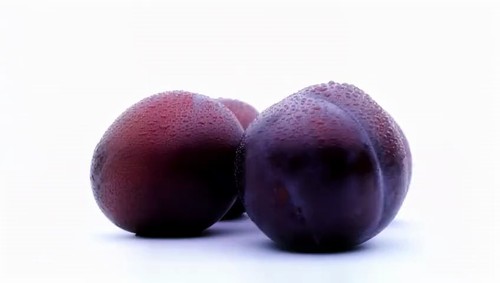}\hfill
\includegraphics[width=0.195\linewidth]{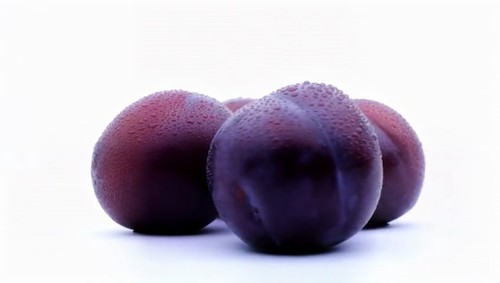}\hfill
\includegraphics[width=0.195\linewidth]{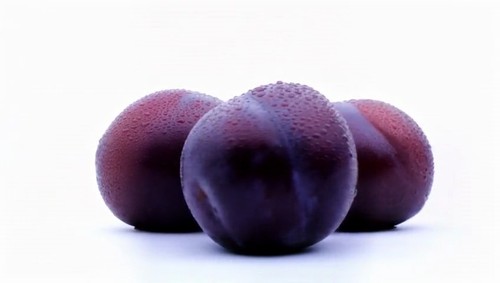}\hfill
\includegraphics[width=0.195\linewidth]{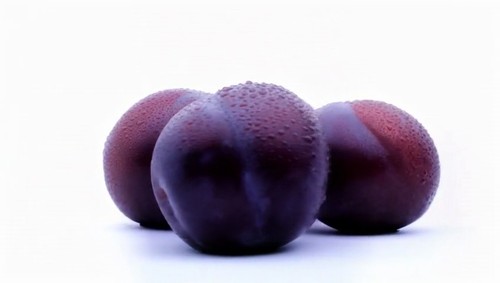}\hfill
\includegraphics[width=0.195\linewidth]{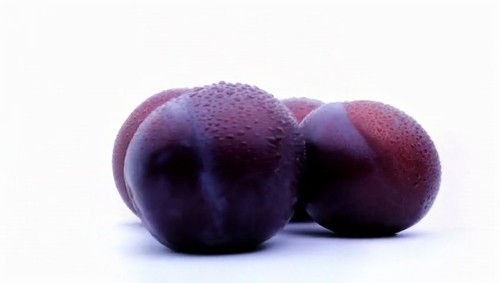}

\vspace{0.4em}

\makebox[\linewidth]{%
    \llap{\textcolor{green!60!black}{sparsity=69\%}\,\,\hspace{\marginMetrics}}
    \makebox[0pt][c]{\algoname}%
    \makebox[0pt][l]{\hspace{\marginMetrics}
        \textcolor{green!60!black}{latency=2m41s}
    }%
}

\vspace{0.2em}
\includegraphics[width=0.195\linewidth]{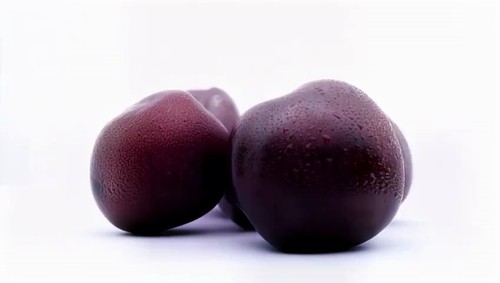}\hfill
\includegraphics[width=0.195\linewidth]{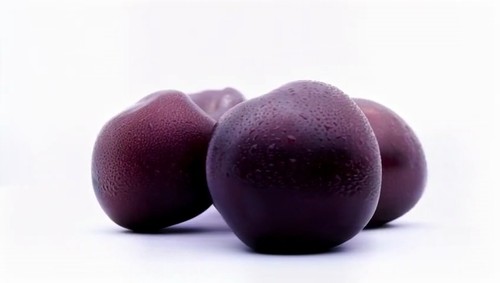}\hfill
\includegraphics[width=0.195\linewidth]{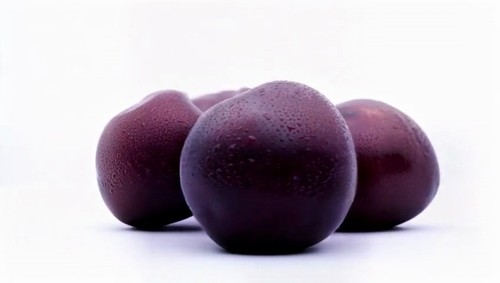}\hfill
\includegraphics[width=0.195\linewidth]{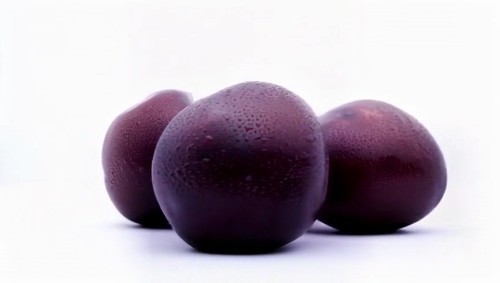}\hfill
\includegraphics[width=0.195\linewidth]{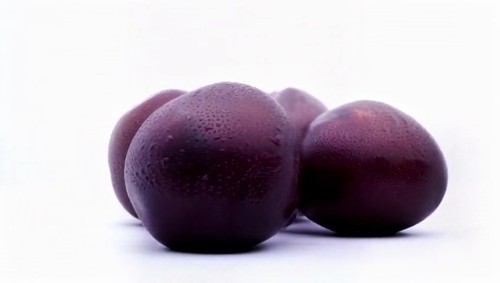}

\parbox{\linewidth}{\centering
\footnotesize \textit{\textcolor{black!70}{``\textB''}}
}

}

\vspace{1.0em}

\parbox{\linewidth}{%

\makebox[\linewidth]{%
    \llap{\textcolor{red}{sparsity=0\%}\,\,\hspace{\marginMetrics}}
    \makebox[0pt][c]{Dense Attention}%
    \makebox[0pt][l]{\hspace{\marginMetrics}
        \textcolor{red}{latency=3m08s}
    }%
}

\vspace{0.2em}
\includegraphics[width=0.195\linewidth]{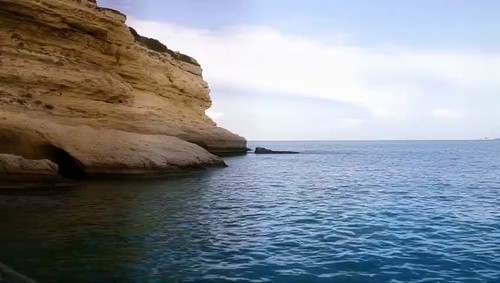}\hfill
\includegraphics[width=0.195\linewidth]{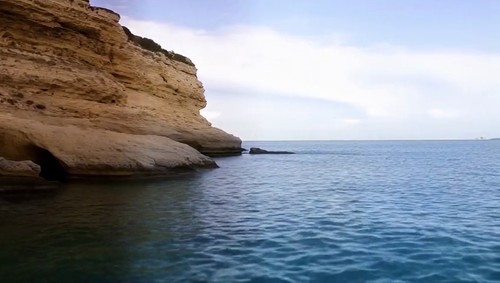}\hfill
\includegraphics[width=0.195\linewidth]{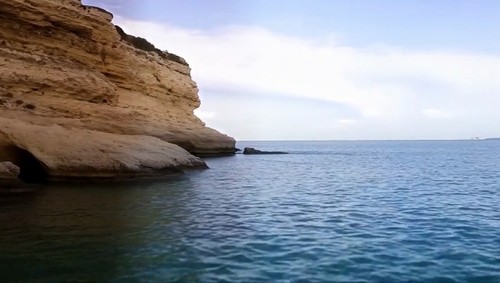}\hfill
\includegraphics[width=0.195\linewidth]{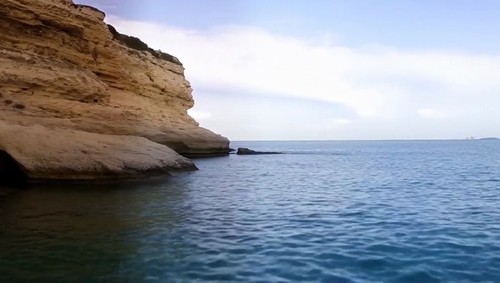}\hfill
\includegraphics[width=0.195\linewidth]{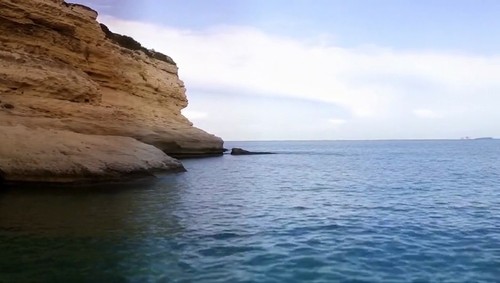}

\vspace{0.4em}

\makebox[\linewidth]{%
    \llap{\textcolor{green!60!black}{sparsity=69\%}\,\,\hspace{\marginMetrics}}
    \makebox[0pt][c]{\algoname}%
    \makebox[0pt][l]{\hspace{\marginMetrics}
        \textcolor{green!60!black}{latency=2m41s}
    }%
}

\vspace{0.2em}
\includegraphics[width=0.195\linewidth]{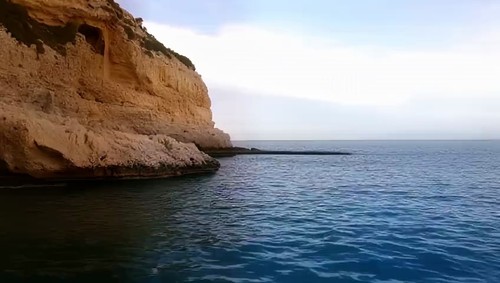}\hfill
\includegraphics[width=0.195\linewidth]{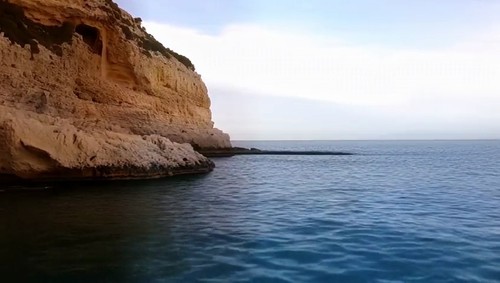}\hfill
\includegraphics[width=0.195\linewidth]{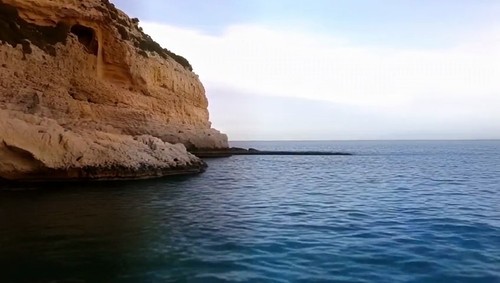}\hfill
\includegraphics[width=0.195\linewidth]{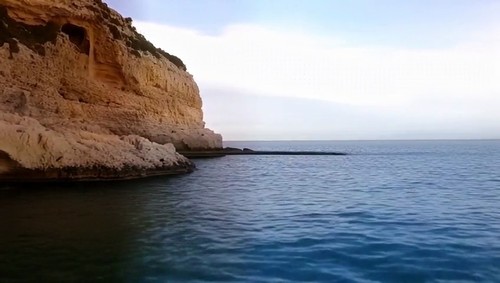}\hfill
\includegraphics[width=0.195\linewidth]{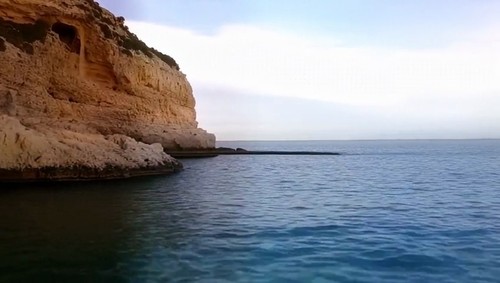}

\parbox{\linewidth}{\centering
\footnotesize \textit{\textcolor{black!70}{``\textC''}}
}

}

\vspace{1.0em}

\parbox{\linewidth}{%

\makebox[\linewidth]{%
    \llap{\textcolor{red}{sparsity=0\%}\,\,\hspace{\marginMetrics}}
    \makebox[0pt][c]{Dense Attention}%
    \makebox[0pt][l]{\hspace{\marginMetrics}
        \textcolor{red}{latency=3m08s}
    }%
}

\vspace{0.2em}
\includegraphics[width=0.195\linewidth]{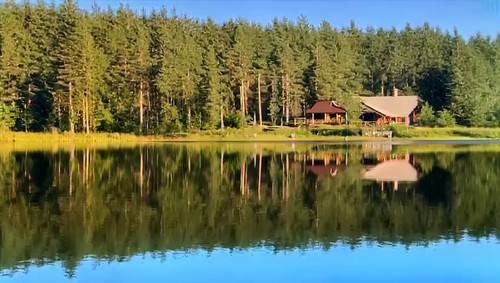}\hfill
\includegraphics[width=0.195\linewidth]{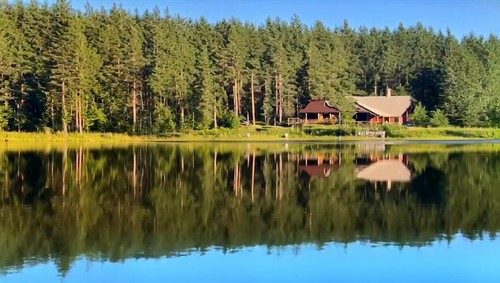}\hfill
\includegraphics[width=0.195\linewidth]{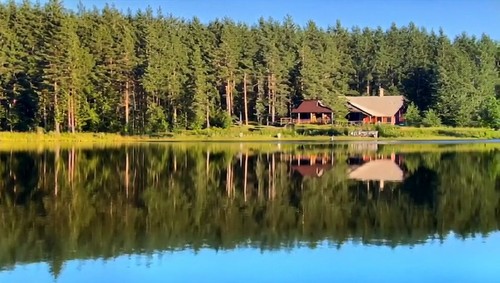}\hfill
\includegraphics[width=0.195\linewidth]{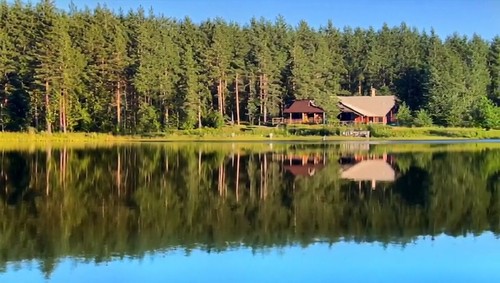}\hfill
\includegraphics[width=0.195\linewidth]{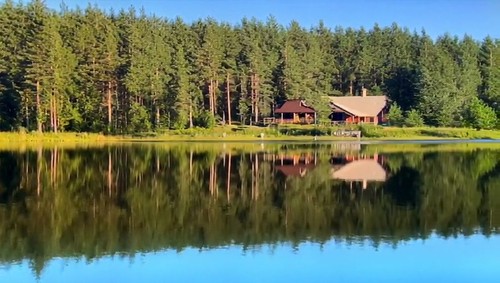}

\vspace{0.4em}

\makebox[\linewidth]{%
    \llap{\textcolor{green!60!black}{sparsity=69\%}\,\,\hspace{\marginMetrics}}
    \makebox[0pt][c]{\algoname}%
    \makebox[0pt][l]{\hspace{\marginMetrics}
        \textcolor{green!60!black}{latency=2m41s}
    }%
}

\vspace{0.2em}
\includegraphics[width=0.195\linewidth]{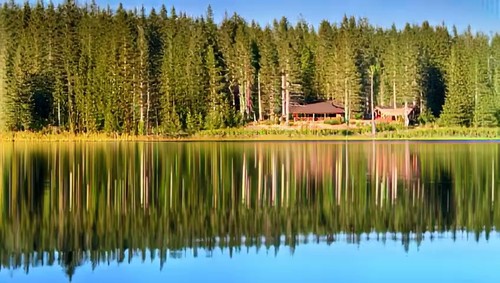}\hfill
\includegraphics[width=0.195\linewidth]{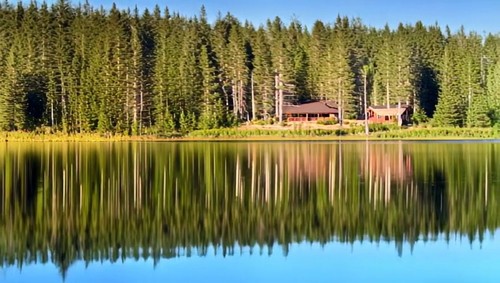}\hfill
\includegraphics[width=0.195\linewidth]{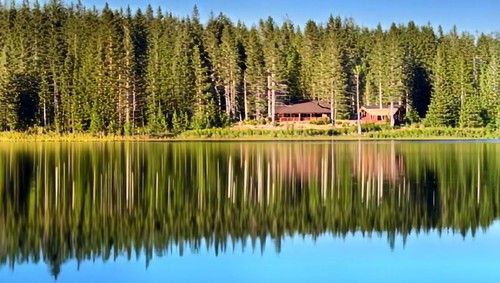}\hfill
\includegraphics[width=0.195\linewidth]{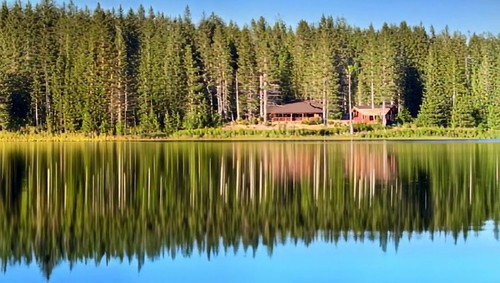}\hfill
\includegraphics[width=0.195\linewidth]{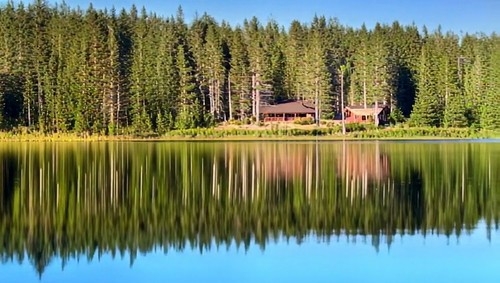}

\parbox{\linewidth}{\centering
\footnotesize \textit{\textcolor{black!70}{``\textD''}}
}

}

\caption{
Comparison of four prompts generated with the same seed on Mochi.
\algoname\ achieves higher attention sparsity and lower end-to-end latency while maintaining visual quality and prompt alignment.
}
\label{fig:sup_two_prompts_four_rows_mochi}

\end{figure}

%% file: figures/video_frame_comp_sup_480p_dist.tex
\def\promptA{Bigfoot}
\def\promptB{cat_pool}
\def\promptC{Couple}
\def\promptD{corgi_drum}

\def\textA{A bigfoot walking in the snowstorm.}
\def\textB{A cat wearing sunglasses and working as a lifeguard at a pool.}
\def\textC{A couple in formal evening wear going home... downpour with umbrellas.}
\def\textD{A corgi is playing drum kit.}

\def\marginMetrics{5.25em}

\begin{figure}[!h]
\centering
\vspace{-1.0em}

\setlength{\fboxsep}{2pt}

\parbox{\linewidth}{%

\makebox[\linewidth]{%
    \llap{\textcolor{red}{sparsity=0\%}\,\,\hspace{\marginMetrics}}
    \makebox[0pt][c]{Dense Attention}%
    \makebox[0pt][l]{\hspace{\marginMetrics}
        \textcolor{red}{latency=14s}
    }%
}

\vspace{0.2em}
\includegraphics[width=0.195\linewidth]{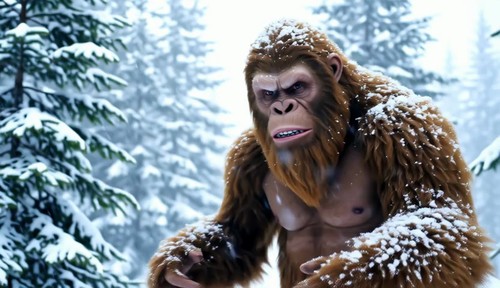}\hfill
\includegraphics[width=0.195\linewidth]{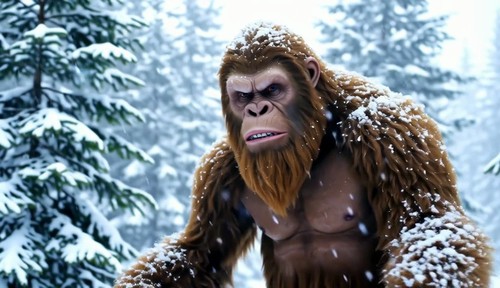}\hfill
\includegraphics[width=0.195\linewidth]{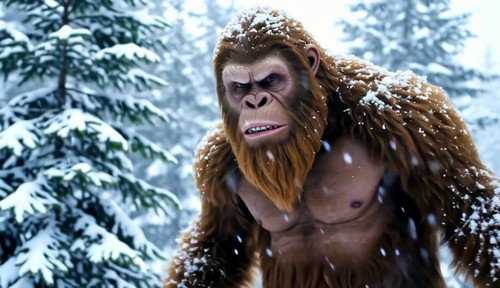}\hfill
\includegraphics[width=0.195\linewidth]{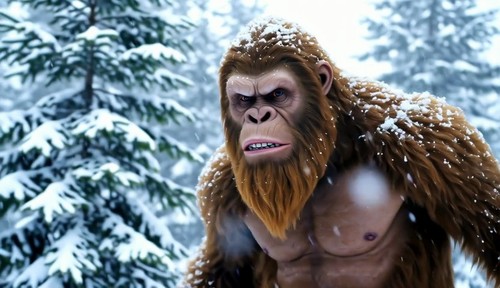}\hfill
\includegraphics[width=0.195\linewidth]{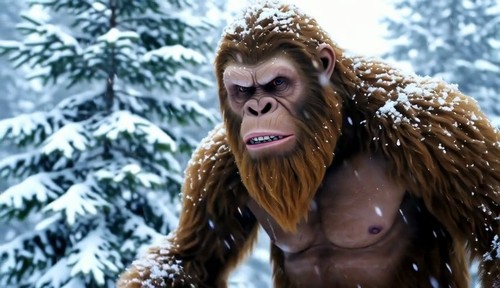}

\vspace{0.4em}

\makebox[\linewidth]{%
    \llap{\textcolor{green!60!black}{sparsity=70\%}\,\,\hspace{\marginMetrics}}
    \makebox[0pt][c]{\algoname}%
    \makebox[0pt][l]{\hspace{\marginMetrics}
        \textcolor{green!60!black}{latency=11s}
    }%
}

\vspace{0.2em}
\includegraphics[width=0.195\linewidth]{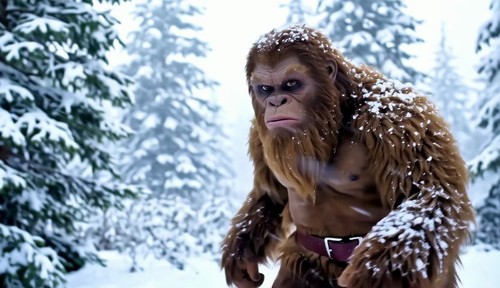}\hfill
\includegraphics[width=0.195\linewidth]{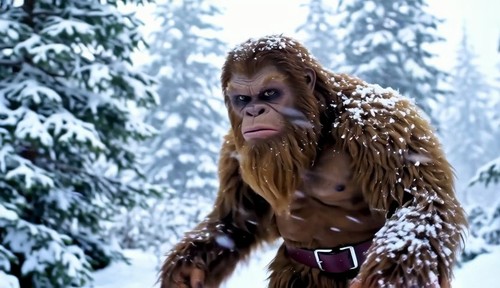}\hfill
\includegraphics[width=0.195\linewidth]{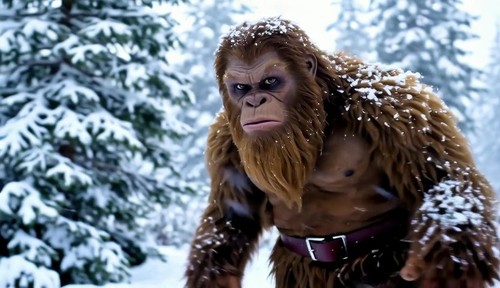}\hfill
\includegraphics[width=0.195\linewidth]{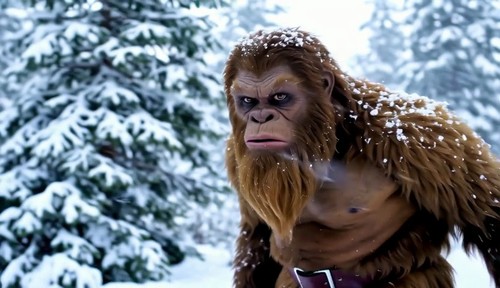}\hfill
\includegraphics[width=0.195\linewidth]{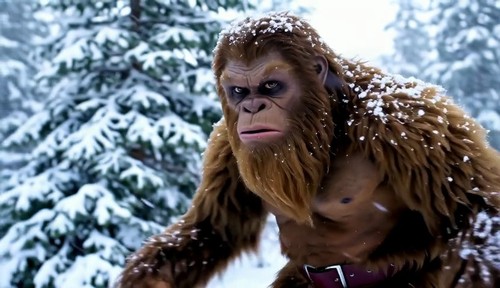}

\parbox{\linewidth}{\centering
\footnotesize \textit{\textcolor{black!70}{``\textA''}}
}

}

\vspace{1.0em}

\parbox{\linewidth}{%

\makebox[\linewidth]{%
    \llap{\textcolor{red}{sparsity=0\%}\,\,\hspace{\marginMetrics}}
    \makebox[0pt][c]{Dense Attention}%
    \makebox[0pt][l]{\hspace{\marginMetrics}
        \textcolor{red}{latency=14s}
    }%
}

\vspace{0.2em}
\includegraphics[width=0.195\linewidth]{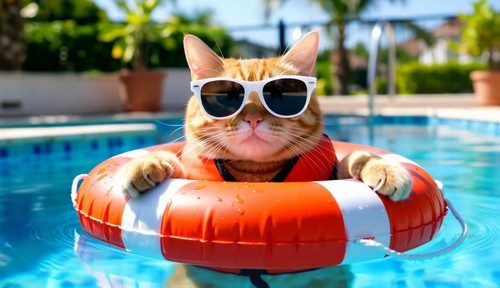}\hfill
\includegraphics[width=0.195\linewidth]{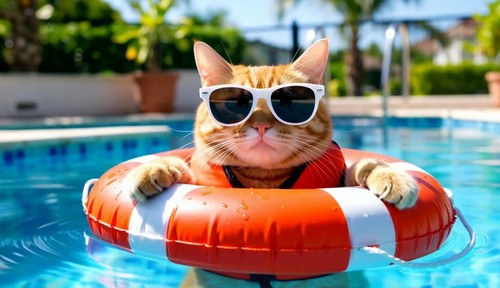}\hfill
\includegraphics[width=0.195\linewidth]{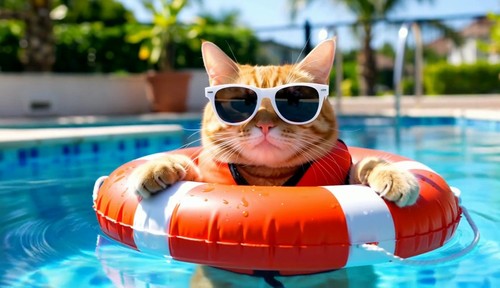}\hfill
\includegraphics[width=0.195\linewidth]{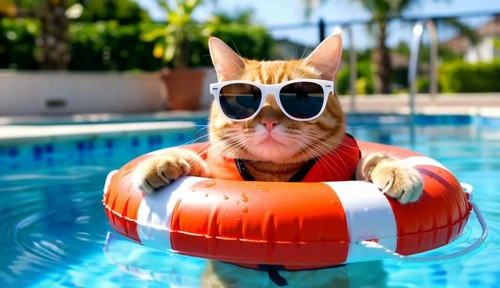}\hfill
\includegraphics[width=0.195\linewidth]{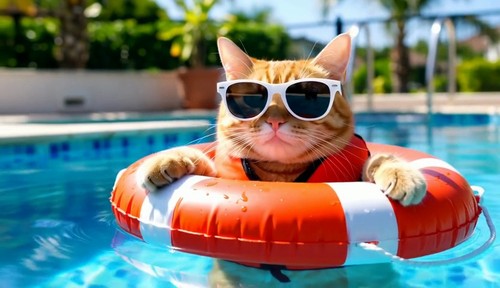}

\vspace{0.4em}

\makebox[\linewidth]{%
    \llap{\textcolor{green!60!black}{sparsity=70\%}\,\,\hspace{\marginMetrics}}
    \makebox[0pt][c]{\algoname}%
    \makebox[0pt][l]{\hspace{\marginMetrics}
        \textcolor{green!60!black}{latency=11s}
    }%
}

\vspace{0.2em}
\includegraphics[width=0.195\linewidth]{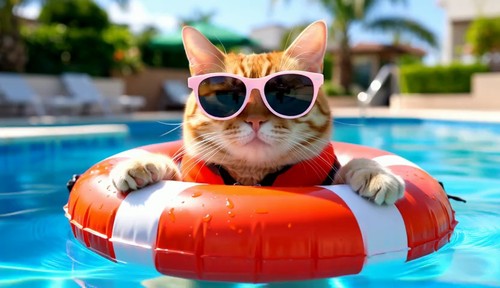}\hfill
\includegraphics[width=0.195\linewidth]{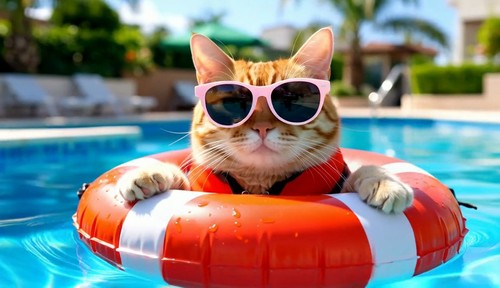}\hfill
\includegraphics[width=0.195\linewidth]{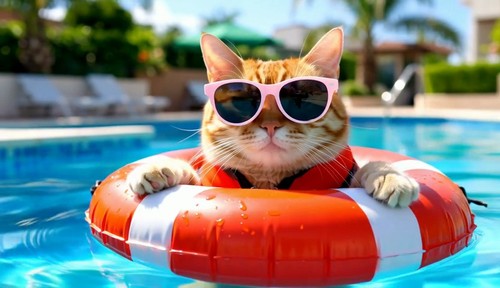}\hfill
\includegraphics[width=0.195\linewidth]{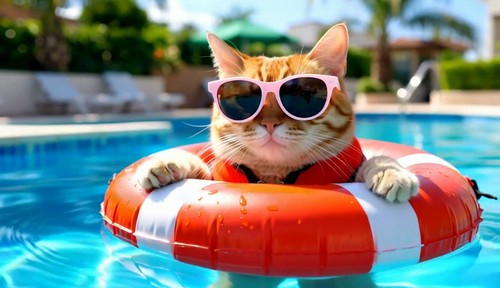}\hfill
\includegraphics[width=0.195\linewidth]{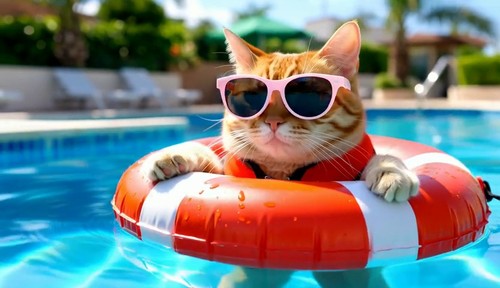}

\parbox{\linewidth}{\centering
\footnotesize \textit{\textcolor{black!70}{``\textB''}}
}

}

\vspace{1.0em}

\parbox{\linewidth}{%

\makebox[\linewidth]{%
    \llap{\textcolor{red}{sparsity=0\%}\,\,\hspace{\marginMetrics}}
    \makebox[0pt][c]{Dense Attention}%
    \makebox[0pt][l]{\hspace{\marginMetrics}
        \textcolor{red}{latency=14s}
    }%
}

\vspace{0.2em}
\includegraphics[width=0.195\linewidth]{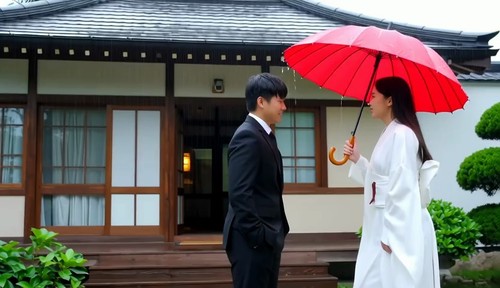}\hfill
\includegraphics[width=0.195\linewidth]{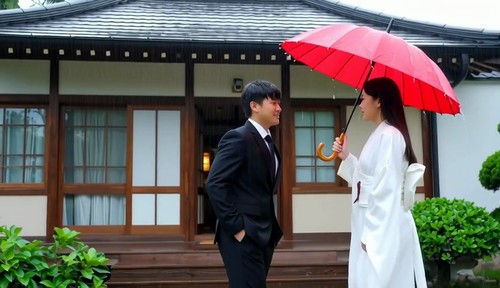}\hfill
\includegraphics[width=0.195\linewidth]{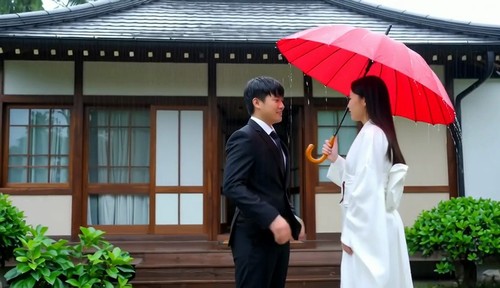}\hfill
\includegraphics[width=0.195\linewidth]{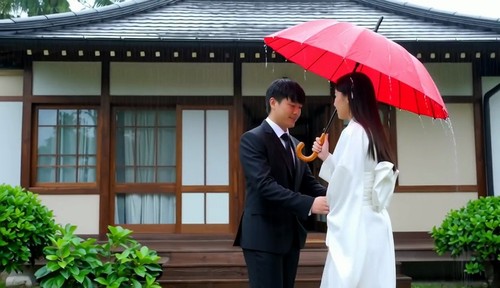}\hfill
\includegraphics[width=0.195\linewidth]{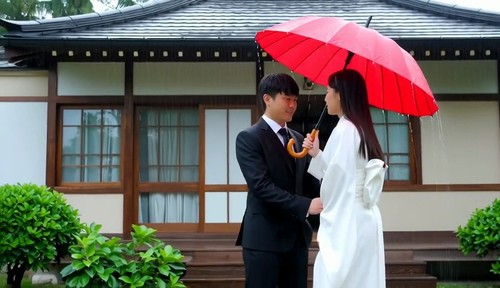}

\vspace{0.4em}

\makebox[\linewidth]{%
    \llap{\textcolor{green!60!black}{sparsity=70\%}\,\,\hspace{\marginMetrics}}
    \makebox[0pt][c]{\algoname}%
    \makebox[0pt][l]{\hspace{\marginMetrics}
        \textcolor{green!60!black}{latency=11s}
    }%
}

\vspace{0.2em}
\includegraphics[width=0.195\linewidth]{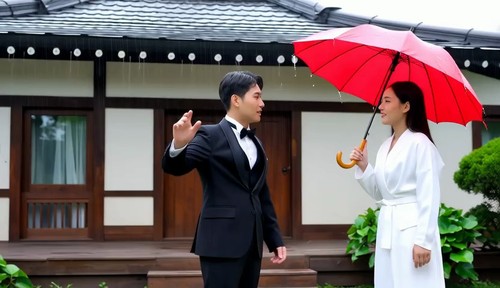}\hfill
\includegraphics[width=0.195\linewidth]{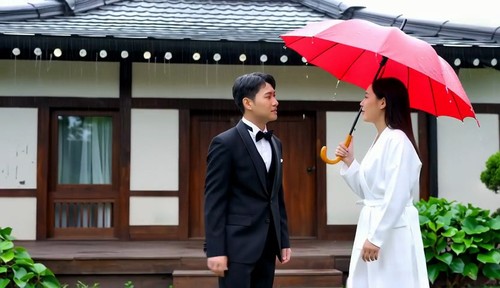}\hfill
\includegraphics[width=0.195\linewidth]{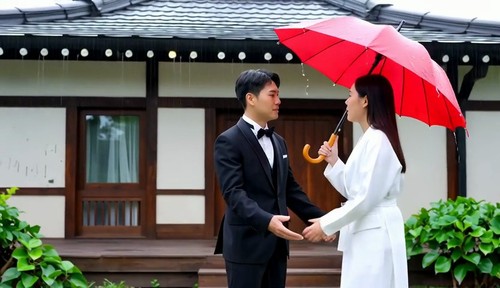}\hfill
\includegraphics[width=0.195\linewidth]{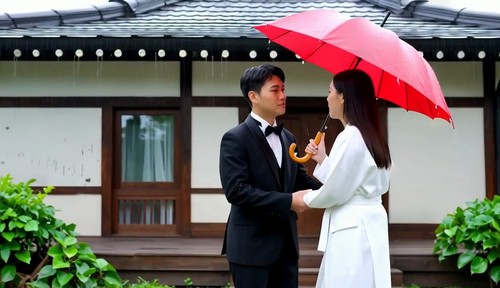}\hfill
\includegraphics[width=0.195\linewidth]{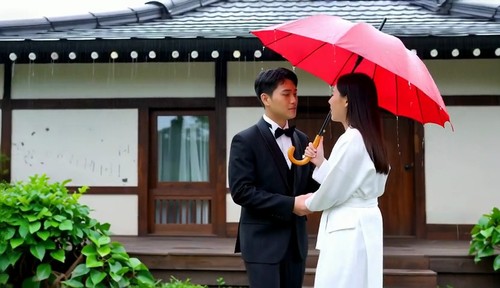}

\parbox{\linewidth}{\centering
\footnotesize \textit{\textcolor{black!70}{``\textC''}}
}

}

\vspace{1.0em}

\parbox{\linewidth}{%

\makebox[\linewidth]{%
    \llap{\textcolor{red}{sparsity=0\%}\,\,\hspace{\marginMetrics}}
    \makebox[0pt][c]{Dense Attention}%
    \makebox[0pt][l]{\hspace{\marginMetrics}
        \textcolor{red}{latency=14s}
    }%
}

\vspace{0.2em}
\includegraphics[width=0.195\linewidth]{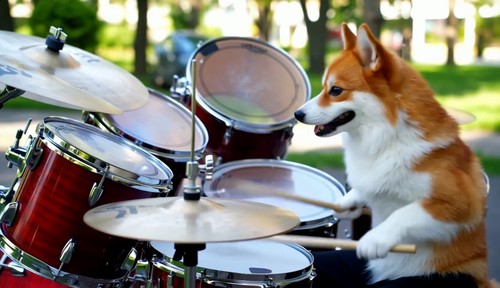}\hfill
\includegraphics[width=0.195\linewidth]{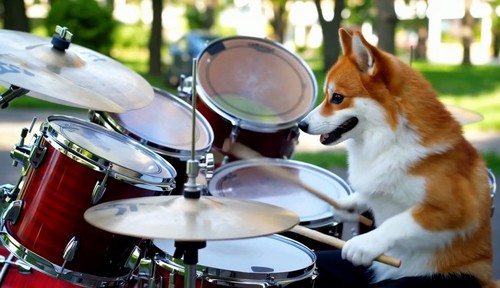}\hfill
\includegraphics[width=0.195\linewidth]{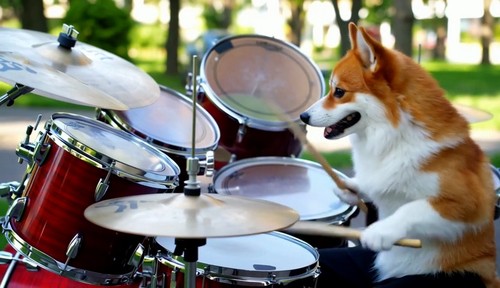}\hfill
\includegraphics[width=0.195\linewidth]{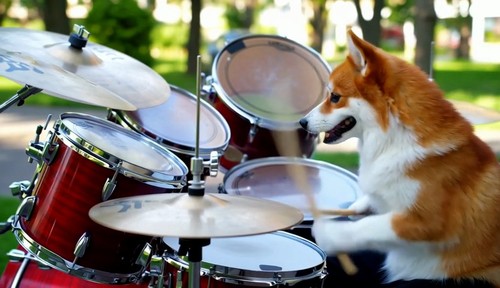}\hfill
\includegraphics[width=0.195\linewidth]{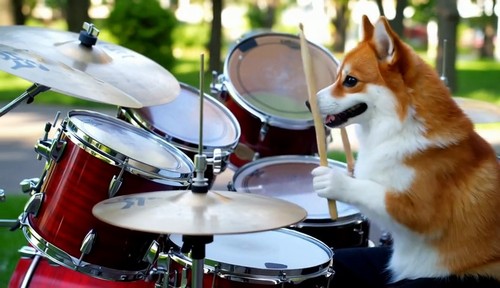}

\vspace{0.4em}

\makebox[\linewidth]{%
    \llap{\textcolor{green!60!black}{sparsity=70\%}\,\,\hspace{\marginMetrics}}
    \makebox[0pt][c]{\algoname}%
    \makebox[0pt][l]{\hspace{\marginMetrics}
        \textcolor{green!60!black}{latency=11s}
    }%
}

\vspace{0.2em}
\includegraphics[width=0.195\linewidth]{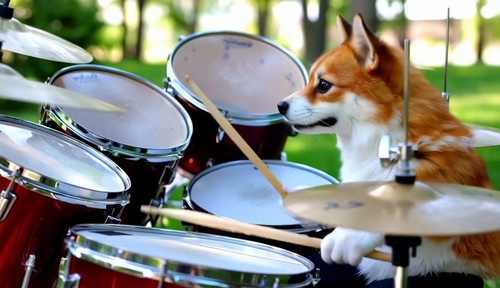}\hfill
\includegraphics[width=0.195\linewidth]{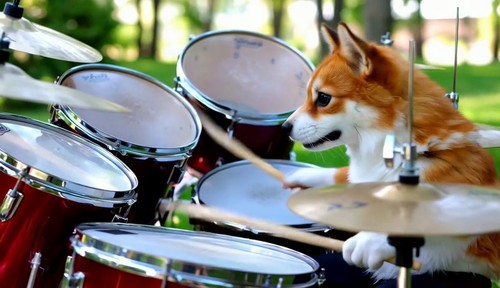}\hfill
\includegraphics[width=0.195\linewidth]{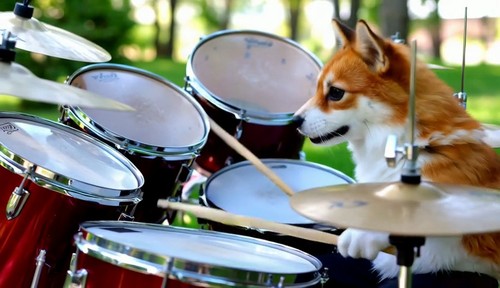}\hfill
\includegraphics[width=0.195\linewidth]{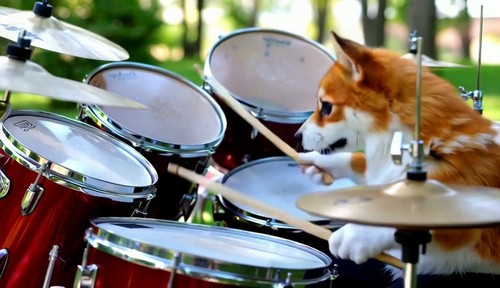}\hfill
\includegraphics[width=0.195\linewidth]{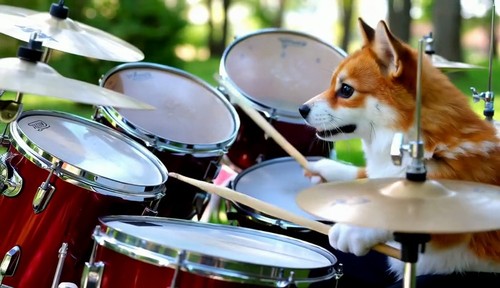}

\parbox{\linewidth}{\centering
\footnotesize \textit{\textcolor{black!70}{``\textD''}}
}

}

\caption{
Comparison of four prompts generated with the same seed on LightX2V (distilled Wan2.1 14B) at 480p.
}
\label{fig:sup_two_prompts_four_rows_distilled_480p}

\end{figure}

%% file: figures/video_frame_comp_sup_720p_dist.tex
\def\promptA{Bunny}
\def\promptB{Camp}
\def\promptC{Teddy}
\def\promptD{astro_duck}

\def\textA{A fat rabbit wearing a purple robe walking through a fantasy landscape.}
\def\textB{A happy fuzzy panda playing guitar nearby a campfire, snow mountain...}
\def\textC{A teddy bear is playing drum kit in NYC Times Square.}
\def\textD{An astronaut feeding ducks on a sunny afternoon, reflection from the water.}

\def\marginMetrics{5.25em}

\begin{figure}[!h]
\centering
\vspace{-1.0em}

\setlength{\fboxsep}{2pt}

\parbox{\linewidth}{%

\makebox[\linewidth]{%
    \llap{\textcolor{red}{sparsity=0\%}\,\,\hspace{\marginMetrics}}
    \makebox[0pt][c]{Dense Attention}%
    \makebox[0pt][l]{\hspace{\marginMetrics}
        \textcolor{red}{latency=48s}
    }%
}

\vspace{0.2em}
\includegraphics[width=0.195\linewidth]{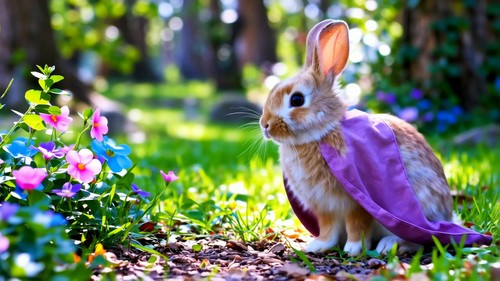}\hfill
\includegraphics[width=0.195\linewidth]{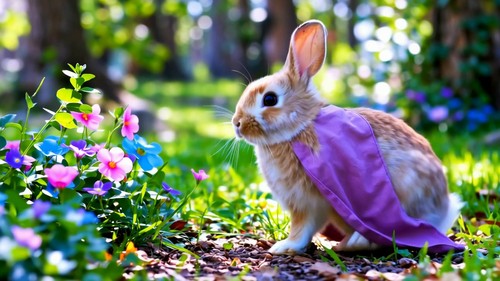}\hfill
\includegraphics[width=0.195\linewidth]{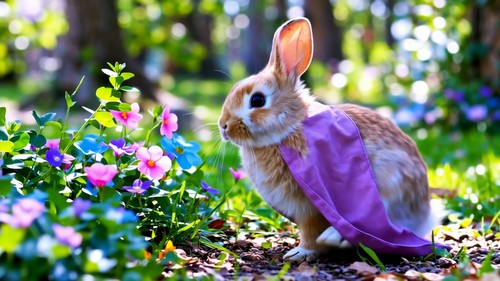}\hfill
\includegraphics[width=0.195\linewidth]{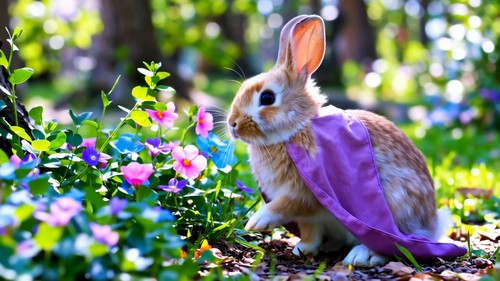}\hfill
\includegraphics[width=0.195\linewidth]{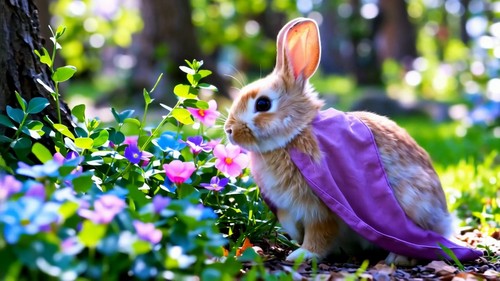}

\vspace{0.4em}

\makebox[\linewidth]{%
    \llap{\textcolor{green!60!black}{sparsity=74\%}\,\,\hspace{\marginMetrics}}
    \makebox[0pt][c]{\algoname}%
    \makebox[0pt][l]{\hspace{\marginMetrics}
        \textcolor{green!60!black}{latency=30s}
    }%
}

\vspace{0.2em}
\includegraphics[width=0.195\linewidth]{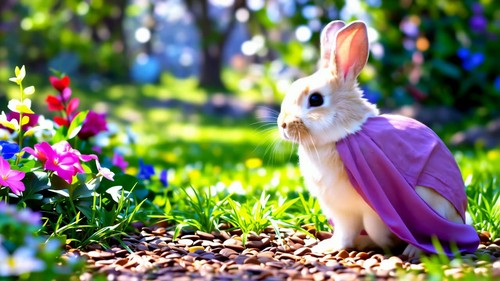}\hfill
\includegraphics[width=0.195\linewidth]{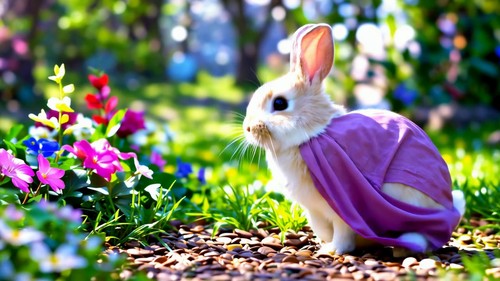}\hfill
\includegraphics[width=0.195\linewidth]{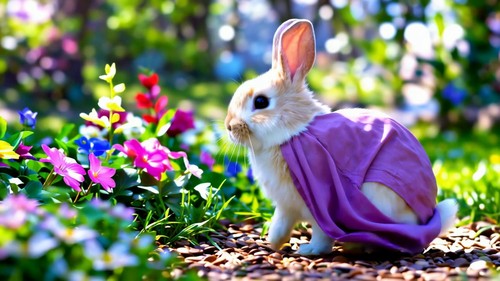}\hfill
\includegraphics[width=0.195\linewidth]{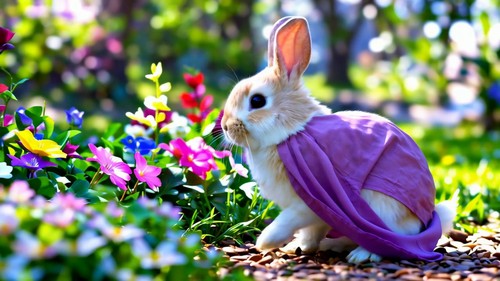}\hfill
\includegraphics[width=0.195\linewidth]{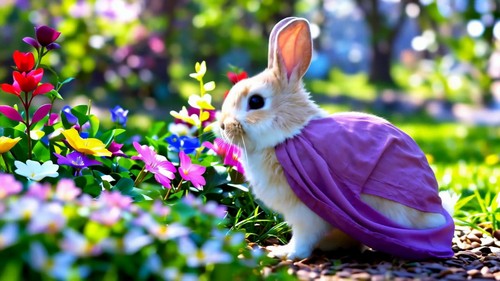}

\parbox{\linewidth}{\centering
\footnotesize \textit{\textcolor{black!70}{``\textA''}}
}

}

\vspace{1.0em}

\parbox{\linewidth}{%

\makebox[\linewidth]{%
    \llap{\textcolor{red}{sparsity=0\%}\,\,\hspace{\marginMetrics}}
    \makebox[0pt][c]{Dense Attention}%
    \makebox[0pt][l]{\hspace{\marginMetrics}
        \textcolor{red}{latency=48s}
    }%
}

\vspace{0.2em}
\includegraphics[width=0.195\linewidth]{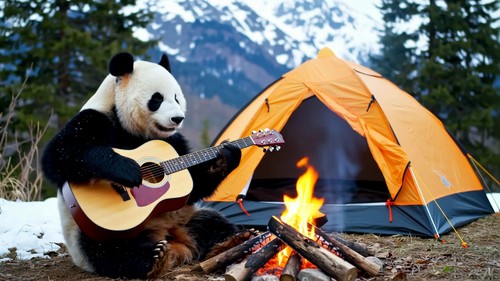}\hfill
\includegraphics[width=0.195\linewidth]{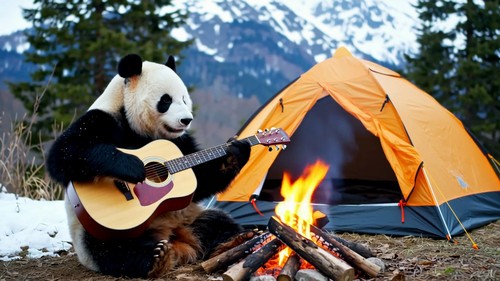}\hfill
\includegraphics[width=0.195\linewidth]{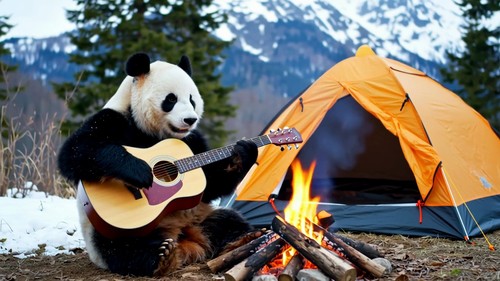}\hfill
\includegraphics[width=0.195\linewidth]{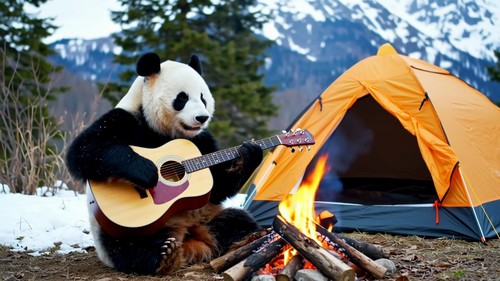}\hfill
\includegraphics[width=0.195\linewidth]{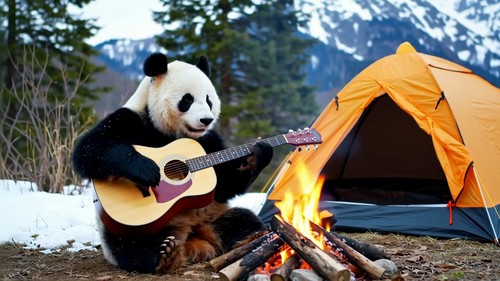}

\vspace{0.4em}

\makebox[\linewidth]{%
    \llap{\textcolor{green!60!black}{sparsity=74\%}\,\,\hspace{\marginMetrics}}
    \makebox[0pt][c]{\algoname}%
    \makebox[0pt][l]{\hspace{\marginMetrics}
        \textcolor{green!60!black}{latency=30s}
    }%
}

\vspace{0.2em}
\includegraphics[width=0.195\linewidth]{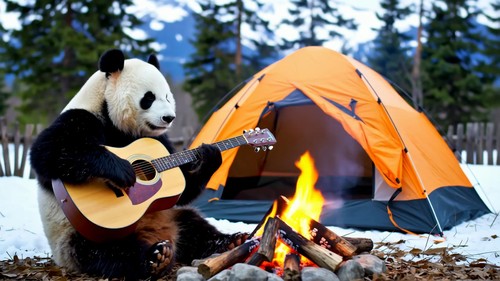}\hfill
\includegraphics[width=0.195\linewidth]{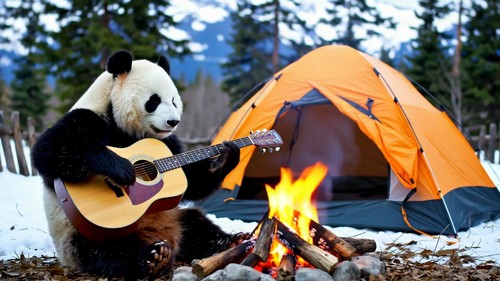}\hfill
\includegraphics[width=0.195\linewidth]{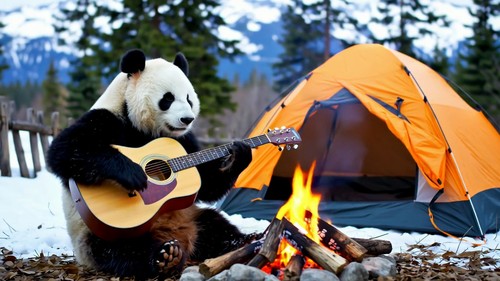}\hfill
\includegraphics[width=0.195\linewidth]{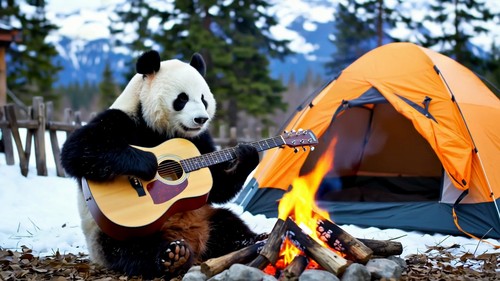}\hfill
\includegraphics[width=0.195\linewidth]{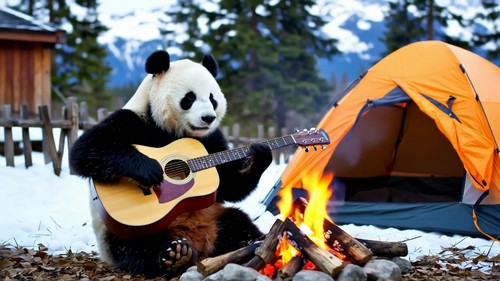}

\parbox{\linewidth}{\centering
\footnotesize \textit{\textcolor{black!70}{``\textB''}}
}

}

\vspace{1.0em}

\parbox{\linewidth}{%

\makebox[\linewidth]{%
    \llap{\textcolor{red}{sparsity=0\%}\,\,\hspace{\marginMetrics}}
    \makebox[0pt][c]{Dense Attention}%
    \makebox[0pt][l]{\hspace{\marginMetrics}
        \textcolor{red}{latency=48s}
    }%
}

\vspace{0.2em}
\includegraphics[width=0.195\linewidth]{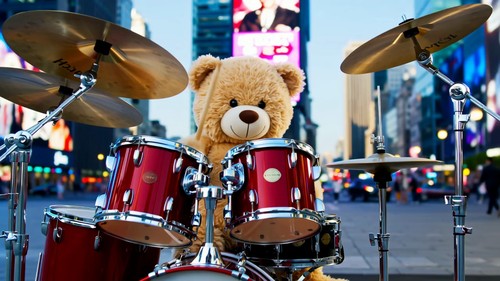}\hfill
\includegraphics[width=0.195\linewidth]{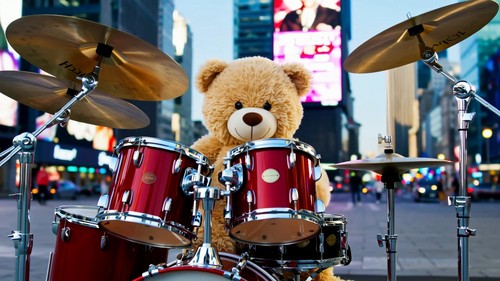}\hfill
\includegraphics[width=0.195\linewidth]{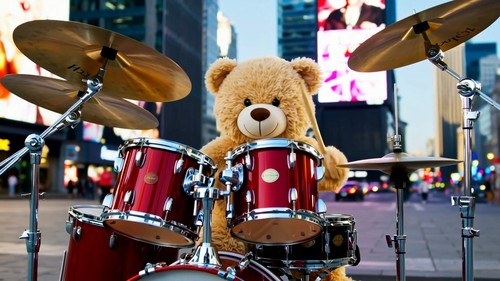}\hfill
\includegraphics[width=0.195\linewidth]{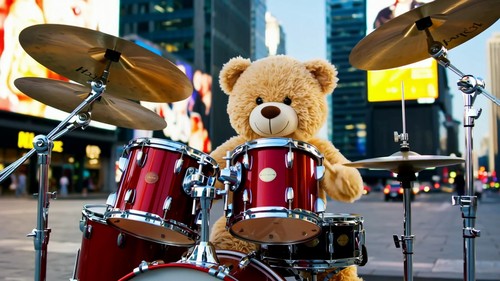}\hfill
\includegraphics[width=0.195\linewidth]{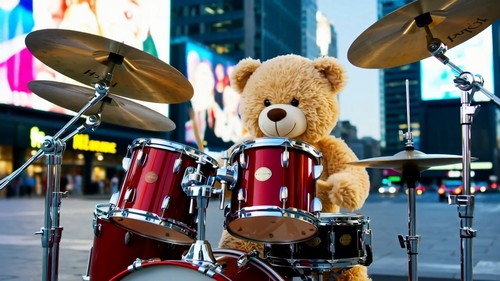}

\vspace{0.4em}

\makebox[\linewidth]{%
    \llap{\textcolor{green!60!black}{sparsity=74\%}\,\,\hspace{\marginMetrics}}
    \makebox[0pt][c]{\algoname}%
    \makebox[0pt][l]{\hspace{\marginMetrics}
        \textcolor{green!60!black}{latency=30s}
    }%
}

\vspace{0.2em}
\includegraphics[width=0.195\linewidth]{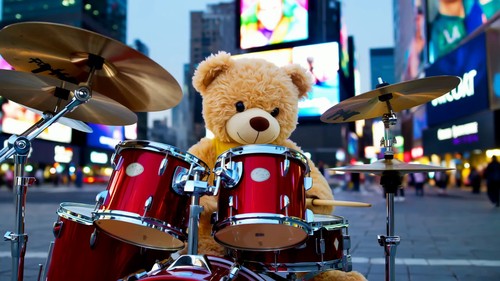}\hfill
\includegraphics[width=0.195\linewidth]{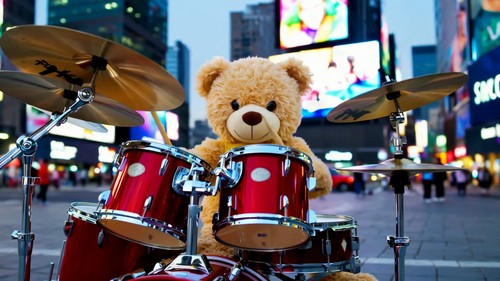}\hfill
\includegraphics[width=0.195\linewidth]{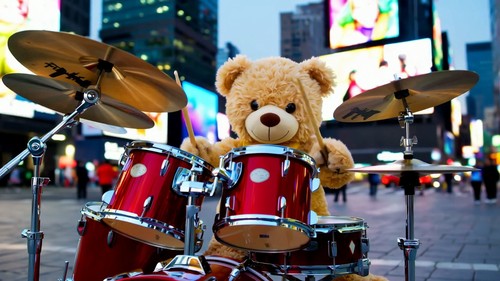}\hfill
\includegraphics[width=0.195\linewidth]{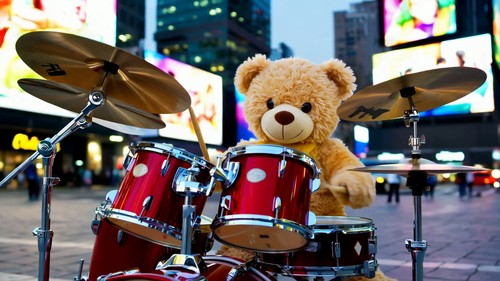}\hfill
\includegraphics[width=0.195\linewidth]{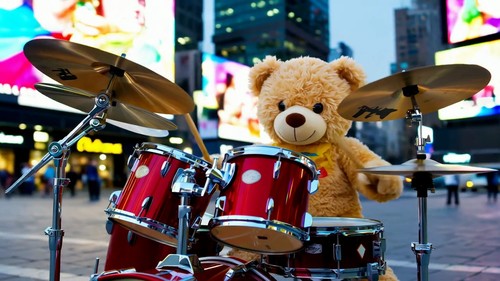}

\parbox{\linewidth}{\centering
\footnotesize \textit{\textcolor{black!70}{``\textC''}}
}

}

\vspace{1.0em}

\parbox{\linewidth}{%

\makebox[\linewidth]{%
    \llap{\textcolor{red}{sparsity=0\%}\,\,\hspace{\marginMetrics}}
    \makebox[0pt][c]{Dense Attention}%
    \makebox[0pt][l]{\hspace{\marginMetrics}
        \textcolor{red}{latency=48s}
    }%
}

\vspace{0.2em}
\includegraphics[width=0.195\linewidth]{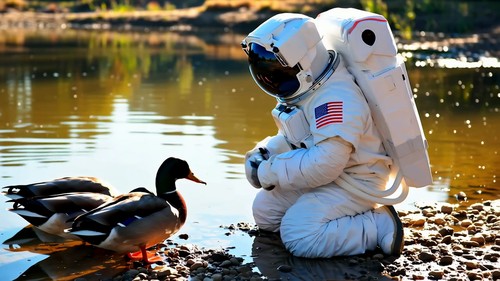}\hfill
\includegraphics[width=0.195\linewidth]{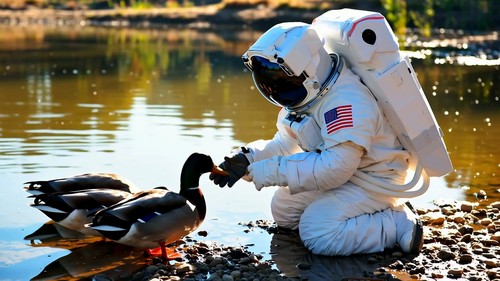}\hfill
\includegraphics[width=0.195\linewidth]{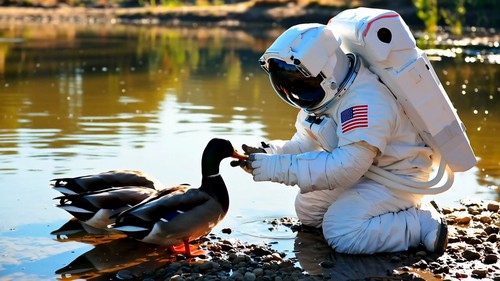}\hfill
\includegraphics[width=0.195\linewidth]{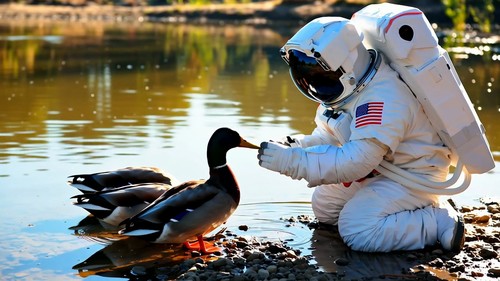}\hfill
\includegraphics[width=0.195\linewidth]{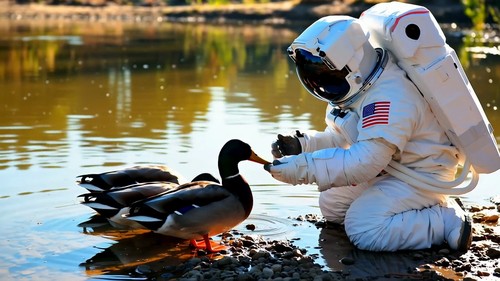}

\vspace{0.4em}

\makebox[\linewidth]{%
    \llap{\textcolor{green!60!black}{sparsity=74\%}\,\,\hspace{\marginMetrics}}
    \makebox[0pt][c]{\algoname}%
    \makebox[0pt][l]{\hspace{\marginMetrics}
        \textcolor{green!60!black}{latency=30s}
    }%
}

\vspace{0.2em}
\includegraphics[width=0.195\linewidth]{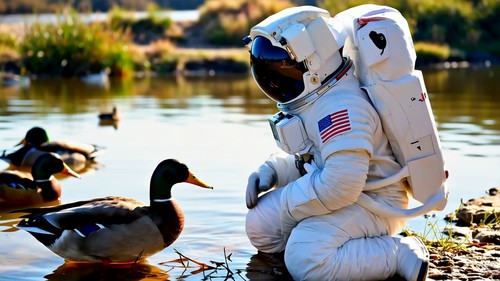}\hfill
\includegraphics[width=0.195\linewidth]{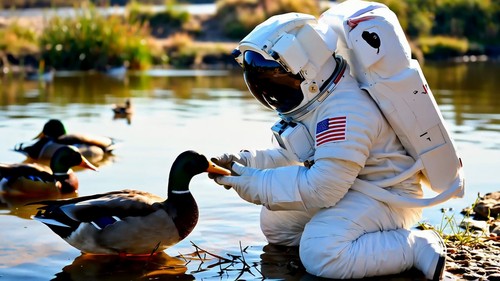}\hfill
\includegraphics[width=0.195\linewidth]{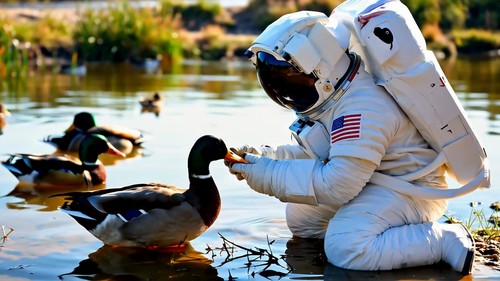}\hfill
\includegraphics[width=0.195\linewidth]{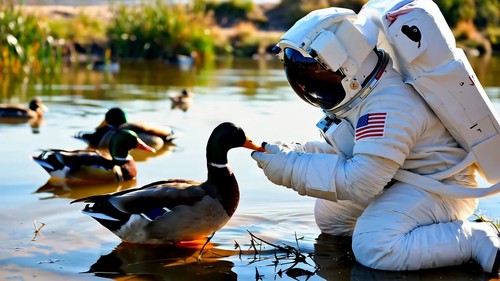}\hfill
\includegraphics[width=0.195\linewidth]{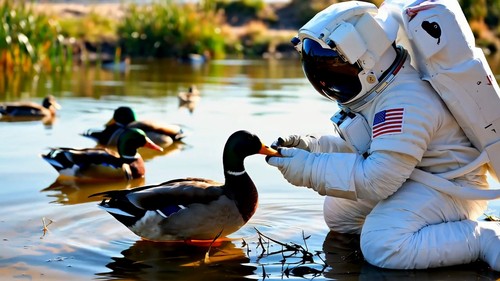}

\parbox{\linewidth}{\centering
\footnotesize \textit{\textcolor{black!70}{``\textD''}}
}

}

\caption{
Comparison of four prompts generated with the same seed on LightX2V (distilled Wan2.1 14B) at 720p.
}
\label{fig:sup_two_prompts_four_rows_distilled_720p}

\end{figure}